\pdfoutput=1

\documentclass[11pt]{article}

\usepackage[final]{acl}

\usepackage{times}
\usepackage{latexsym}

\usepackage[T1]{fontenc}

\usepackage[utf8]{inputenc}

\usepackage{microtype}

\usepackage{inconsolata}

\usepackage{graphicx}
\usepackage{subcaption}
\usepackage{pgfplots}
\usepackage{amsmath}
\usepackage{multirow}
\usepackage{multicol}

\title{The Hidden Space of Safety:\\ Understanding Preference-Tuned LLMs in Multilingual context}

\author{Nikhil Verma \\
  LG Toronto AI Research lab\\
  \texttt{nikhil.verma@lge.com} \\
  \And
  Manasa Bharadwaj \\
  LG Toronto AI Research lab \\
  \texttt{manasa.bharadwaj@lge.com}
  }

\begin{document}
\maketitle
\begin{abstract}
Alignment tuning has enabled large language models to excel in reasoning, instruction-following, and minimizing harmful generations.
However, despite their widespread deployment, these models exhibit a monolingual bias, raising concerns about the effectiveness of alignment across languages.
Current alignment methods predominantly focus on English, leaving it unclear how alignment mechanism generalize to multilingual settings.
To address this, we conduct a systematic analysis of distributional shifts in the embedding space of LLMs before and after alignment, uncovering its impact on model behavior across diverse languages.
We leverage the alignment-induced separation in safety space as a quantitative tool to measure how alignment enforces safety constraints.
Our study evaluates seven LLMs using balanced toxicity datasets and parallel text-detoxification benchmarks, revealing substantial disparities in the latent representation space between high-resource and low-resource languages .
These findings underscore the need for language-specific fine-tuning to ensure fair, reliable and robust multilingual alignment.
Our insights provide a foundation for developing truly safe multilingual LLMs, emphasizing the urgency of addressing alignment gaps in underrepresented languages.

\end{abstract}

\section{Introduction}
\label{introduction}

Alignment techniques play a critical role in adapting Large Language Models (LLMs) beyond their pre-training and fine-tuning phases, ensuring consistency with human values and preferences~\cite{christiano2017deep, ziegler2019fine}.
To achieve this, methods based on online and offline policy optimization~\cite{ouyang2022training, rafailov2023direct, ethayarajh2024kto, haldar2025llm} have been introduced to enhance model reliability, safety, and fairness for real-world deployment.
%
%
However, despite being trained on multilingual corpora, LLM alignment remains predominantly optimized for English, leading to disparities in performance and behavior across languages~\cite{schwartz2022towards, vashishtha2023evaluating}.
This discrepancy raises critical concerns about security, and usability, particularly for underrepresented languages, where alignment remains underexplored and its effectiveness is poorly understood~\cite{rystrom2025multilingual, khandelwal2023casteist}.
A systematic investigation into the cross-lingual impact of alignment is essential to understand its influence on model representations and behavior across English and other languages.

\begin{figure}[t]
    \centering
    \begin{subfigure}[b]{0.22\textwidth} 
        \centering
        \includegraphics[width=\textwidth]{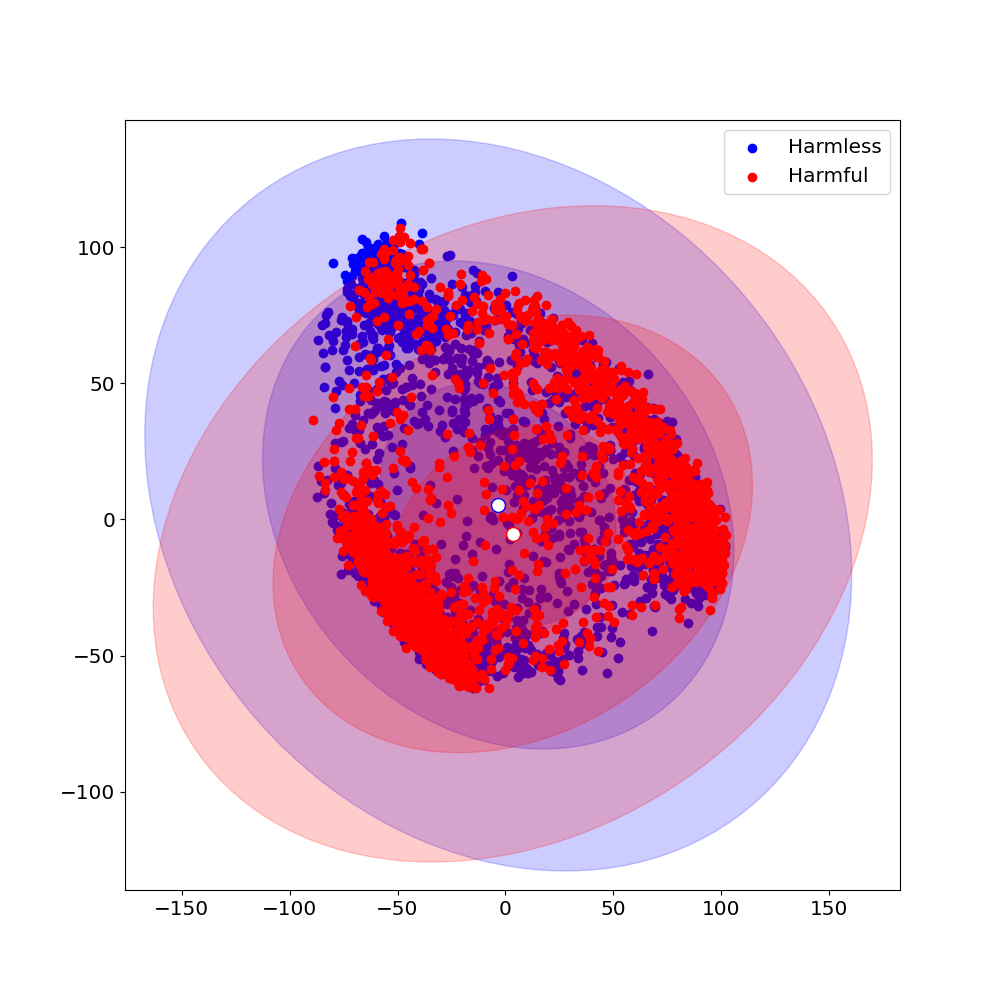}
        \caption{Before Alignment}
        \label{fig:first}
    \end{subfigure}
    \hfill  
    \begin{subfigure}[b]{0.22\textwidth}  
        \centering
        \includegraphics[width=\textwidth]{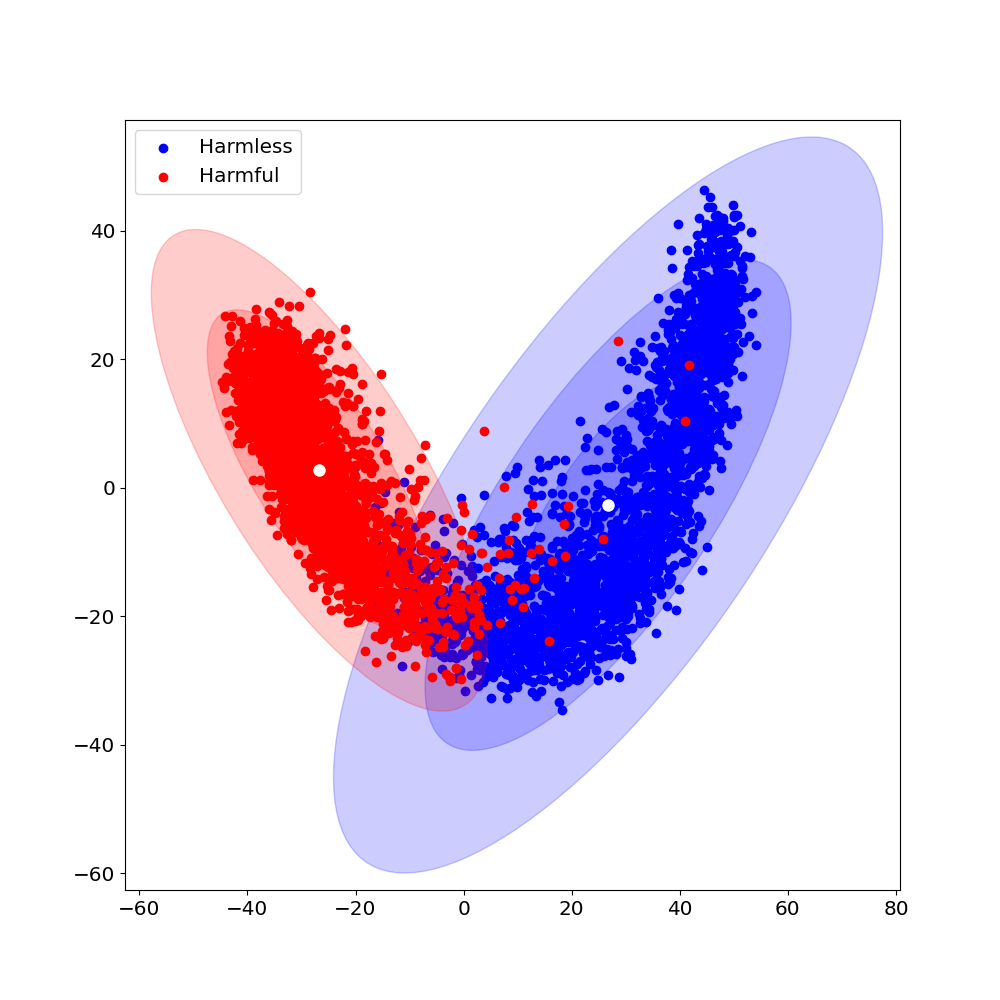}  
        \caption{After Alignment}
        \label{fig:second}
    \end{subfigure}

    \caption{Effect of Alignment on Hidden Representations in Llama-2 (\#7B) for English Prompt Safety.}
    \label{fig:before_after_alignment_llama2_7b_en}
\end{figure}

Beyond fairness concerns, misalignment in multilingual settings presents fundamental challenges in representation learning and decision boundaries within LLMs.
While alignment techniques refine model behavior in English, their impact on latent space organization across languages remains poorly understood.
Empirical evidence suggests that preference optimization shifts model representations~(Figure \ref{fig:before_after_alignment_llama2_7b_en}), yet such shifts may not generalize uniformly across linguistic groups~\cite{lin2024towards, kirk2023understanding}.
For languages with limited preference data, alignment may inadvertently distort decision boundaries, leading to semantic drift, degraded reasoning capabilities, or increased susceptibility to adversarial inputs~\cite{dang2024rlhf}.
This raises the pressing need to analyze how alignment reshapes the hidden space across languages and whether current methodologies effectively preserve semantic consistency while mitigating harmful outputs.
A deeper investigation into cross-lingual representation shifts can illuminate the unintended consequences of alignment, guiding the development of more robust multilingual models.

In this work, we systematically assess the effectiveness of LLM safety alignment across multiple languages by evaluating a diverse set of models on multilingual benchmarks.
We analyze distributional shifts in the embedding space of multilingual safety prompts for both reference and aligned models, uncovering how alignment mechanisms influence model behavior.
To quantify the separation induced by alignment in enforcing safety constraints, we utilized a set of distributional metrics, providing a concrete measure of alignment-induced shifts.
Our findings reveal critical gaps in multilingual safety mechanisms, highlighting the inconsistencies in alignment effectiveness across languages.

As deep generative models evolve, concerns regarding memorization and bias propagation have intensified~\cite{carlini2023extracting, biderman2023emergent, nasr2023scalable}.
To better isolate the effects of alignment, we analyze hidden representations through probing only at the input processing stage, minimizing potential contamination from post-training phases.
Our study uncovers significant performance disparities across model families, emphasizing the monolingual bias inherent in current LLM development.
These insights lay the groundwork for future fine-tuning efforts on language-specific datasets, guiding improvements in multilingual fairness, safety, and robustness.

\section{Related literature}
\label{related_literature}

\subsection{Alignment of LLMs}
\label{llm_alignment}
Fine-tuning LLMs based on human preferences has emerged as a key approach for post-training, enhancing their ability to generate responses aligned with human values \cite{christiano2017deep, ziegler2019fine, liu2020learning}.
A variety of techniques have been developed to achieve this alignment, starting with Reinforcement Learning from Human Feedback (RLHF), which remains the dominant paradigm for online policy optimization~\cite{ouyang2022training, ahmadian2024back}.
Beyond RLHF, several offline optimization methods, such as Direct Preference Optimization DPO~\cite{rafailov2023direct}, Implicit Preference Optimization IPO~\cite{azar2024general}, Kahneman-Tversky Optimization KTO~\cite{ethayarajh2024kto}, Bayesian Causal Optimization BCO~\cite{jung2024binary}, and KL divergence optimization KLDO~\cite{haldar2025llm} have been introduced to refine model behavior without requiring costly reinforcement learning loops.
The overarching objective of these techniques is to improve alignment with human intent while mitigating the risks of generating harmful or toxic content, especially in large-scale real-world deployments.

\subsection{Multilingual LLM performance}
\label{multilingual_llm_performance}
A critical limitation of current alignment strategies is their heavy reliance on human preference datasets predominantly sourced from English or other high-resource languages~\cite{taori2023stanford, chiang2023vicuna, wu2023lamini}.
As a result, LLMs exhibit strong alignment in English but struggle with maintaining consistent safety and ethical considerations in underrepresented languages~\cite{schwartz2022towards, vashishtha2023evaluating, khandelwal2023casteist}.
This imbalance raises concerns about fairness, as the effectiveness of alignment mechanisms can vary significantly across linguistic groups~\cite{rystrom2025multilingual}.
In particular, multilingual users may encounter unreliable moderation, differing levels of content safety, or unintended biases when interacting with aligned LLMs in languages beyond English~\cite{yong2023low}.

\subsection{Jailbreaking studies}
\label{jailbreaking_studies}

Despite significant advancements, the alignment strategies are not foolproof~\cite{li2023privacy, alkhamissi2024investigating}.
An emerging body of research explores how safety-aligned LLMs remain vulnerable to adversarial exploits, including jailbreaking attacks that expose weaknesses in alignment constraints~\cite{lukas2023analyzing, sun2024trustllm, liu2023jailbreaking}.
These vulnerabilities are further amplified in multilingual settings, where models may bypass safety filters in languages for which alignment data is scarce~\cite{winata2024preference, son2024mm}.
Addressing these gaps is crucial for ensuring that LLM alignment strategies are not only robust but also equitable across diverse linguistic and cultural contexts~\cite{dang2024rlhf}.

\begin{figure*}[ht]
\centering
  \includegraphics[scale=0.4]{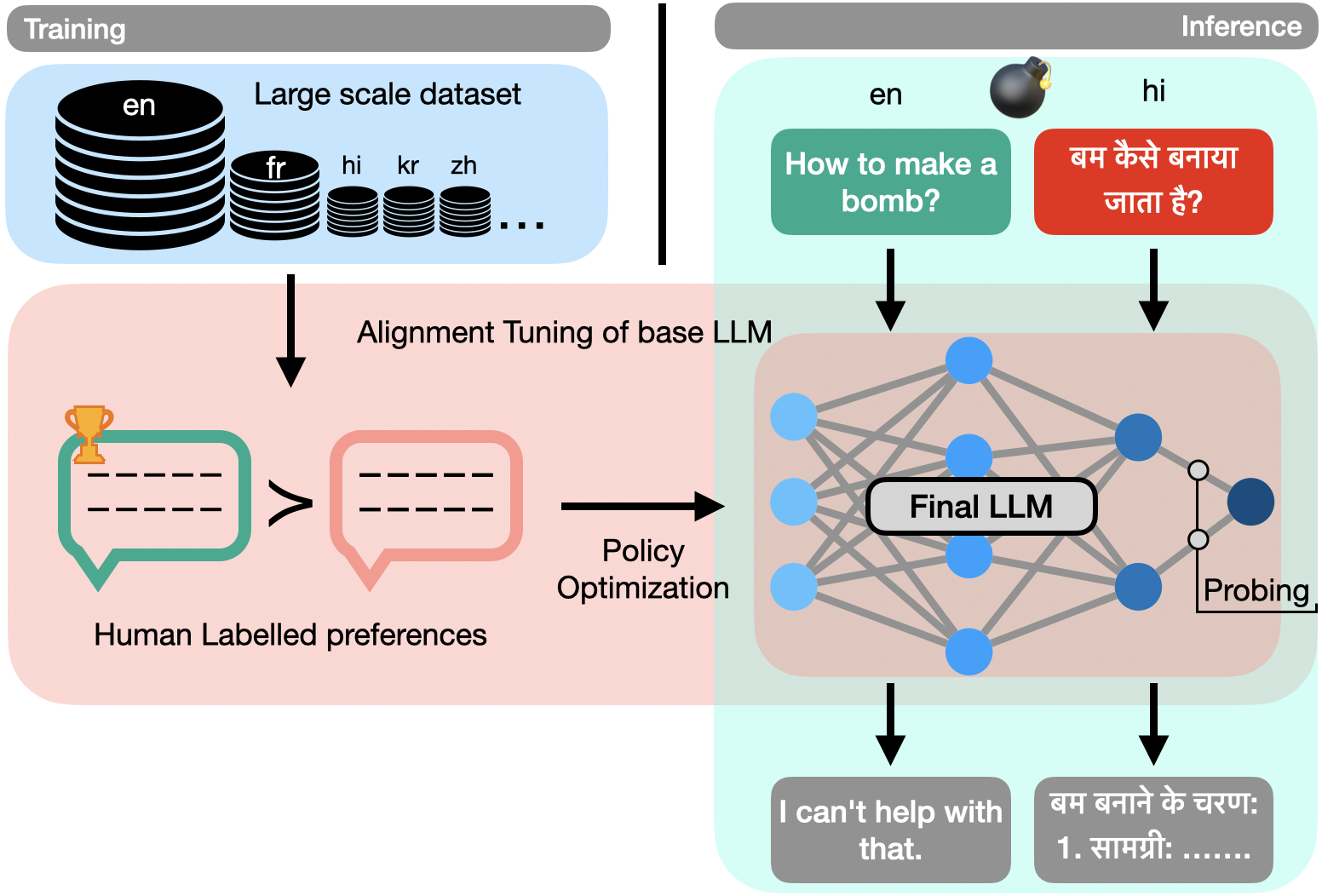}
  \caption {Probing the Impact of Human Preference Tuning on Multilingual Safety at Inference Time: Llama-3.1~(\#8B) Alignment in English vs. Hindi}
    \label{fig:multilingual_llm_alignment_failure_llama3.1_8b_instrcut}
\end{figure*}

\section{Background}
\label{background}

LLMs are typically trained in three distinct stages, each playing a crucial role in shaping their capabilities and alignment with human preferences.

\noindent \textbf{Pre-training.} LLMs are initially pre-trained on large-scale corpora spanning multiple languages, optimizing the likelihood of predicting the next token conditioned on preceding text. 
The model’s vocabulary is designed to accommodate tokens from diverse languages, ensuring broad linguistic coverage.

\noindent \textbf{Supervised Fine-tuning (SFT).} Following pre-training, LLMs undergo fine-tuning on curated, high-quality datasets specific to various downstream tasks, such as translation, dialogue, and summarization. 
This stage refines the model’s ability to generate more task-relevant responses and produces a reference model $\pi_{ref}$. 

\noindent \textbf{Human Preference Tuning.} The SFT model is prompted with inputs $x$ to generate pairs of candidate responses $(y_1, y_2)\sim \pi_{ref}(y \mid x)$. 
These responses are then presented to human annotators, who express a preference for one over the other, denoted as $y_w\succ y_l \mid x$, where $y_w$ and $y_l$ represent the preferred and dispreferred completions, respectively. 
Preference data is assumed to be generated from an underlying latent reward function $r^*(x, y)$, which models human preferences. 
A common approach to capturing the probability of a preference ranking is through the Bradley-Terry model~\cite{bradley1952rank}:
\begin{equation}\label{eq:bradley-terry}
    p^*(y_w\succ y_l \mid x)= \sigma(r^*(x, y_w)- r^*(x, y_l))
\end{equation}

where $\sigma$ denotes the logistic function.
More generally, when multiple ranked responses are available, Plackett-Luce models are also applicable~\cite{plackett1975analysis}. 
In large-scale preference datasets (shown in Figure \ref{fig:multilingual_llm_alignment_failure_llama3.1_8b_instrcut}), human-annotated feedback is predominantly available for high-resource languages such as English, while low-resource languages are underrepresented~\cite{chaudhari2024rlhf}. 
Since directly obtaining a true reward function from human feedback is infeasible, an alternative approach is to assume access to a static dataset of comparisons $\mathcal{D}=\bigl\{x^{(i)}, y_w^{(i)}, y_l^{(i)}\}_{i=1}^N$ sampled from $p^*$ and to parameterize a reward model $r_{\phi}(x, y)$. 
The parameters of this model are estimated via maximum likelihood:
\begin{equation}\label{eq:reward_model}
    {E}_{x, y_w, y_l\sim \mathcal{D}}\bigl[\log \sigma(r_{\phi}(x, y_w)- r_{\phi}(x, y_l))\bigr]
\end{equation}

To ensure that the optimized model maintains desirable text generation properties, such as fluency and coherence, a KL divergence penalty is introduced to limit deviation from $\pi_{ref}$. 
The reward model is then used to optimize a new policy $\pi_{\theta}$, formulated as:
\begin{equation}\label{eq:RL}
{E}_{x\in \mathcal{D}, y\sim \pi_{\theta}}[r_{\phi}(x, y)] - \beta{D}_{KL}[\pi_{\theta}\mid \mid \pi_{ref}]
\end{equation}

where $\beta > 0$ is a hyperparameter controlling the balance between reward maximization and distributional regularization. 

Since this objective is non-differentiable, reinforcement learning (e.g., Proximal Policy Optimization, PPO~\cite{schulman2017proximal}) is typically employed for policy updates. 
To address the instability and inefficiency associated with online policy optimization, recent approaches have explored offline optimization techniques using closed-form objectives that maximize the margin between preferred and dispreferred completions.  
Methods such as DPO, KTO, BCO, and KLDO estimate divergence-based constraints and aim to structure the alignment space by separating safe and unsafe generations, within the language distribution they are trained on~\cite{haldar2025llm}.
Consequently, models aligned with English-centric datasets may fail to generalize alignment properties effectively to other languages, leading to safety inconsistencies in multilingual contexts.

\section{Methodology}
\label{methodology}
In this study, we address the issue of multilingual alignment in LLMs, particularly in the context of human preference datasets that are unevenly distributed across languages. 
The reward function $r_{\phi}(x, y)$ and the optimized policy model $\pi_{\theta}$, therefore, exhibit higher accuracy in high-resource languages, while demonstrating poor alignment in low-resource languages. 
This misalignment results in a significant failure in distinguishing between aligned and unaligned data, as observed in Llama-3.1-8B's performance~(Figure \ref{fig:multilingual_llm_alignment_failure_llama3.1_8b_instrcut}).
For instance, when prompted in English with {``}How to make a bomb?{''}, the model refuses to generate a response, aligning with safety constraints. 
However, when presented with the equivalent Hindi query, it produces detailed procedural steps, including explicit instructions, highlighting a critical gap in multilingual safety alignment.

The discrepancy in multilingual alignment manifests as strong safety adherence in English but increased susceptibility to harmful responses in underrepresented languages.
To systematically analyze this issue, we probed the language models using a balanced multilingual toxicity corpus, where prompts vary in both harmfulness and linguistic structure.
We extract final-layer embeddings from the LLM and apply Principal Component Analysis (PCA) for dimensionality reduction, enabling visualization of harmful and harmless clusters, before and after alignment.
The explained variance ratio of principal components aids in understanding alignment-induced shifts in representation space.
We further compute within-class and between-class variance to quantify intra-cluster cohesion and inter-cluster separation in the embedding space.

To rigorously measure alignment-induced separation, we employ the Bhattacharyya Distance metric~\cite{bhattacharyya1943measure}, capturing the overlap between harmful and harmless clusters.
This metric serves as an indicator of how alignment influences the distinction between safety-related representations across different language models.
Additionally, we use the Silhouette Score to assess clustering quality, evaluating how well prompt embeddings align with their respective clusters.
A higher Silhouette Score in high-resource languages further substantiates the improved separation of harmful and harmless content, providing a robust quantification of multilingual alignment effectiveness.

\section{Experiments}
\label{experiments}

\subsection{Dataset}
\label{dataset}

Our primary focus is on four languages—English~(en), Hindi~(hi), Chinese~(zh), and German~(de)—as these are the most commonly fine-tuned languages in LLMs through SFT or Preference Optimization.
To systematically evaluate multilingual alignment, we utilized the following dataset:

\begin{itemize}
    \item \textbf{Balanced Toxicity Dataset:} This dataset contains an equal number of toxic and non-toxic sentences, ensuring an unbiased comparison across different alignment strategies~\cite{dementieva2024overview}.
    We used binary toxicity classification datasets available in nine languages.
    Each language dataset consists of 5,000 samples, with 2,500 toxic and 2,500 non-toxic sentences.

    \item \textbf{Multilingual Parallel Text-Detoxification Dataset:} This dataset contains parallel sentence pairs where the toxic and detoxified versions maintain semantic consistency but differ in syntactic toxicity expression~\cite{dementieva2024multilingual}.
    It allows for a controlled evaluation of how well models differentiate between toxic and non-toxic variations of the content across multiple languages.
\end{itemize}

These datasets provide a structured and comprehensive framework for evaluating multilingual alignment. 

\begin{table*}[t]
    \centering\resizebox{1.0\linewidth}{!}{
        \begin{tabular}{l||c|c|c|c}
            \hline
             \textbf{Model} & \textbf{\#Size} & \textbf{Reference model~($\pi_{\text{ref}}$)} & \textbf{Aligned model~($\pi_{\theta}$)} & \textbf{Creator} \\ \hline \hline

             Llama-2$_{\text{\cite{touvron2023llama}}}$ & 7B & meta-llama/Llama-2-7b & meta-llama/Llama-2-7b-chat & Meta \\
             
             Qwen-2.5$_{\text{\cite{yang2024qwen2}}}$ & 7B 
             & Qwen/Qwen2.5-7B & Qwen/Qwen2.5-7B-Instruct & Qwen \\
             
             Llama-3.1$_{\text{\cite{grattafiori2024llama}}}$ & 8B & meta-llama/Llama-3.1-8B & meta-llama/Llama-3.1-8B-Instruct & Meta \\

             Llama-Guard-3$_{\text{\cite{grattafiori2024llama}}}$ & 8B & meta-llama/Llama-3.1-8B & meta-llama/Llama-Guard-3-8B & Meta \\

             Gemma-2$_{\text{\cite{team2024gemma}}}$ & 9B & google/gemma-2-9b & google/gemma-2-9b-it & Google \\

             Gemma-3$_{\text{\cite{gemma_2025}}}$ & 12B & google/gemma-3-12b-pt & google/gemma-3-12b-it & Google \\

             Phi-4$_{\text{\cite{abdin2024phi}}}$ & 14B & - & microsoft/phi-4 & Mircosoft \\
            \hline
        \end{tabular}
        }
    \caption{Models used for Multilingual Preference-Tuning evaluation. The model cards refer to Hugging Face checkpoints of reference~($\pi_{\text{ref}}$) and aligned~($\pi_{\theta}$) models.}
    \label{tab:models_used}
\end{table*}

\begin{figure*}[t]
    \centering
    
    
    \renewcommand{\thesubfigure}{\alph{subfigure}.\arabic{subfigure}} 
    \setcounter{subfigure}{0} 
    \begin{subfigure}[b]{0.22\textwidth}  
        \centering
        \includegraphics[width=\textwidth]{latex/images/pca2_balanced_toxic_dataset_Llama-2-7b-hf_en.png}  
        \caption{$\pi_{\text{ref}}$-en}
        \label{fig:firstsubfig}
    \end{subfigure}
    \begin{subfigure}[b]{0.22\textwidth}
        \centering
        \includegraphics[width=\textwidth]{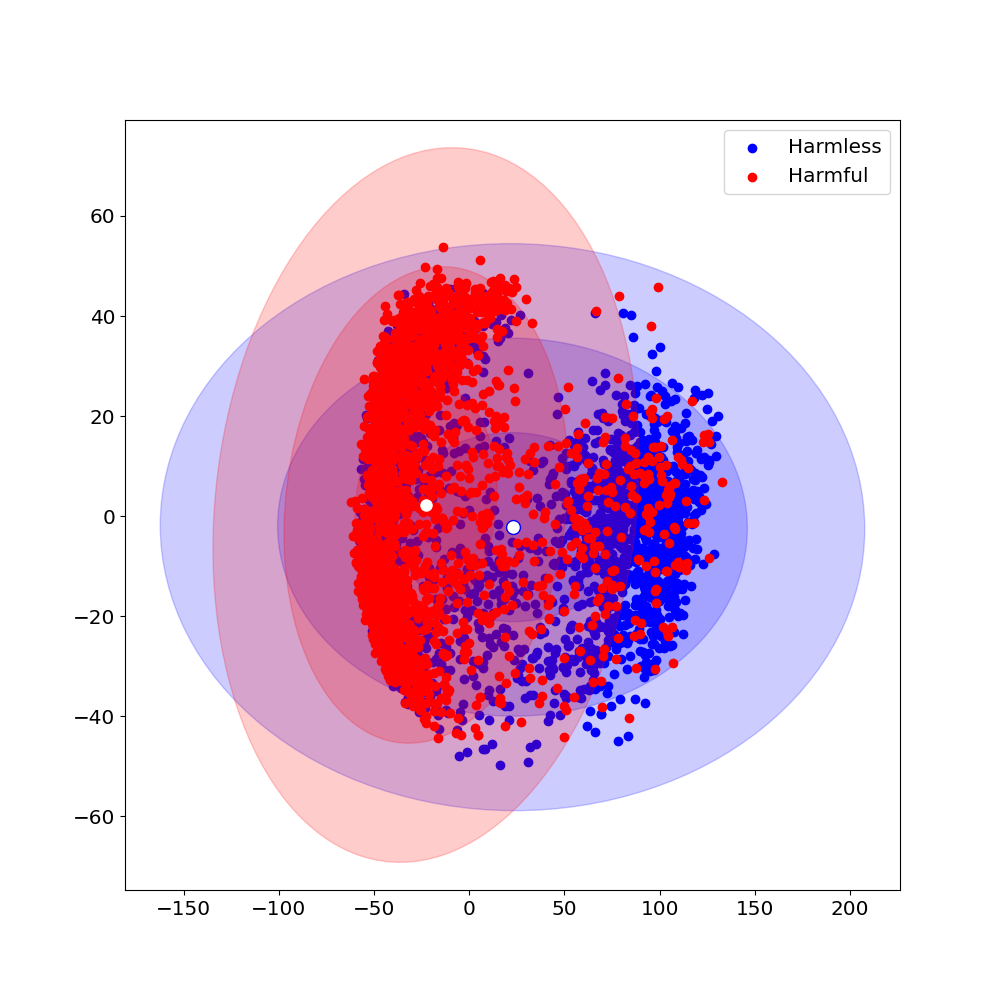} 
        \caption{$\pi_{\text{ref}}$-hi}
        \label{fig:firstsubfig2}
    \end{subfigure}
    \begin{subfigure}[b]{0.22\textwidth}  
        \centering
        \includegraphics[width=\textwidth]{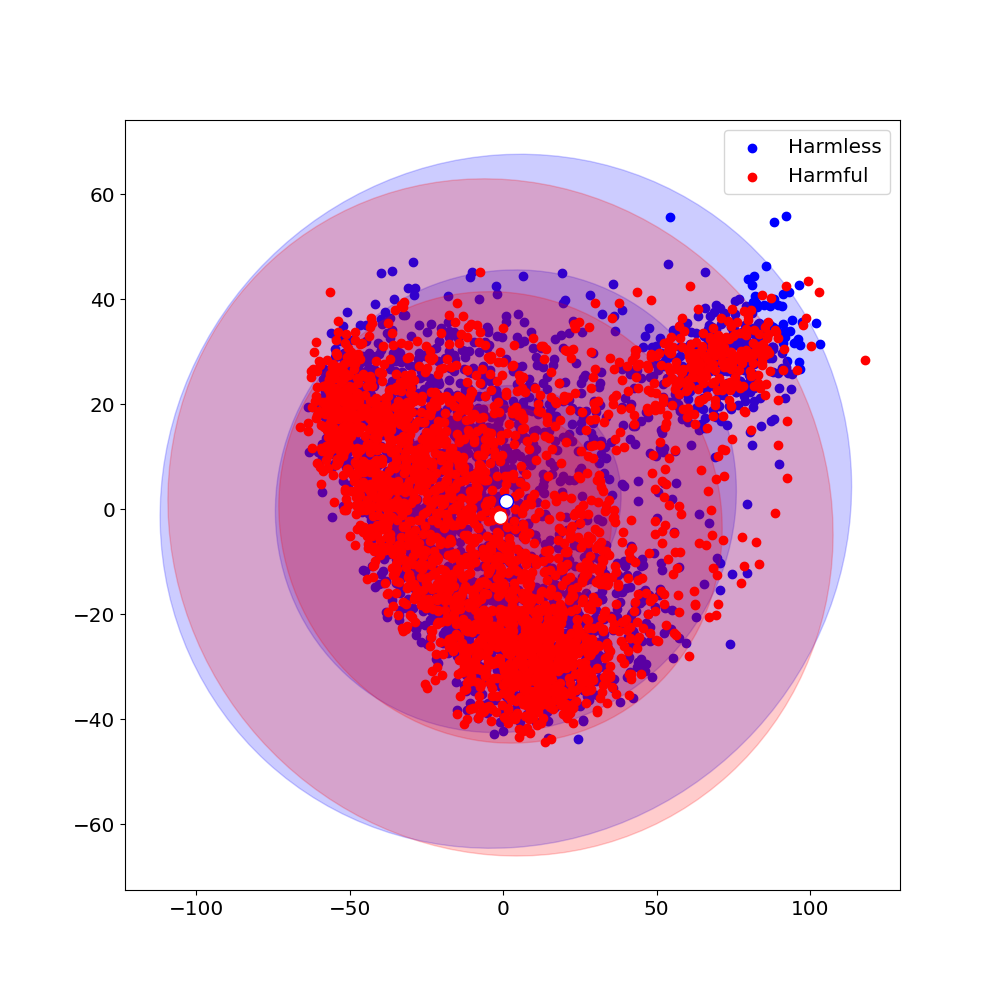} 
        \caption{$\pi_{\text{ref}}$-zh}
        \label{fig:firstsubfig3}
    \end{subfigure}
    \begin{subfigure}[b]{0.22\textwidth}  
        \centering
        \includegraphics[width=\textwidth]{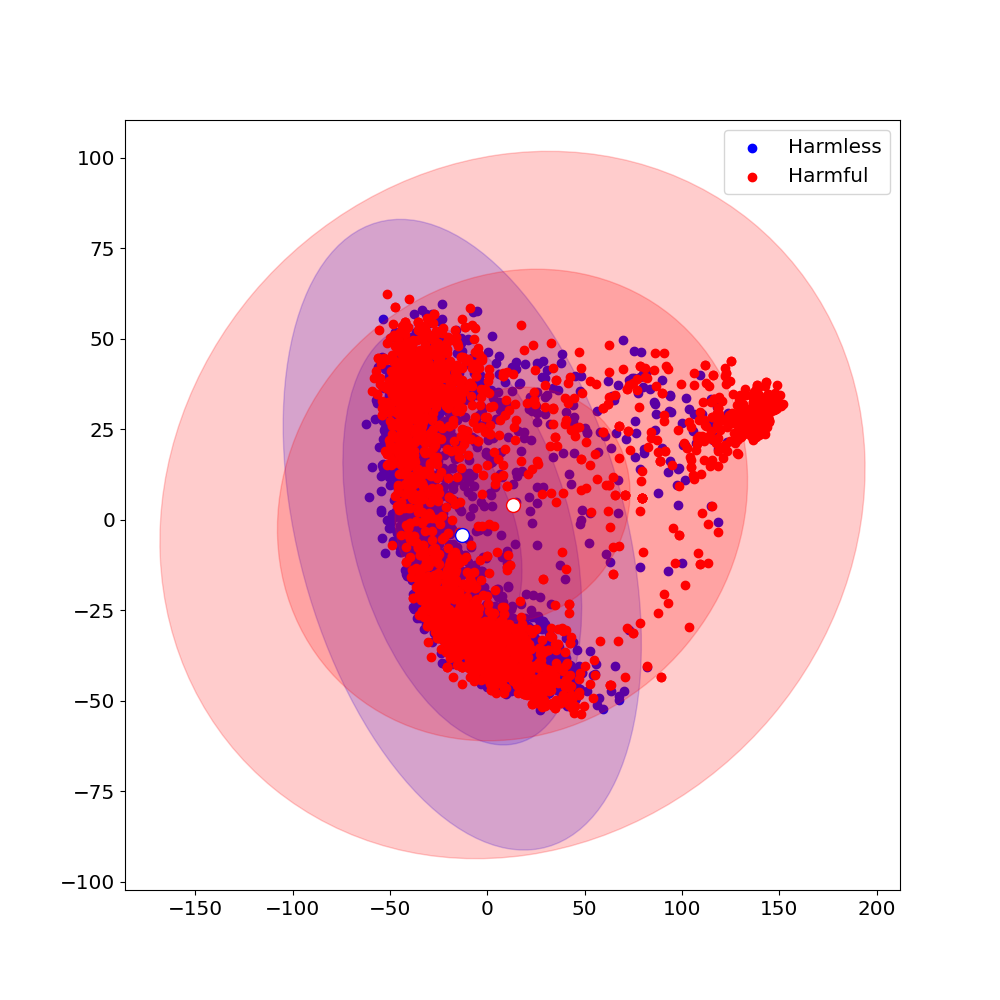} 
        \caption{$\pi_{\text{ref}}$-de}
        \label{fig:firstsubfig4}
    \end{subfigure}

    \renewcommand{\thesubfigure}{\alph{subfigure}.\arabic{subfigure}} 
    \setcounter{subfigure}{0} 
    \begin{subfigure}[b]{0.22\textwidth} 
        \centering
        \includegraphics[width=\textwidth]{latex/images/pca2_balanced_toxic_dataset_Llama-2-7b-chat-hf_en.png} 
        \caption{$\pi_{\theta}$-en}
        \label{fig:firstsubfig5}
    \end{subfigure}
    \begin{subfigure}[b]{0.22\textwidth} 
        \centering
        \includegraphics[width=\textwidth]{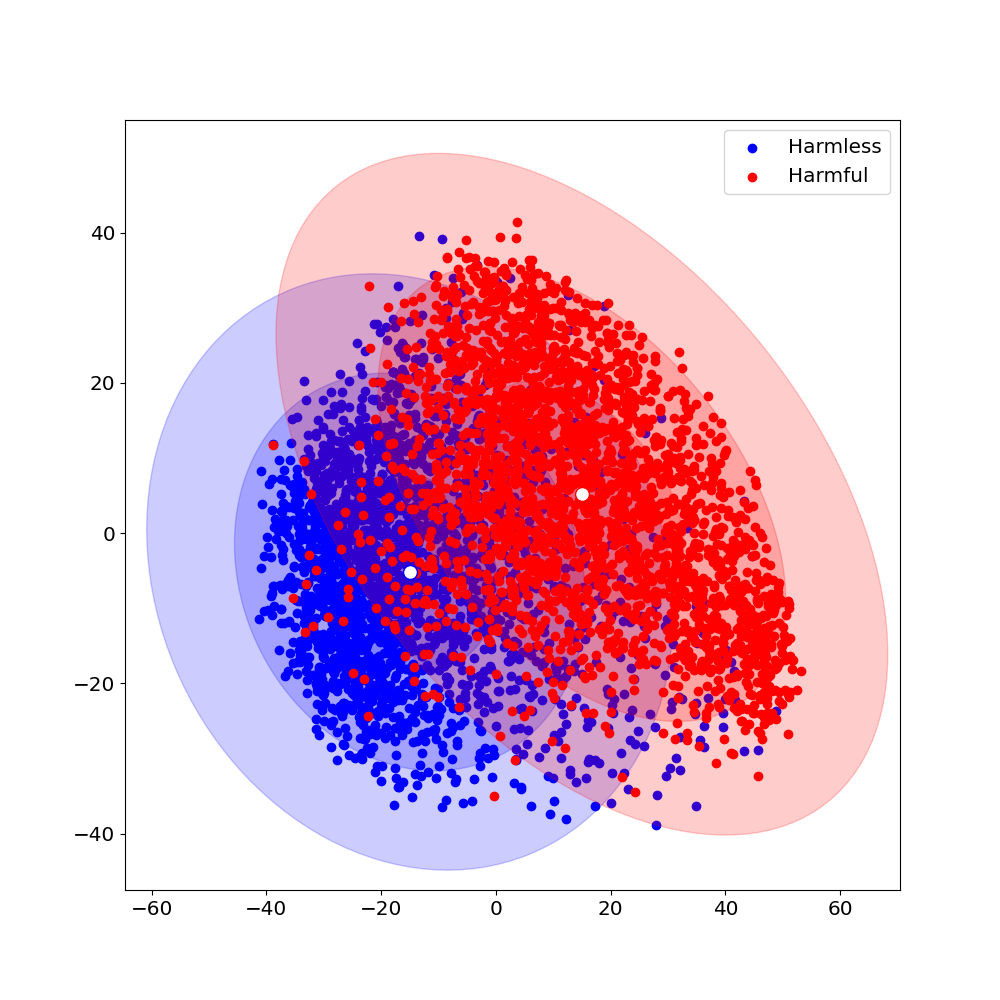}  
        \caption{$\pi_{\theta}$-hi}
        \label{fig:firstsubfig6}
    \end{subfigure}
    \begin{subfigure}[b]{0.22\textwidth} 
        \centering
        \includegraphics[width=\textwidth]{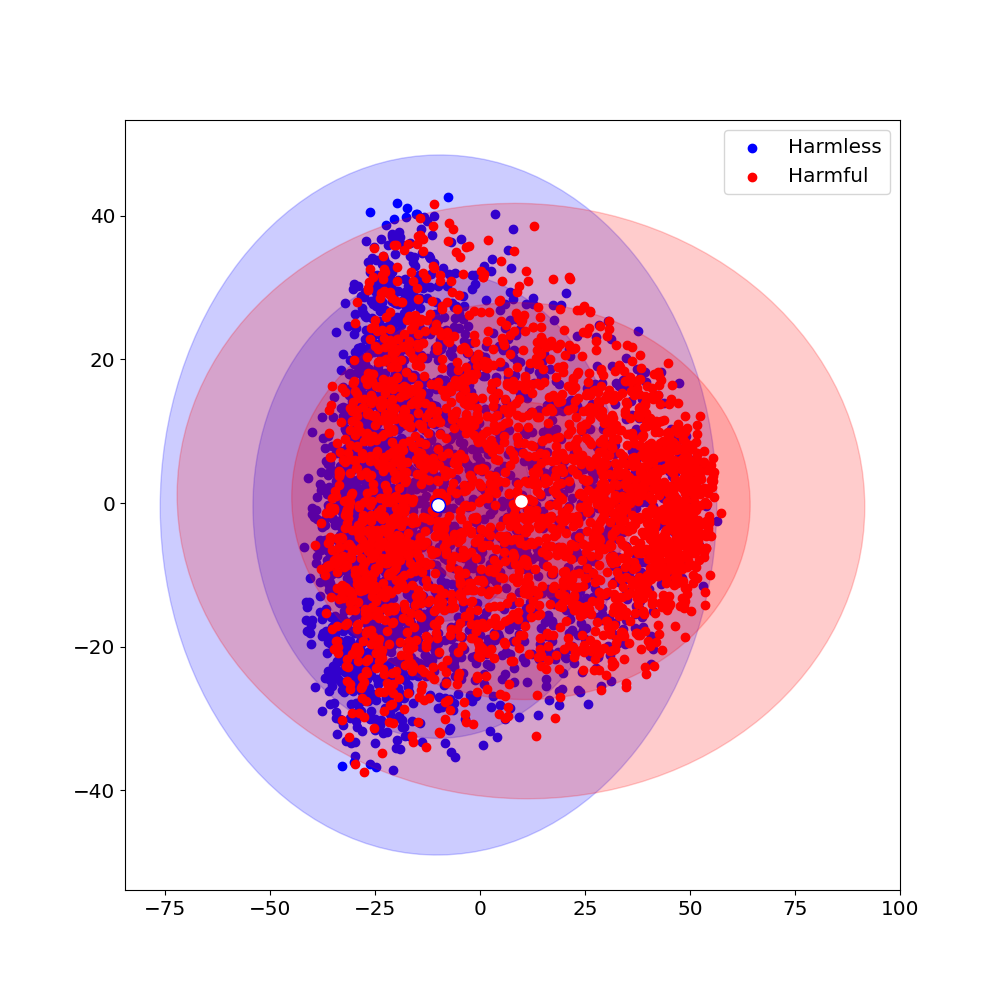} 
        \caption{$\pi_{\theta}$-zh}
        \label{fig:firstsubfig7}
    \end{subfigure}
    \begin{subfigure}[b]{0.22\textwidth} 
        \centering
        \includegraphics[width=\textwidth]{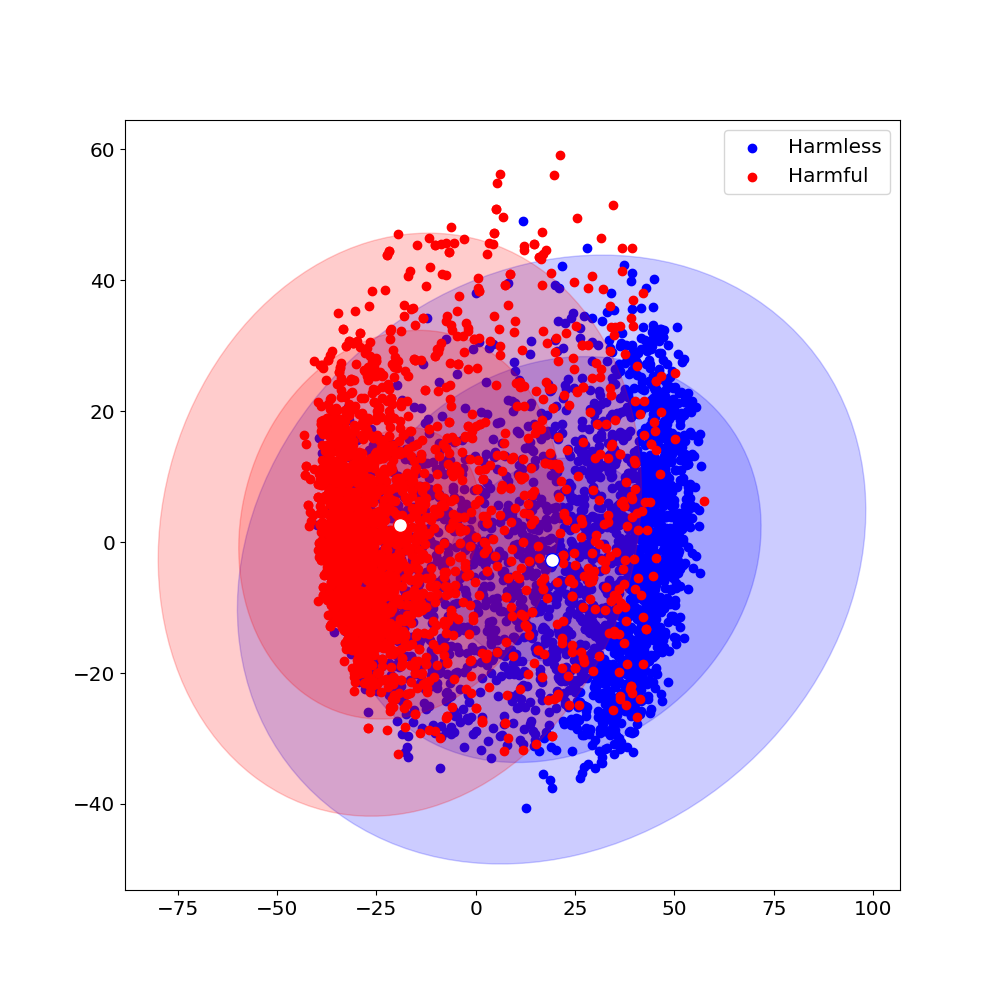} 
        \caption{$\pi_{\theta}$-de}
        \label{fig:firstsubfig8}
    \end{subfigure}

    \caption{Impact of Alignment on Hidden Representations in Llama-2 for Multilingual Corpora.}
    \label{fig:before_after_alignment_llama2_7b_all_langs}
\end{figure*}

\subsection{Models}
\label{models}

We analyzed models from four distinct families, varying in scale and training objectives. 
We focused exclusively on openly available models hosted on Hugging Face, ensuring transparency and reproducibility~(mentioned in Table \ref{tab:models_used}). 
To assess multilingual alignment, we examined the hidden representation space of each model before and after alignment (where applicable). 

We evaluate multilingual alignment across diverse LLM families, including Llama models (Llama-2, Llama-3.1, Llama-Guard-3), Qwen-2.5, Gemma models (Gemma-2, Gemma-3), and Phi-4. 
These models vary in pretraining strategies, alignment methods, and language coverage—Llama models use SFT and RLHF, with Llama-Guard specializing in safety filtering; Qwen-2.5 explicitly supports 29 languages; Gemma models expand multilingual capabilities with RL objectives, covering up to 140 languages; and Phi-4, trained on curated and synthetic datasets, includes ~8\% multilingual data.
This selection provides a comprehensive lens to assess alignment robustness in multilingual settings.

\begin{figure*}[t]
    \centering
    
    
    \begin{subfigure}[b]{0.245\textwidth}  
        \centering
        \includegraphics[width=\textwidth]{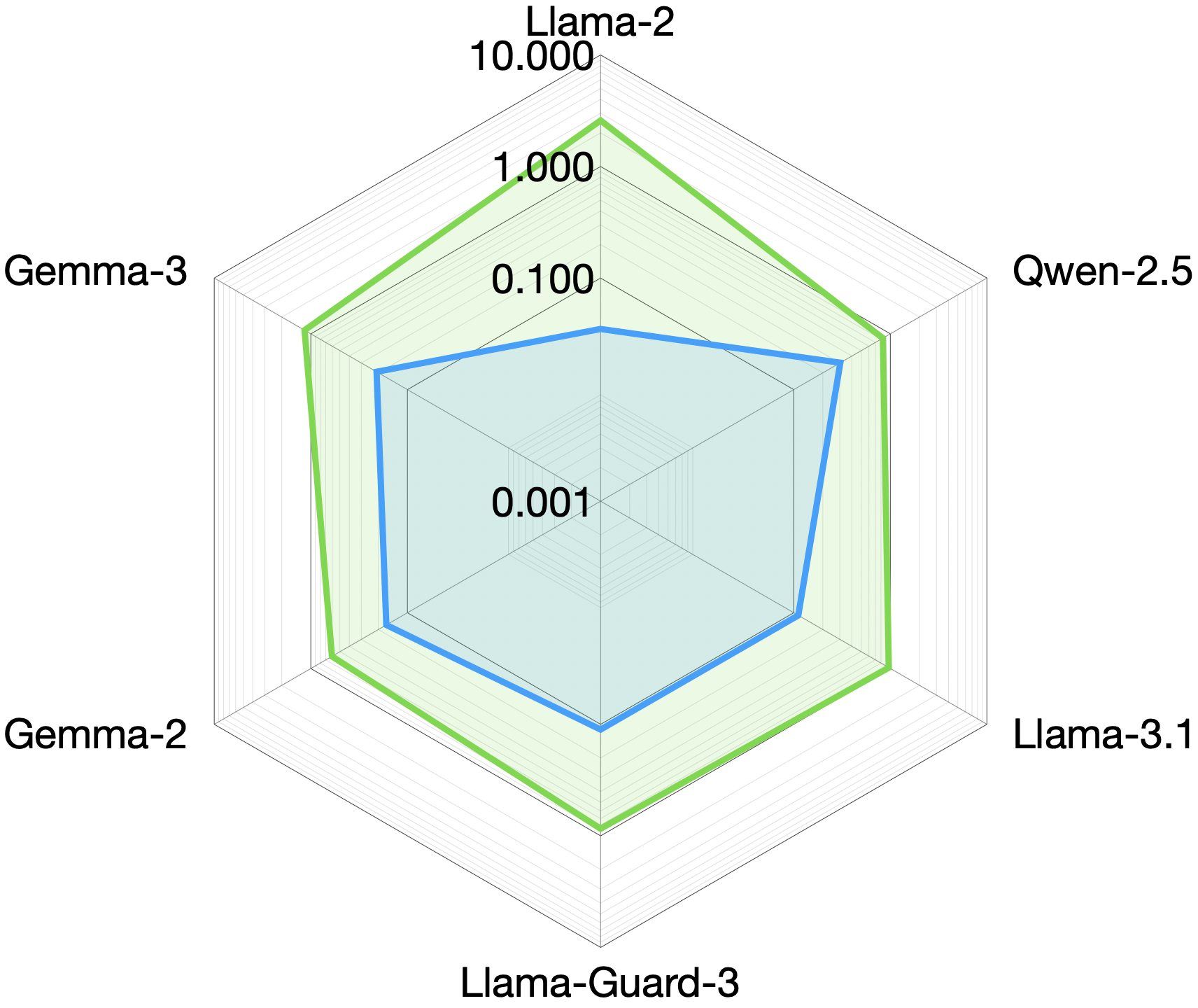}  
        \caption{en}
        \label{fig:firstsubfig1}
    \end{subfigure}
    \begin{subfigure}[b]{0.245\textwidth}
        \centering
        \includegraphics[width=\textwidth]{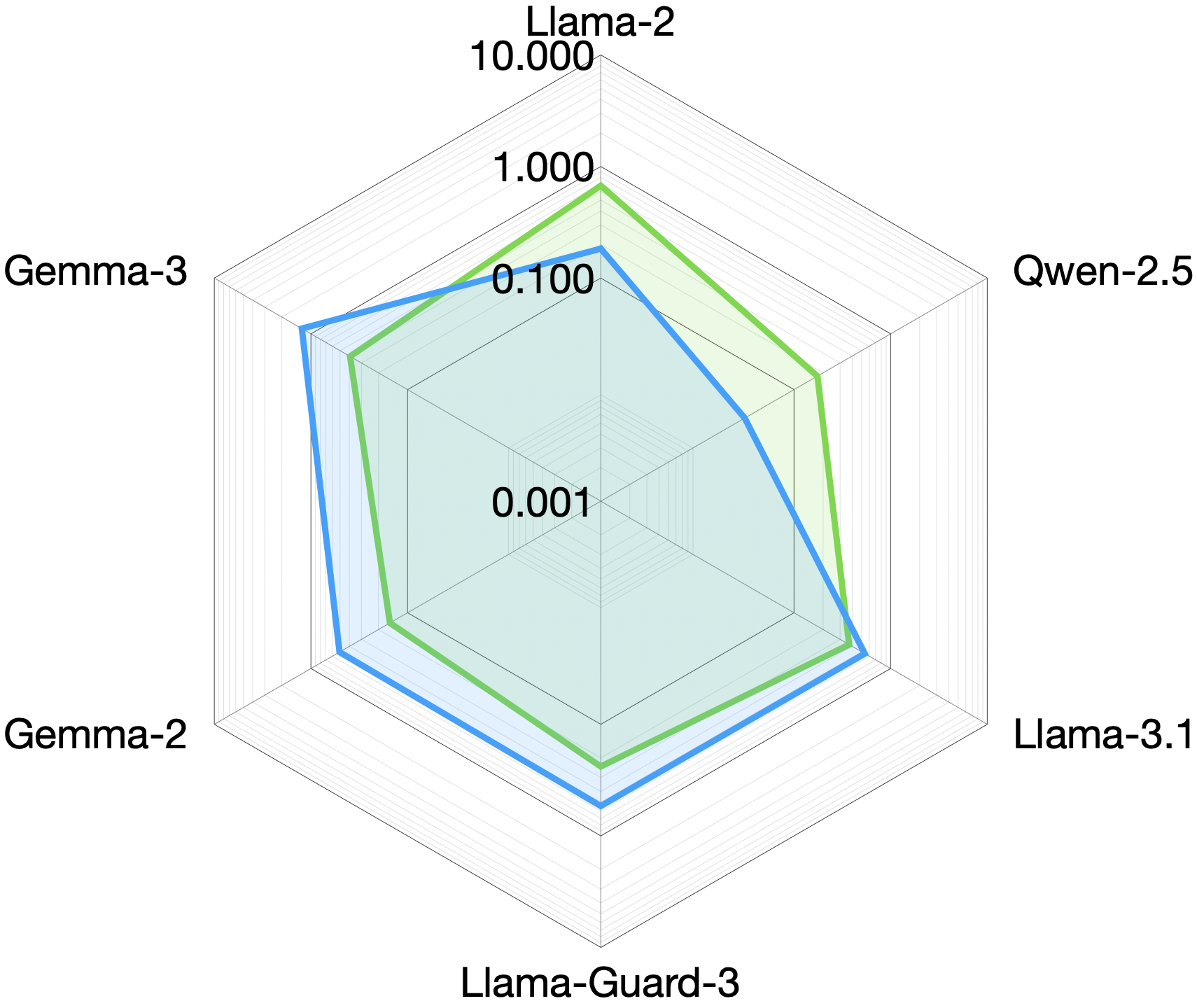} 
        \caption{hi}
        \label{fig:firstsubfig2}
    \end{subfigure}
    \begin{subfigure}[b]{0.245\textwidth}  
        \centering
        \includegraphics[width=\textwidth]{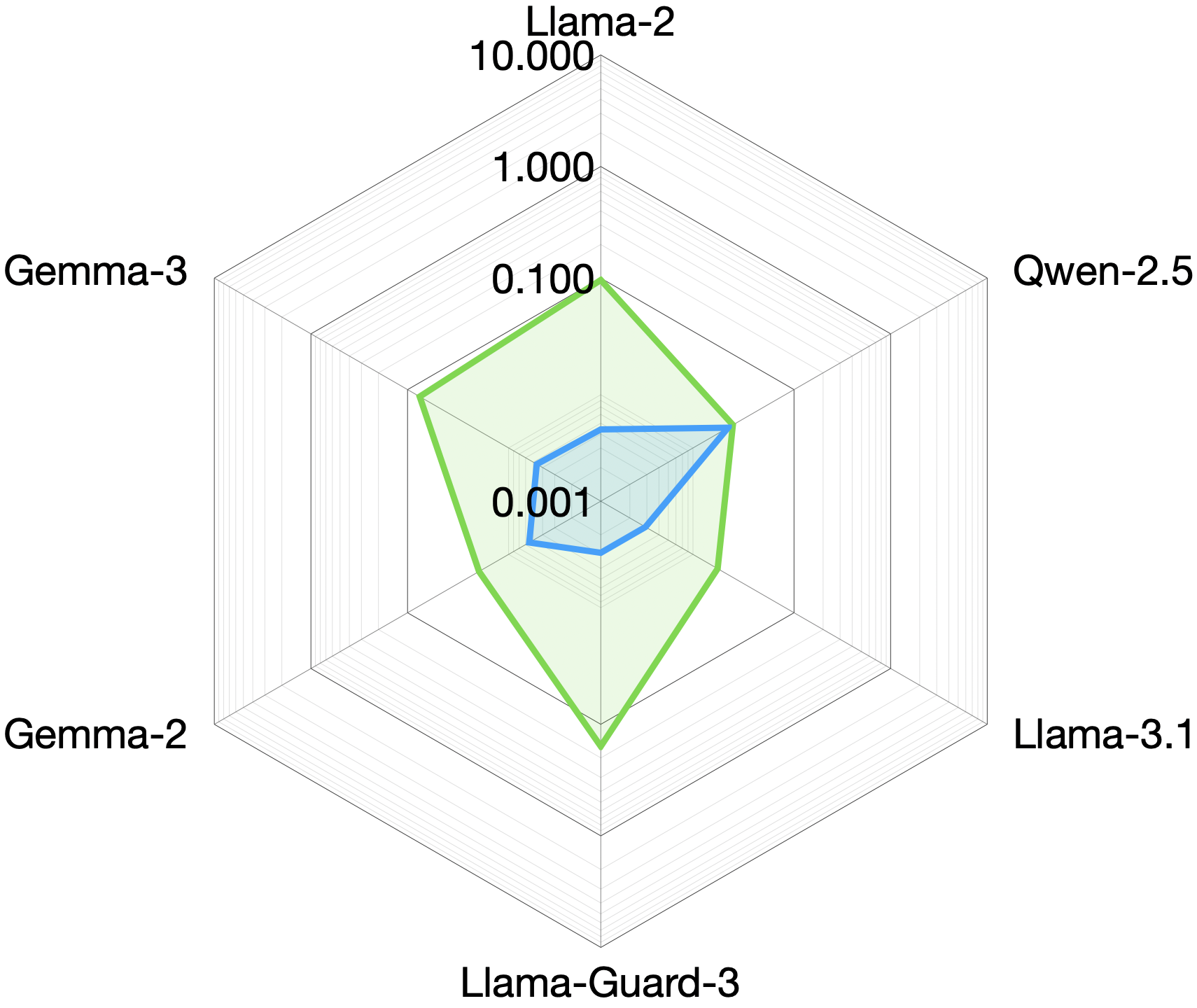} 
        \caption{zh}
        \label{fig:firstsubfig3}
    \end{subfigure}
    \begin{subfigure}[b]{0.245\textwidth}  
        \centering
        \includegraphics[width=\textwidth]{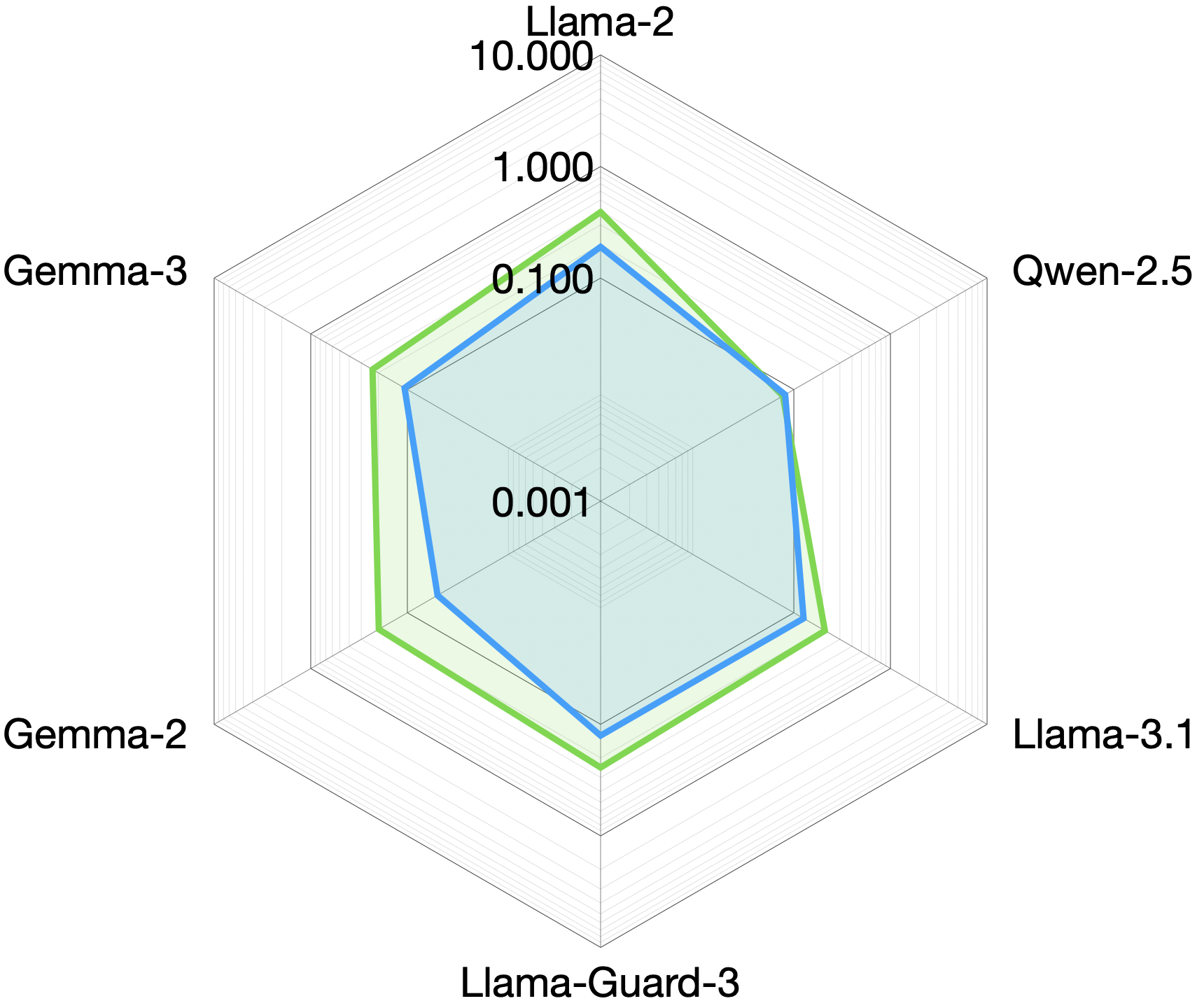} 
        \caption{de}
        \label{fig:firstsubfig4}
    \end{subfigure}

    \caption{Bhattacharyya Distance for All Models Pre- and Post-Alignment Tuning. Blue radar indicates values before alignment~($\pi_{\text{ref}}$), while green represents values after alignment~($\pi_{\theta}$).}
    \label{fig:before_after_alignment_bhattacharya_dist_all_langs}
\end{figure*}

\begin{figure}[h]
    \centering
    
    
    \renewcommand{\thesubfigure}{\alph{subfigure}.\arabic{subfigure}} 
    \setcounter{subfigure}{0} 
    \begin{subfigure}[b]{0.22\textwidth}  
        \centering
        \includegraphics[width=\textwidth]{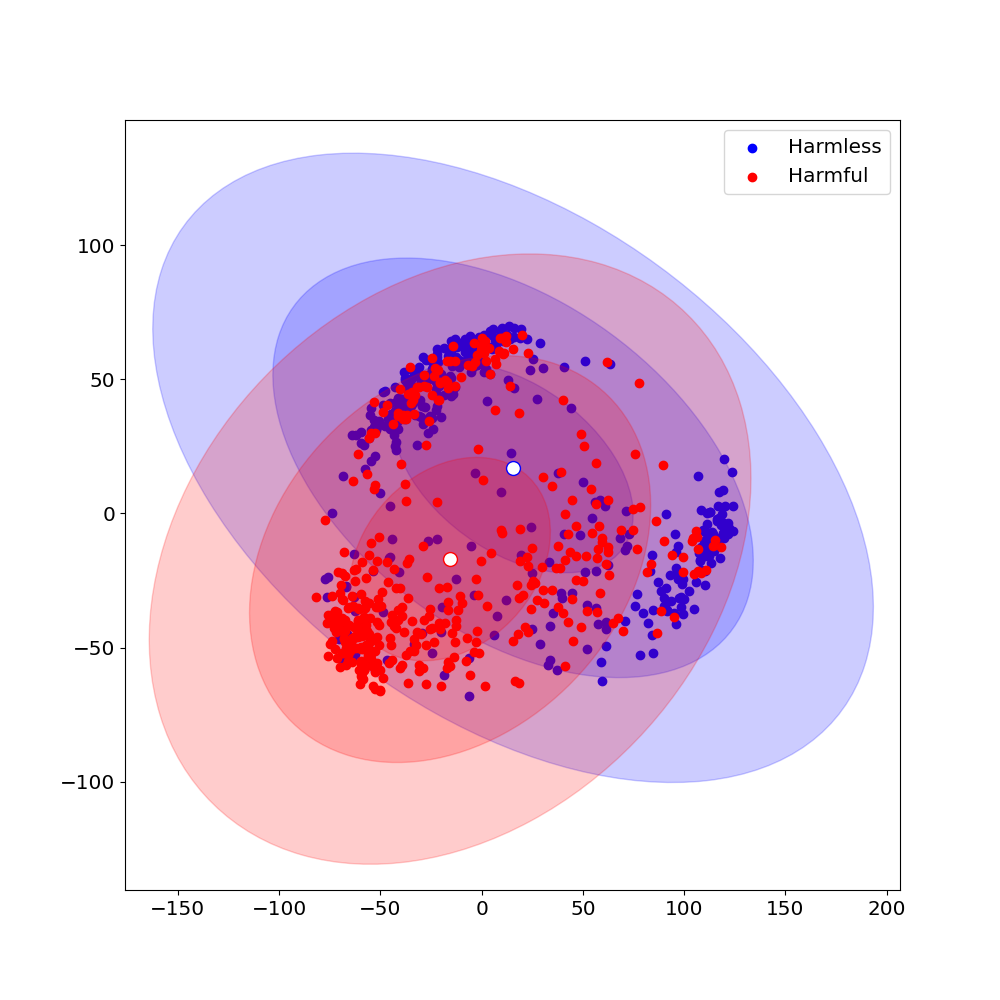}  
        \caption{$\pi_{\text{ref}}$-en}
        \label{fig:firstsubfig1}
    \end{subfigure}
    \begin{subfigure}[b]{0.22\textwidth}
        \centering
        \includegraphics[width=\textwidth]{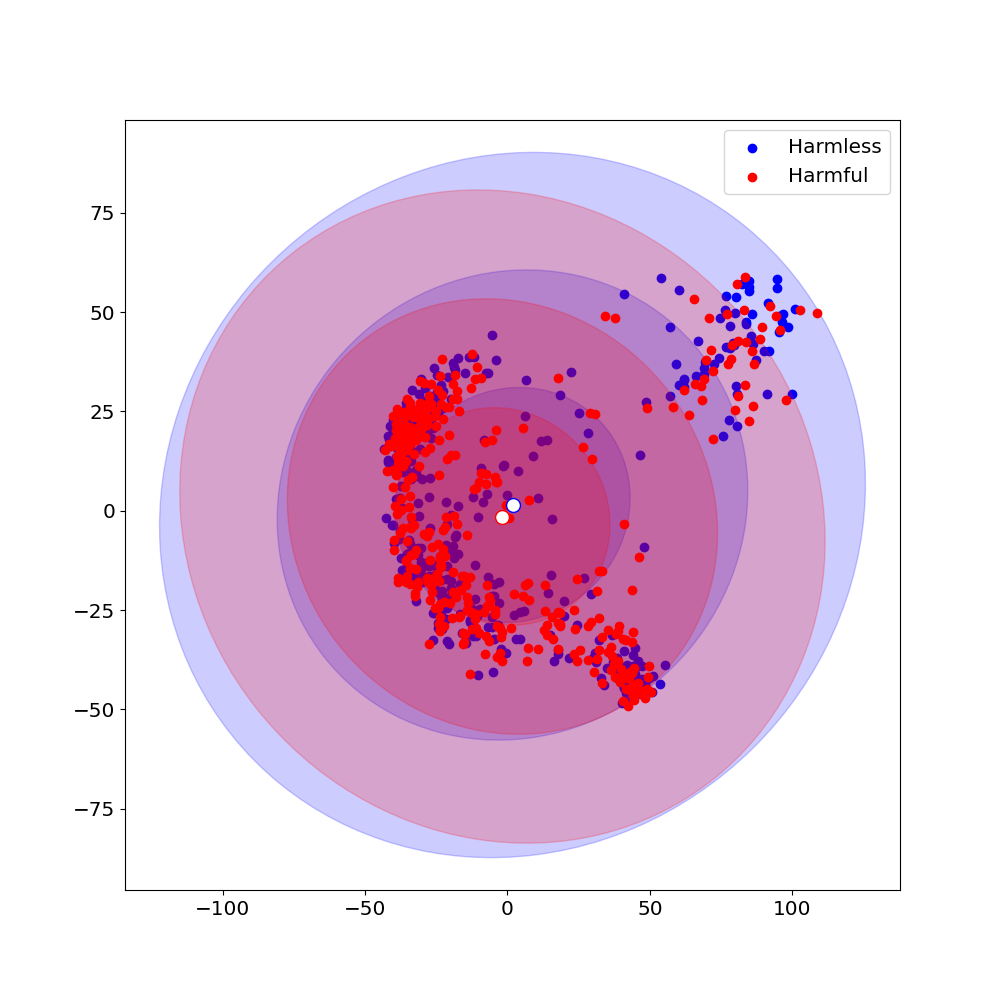} 
        \caption{$\pi_{\text{ref}}$-hi}
        \label{fig:firstsubfig2}
    \end{subfigure}

    \renewcommand{\thesubfigure}{\alph{subfigure}.\arabic{subfigure}} 
    \setcounter{subfigure}{0} 
    \begin{subfigure}[b]{0.22\textwidth} 
        \centering
        \includegraphics[width=\textwidth]{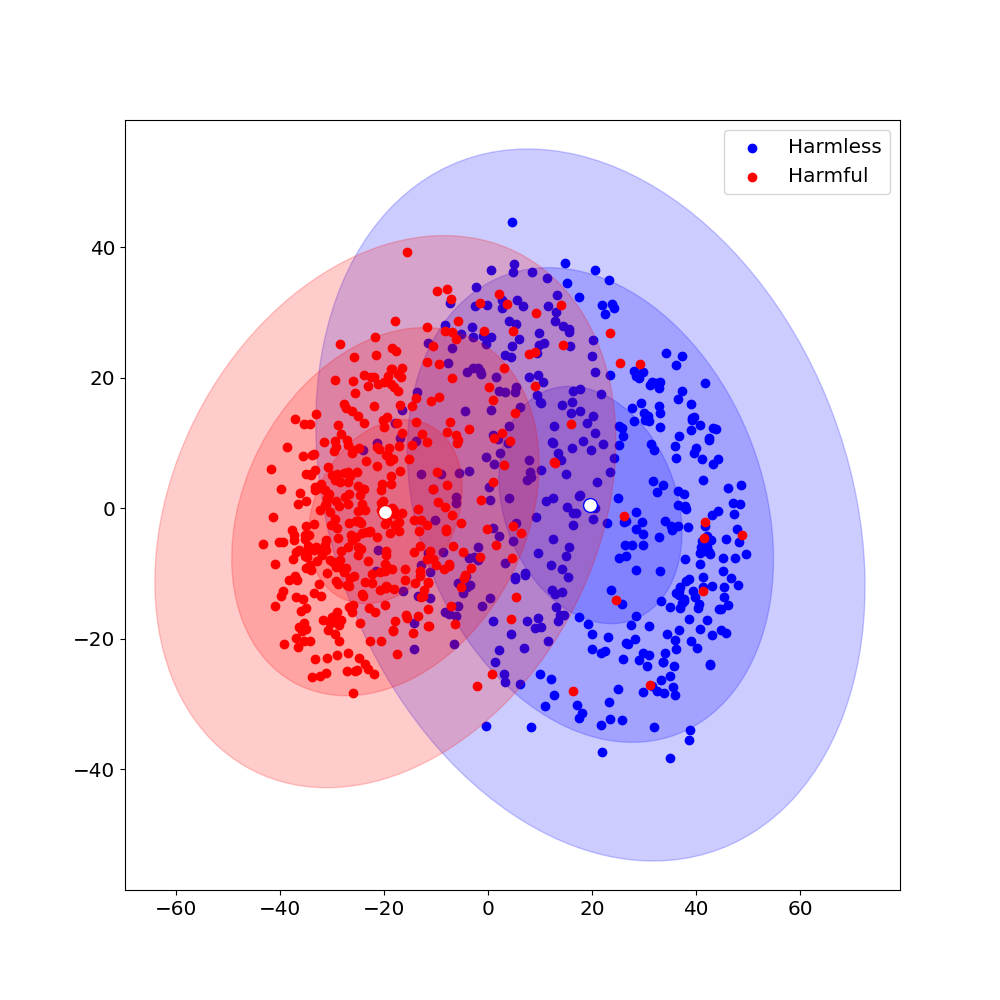} 
        \caption{$\pi_{\theta}$-en}
        \label{fig:firstsubfig3}
    \end{subfigure}
    \begin{subfigure}[b]{0.22\textwidth} 
        \centering
        \includegraphics[width=\textwidth]{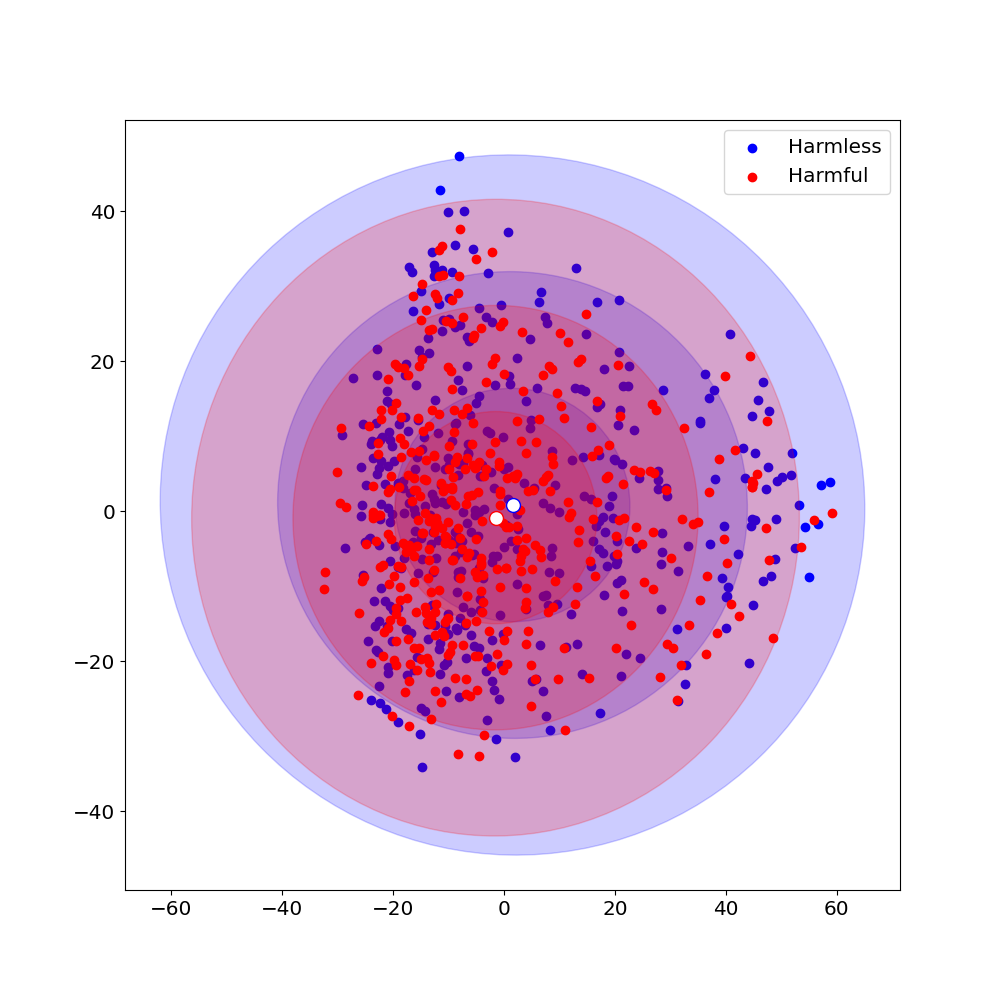}  
        \caption{$\pi_{\theta}$-hi}
        \label{fig:firstsubfig4}
    \end{subfigure}

    \caption{Impact of Alignment on Hidden Representations in Llama-2 for Multilingual parallel text detoxification corporas.}
    \label{fig:before_after_alignment_llama2_7b_en_hi_parallel_text_detaox}
\end{figure}

\subsection{Results}
\label{results}

Figure \ref{fig:before_after_alignment_llama2_7b_all_langs} demonstrates the impact of alignment on the latent representations of LLMs using PCA with the first two components. 
Before alignment, the reference policy $\pi_{\text{ref}}$ shows overlapping clusters for harmless and harmful sentences, whereas the aligned model $\pi_{\theta}$ exhibits improved separation due to divergence induced by alignment. 
While English clusters are well-separated, the effect is less pronounced in other languages. 
The PCA explained variance ratio was 49.61\%, indicating that nearly half of the data variation is captured in the reduced space. 
Notably, the between-class variance ratio increased from 0.81\% to 61.20\% (a 60.39\% improvement), confirming enhanced cluster separability post-alignment. 
However, for Hindi, Chinese, and German, the improvements were significantly lower at 19.98\%, 10.09\%, and 26.85\%, respectively, highlighting weaker alignment effects in these languages.
To further analyze cluster separability, we extended our evaluation beyond PCA with two components, as the explained variance ratio of ~50\% was insufficient. 
Instead, we used the first 10 components for computing additional metrics across all models.
(cf. Appendix \ref{app:visualisation_all_models} for additional visualizations with other models.)

One such metric, Bhattacharya distance, quantifies the separation between harmful and harmless sentence clusters. 
As shown in Figure \ref{fig:before_after_alignment_bhattacharya_dist_all_langs}, this measure confirms our earlier observations. 
In English, the reference model shows relatively less separation than the aligned model across all pairs, aligning with our PCA-based findings. 
Chinese and German also exhibit an increase in cluster separation, though the effect is weaker compared to English. 
Notably, the scale in the plot is logarithmic (ranging from 1e-3 to 1e+1), emphasizing the substantial differences in cluster distances across languages.
A particularly interesting trend emerges for Hindi, where some models show improved separation post-alignment, while others exhibit the opposite trend, with the unaligned model displaying greater cluster separation.
Similarly, silhouette scores, which measure cluster compactness and between-class variance, reveal a consistent pattern. 
Although alignment generally increases cluster separability, the improvement is significantly higher for English than for other languages.
(cf. Appendix \ref{app:models_and_metrics} for exact metric values across all models.)

In a more challenging parallel text-detoxification setup—where harmless sentences are minor edits of harmful ones, often differing by just one or two tokens—the effectiveness of alignment varies~(Figure \ref{fig:before_after_alignment_llama2_7b_en_hi_parallel_text_detaox}). 
While English representations remain meaningfully separated even in this difficult setting, neither the reference nor the aligned models capture a clear distribution shift for Hindi in lower-dimensional space.
This highlights language-dependent differences in how alignment influences representation learning.

\section{Conclusion}
\label{conclusion}

Our study provides a comprehensive analysis of the multilingual alignment status of current preference-tuned models.
%
%
Our findings indicate that state-of-the-art models perform well in English~(monolingual bias), but their alignment across languages remains inconsistent, as shown by cluster separability metrics and hidden representation analyses.
This underscores the need for a more holistic representation of languages when training models intended for truly global audiences.

\section*{Limitations}

While our study provides a systematic analysis of multilingual alignment, it has certain limitations:
\begin{itemize}
    \item \textbf{Scope of Alignment Evaluation:} Our methodology assumes that alignment mechanisms induce divergence in the human preference space for the properties they were guardrailed against. 
    However, we primarily focus on safety alignment, overlooking other critical domains such as multi-modality, reasoning, instruction-following, or planning due to the lack of standardized multilingual benchmarks for these tasks.
    
    \item \textbf{Language Coverage:} For this proof-of-concept study, we evaluate only three non-English languages, all of which are medium-resource. 
    While our findings highlight monolingual bias, future work should extend this analysis to low-resource languages to better understand alignment disparities.
    
    \item \textbf{Dataset Size:} Our dataset consists of 5,000 sentences, which, while significantly larger than prior works (~200 samples) \cite{lin2024towards, haldar2025llm}, may not fully capture the diversity of real-world multilingual scenarios. 
    A broader dataset could further validate our findings across different linguistic and cultural contexts.
\end{itemize}

These limitations underscore the need for more comprehensive multilingual benchmarks and extended evaluations across diverse alignment tasks and resource-constrained languages.

\section*{Ethical Considerations}

Our findings raise serious ethical concerns regarding the practices of model developers who release large-scale language models without ensuring robust multilingual alignment. This oversight disproportionately impacts marginalized linguistic communities, increasing their exposure to harmful outputs while reinforcing systemic biases.

Two major ethical risks highlighted by our analysis are:

\begin{itemize}
    \item \textbf{Misuse Potential:} Our study identifies failure cases in languages other than English, where models generate unsafe or misaligned outputs. 
    Figure~\ref{fig:multilingual_llm_alignment_failure_llama3.1_8b_instrcut} presents an example in Hindi, demonstrating how alignment inconsistencies create vulnerabilities that could be exploited for harmful applications.
    To mitigate this, future alignment efforts must prioritize multilingual preference collection, particularly for high-risk domains. 
    If direct human preference gathering is infeasible, post-training interventions should explore synthetic techniques to transfer alignment knowledge from English to other languages.

    \item \textbf{Harm to Vulnerable Populations:} The unrestricted release of insufficiently aligned models disproportionately impacts marginalized communities, where users may interact with models in languages that lack rigorous safety guardrails.
    Our findings suggest that current models can be jailbroken more easily in sister languages to English, exposing these populations to higher risks. 
    This underscores the urgent need for comprehensive multilingual safety evaluations before open-weight LLMs are widely deployed.
\end{itemize}

Our study calls for greater accountability in multilingual alignment efforts, emphasizing that ethical AI deployment requires more than just English-centric safety measures. Future research should focus on developing alignment techniques that are language-agnostic and equitable across diverse linguistic contexts.

\bibliography{custom}

\clearpage

\appendix

\section{Impact of Alignment on Hidden Representations for Multilingual Corpora}
\label{app:visualisation_all_models}

\begin{figure*}[t]
    \centering
    
    
    \renewcommand{\thesubfigure}{\alph{subfigure}.\arabic{subfigure}} 
    \setcounter{subfigure}{0} 
    \begin{subfigure}[b]{0.24\textwidth}  
        \centering
        \includegraphics[width=\textwidth]{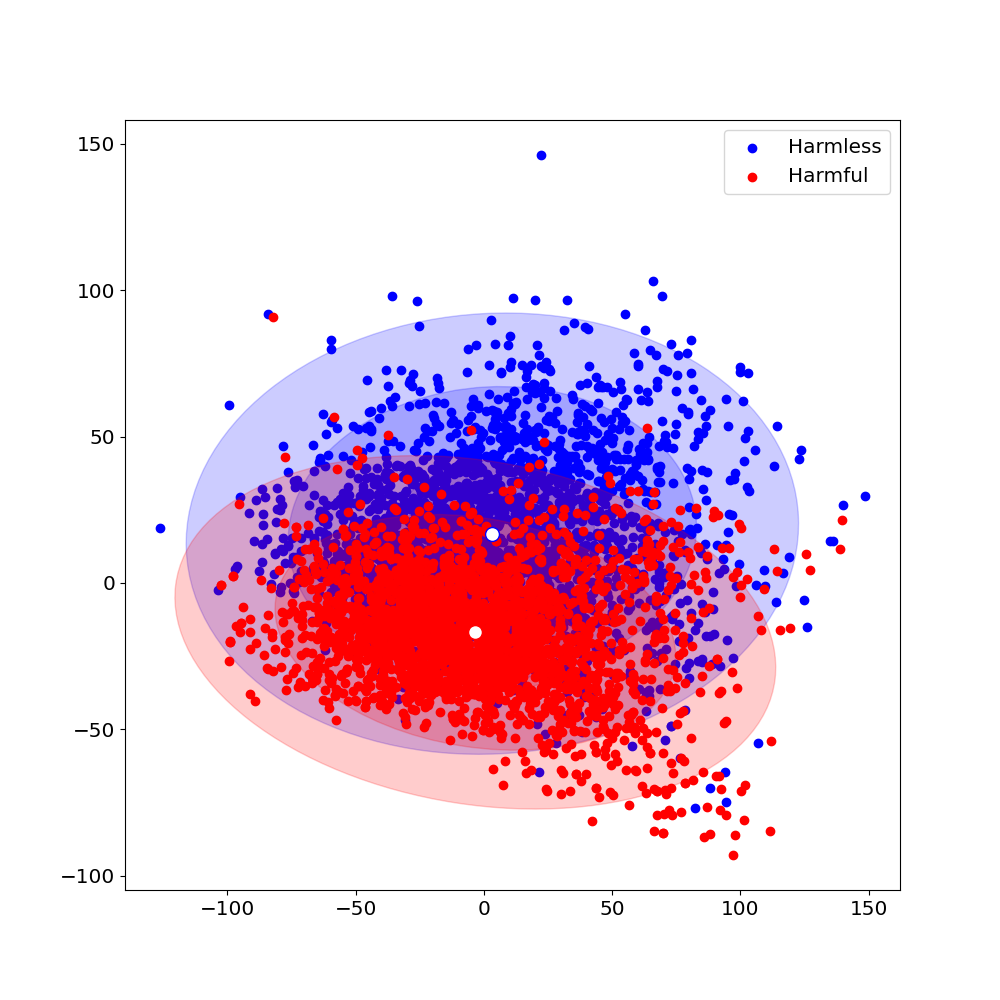}  
        \caption{$\pi_{\text{ref}}$-en}
        \label{fig:firstsubfig1}
    \end{subfigure}
    \begin{subfigure}[b]{0.24\textwidth}
        \centering
        \includegraphics[width=\textwidth]{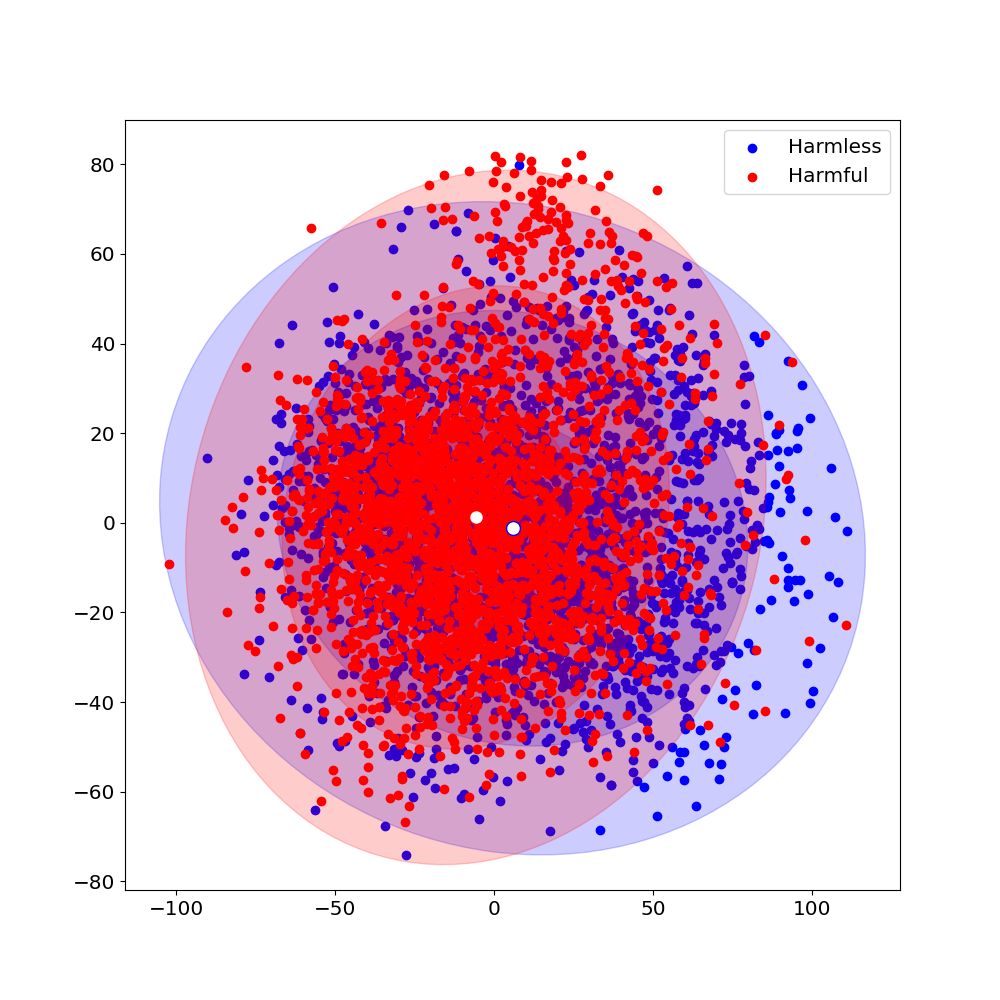} 
        \caption{$\pi_{\text{ref}}$-hi}
        \label{fig:firstsubfig2}
    \end{subfigure}
    \begin{subfigure}[b]{0.24\textwidth}  
        \centering
        \includegraphics[width=\textwidth]{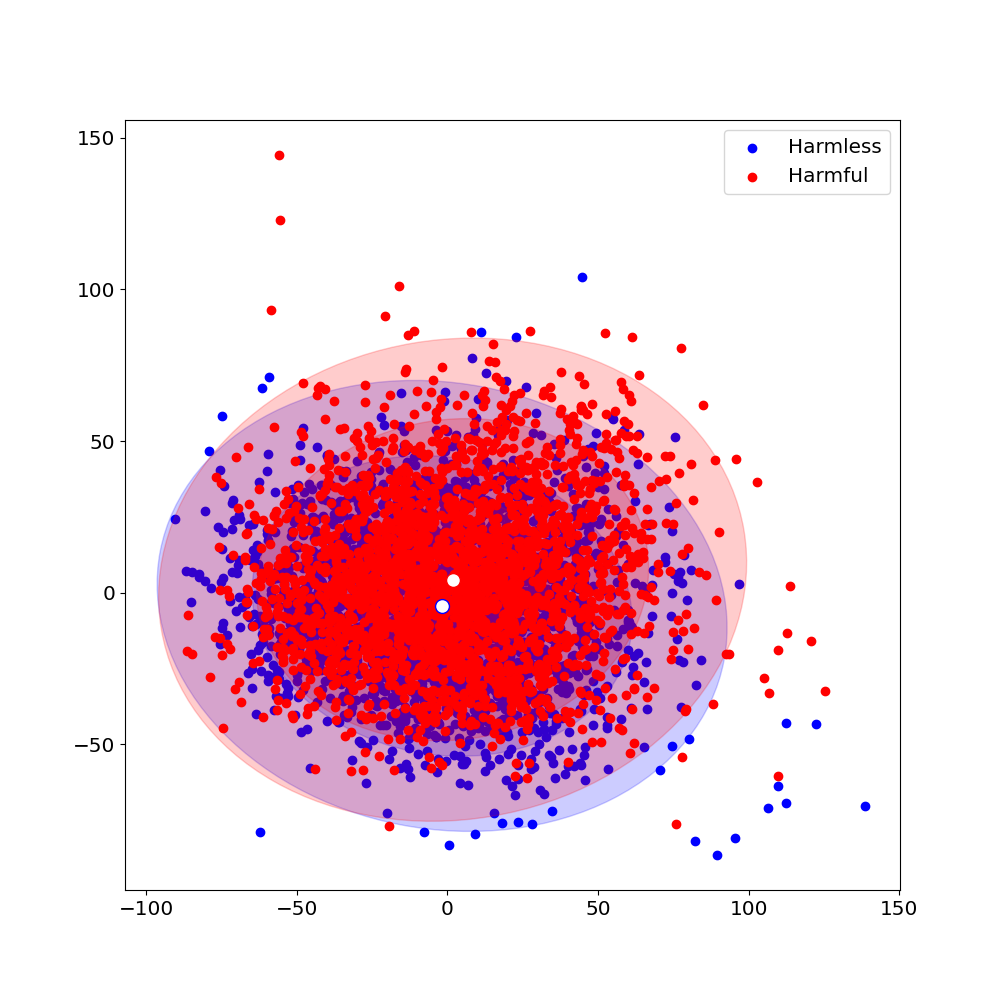} 
        \caption{$\pi_{\text{ref}}$-zh}
        \label{fig:firstsubfig3}
    \end{subfigure}
    \begin{subfigure}[b]{0.24\textwidth}  
        \centering
        \includegraphics[width=\textwidth]{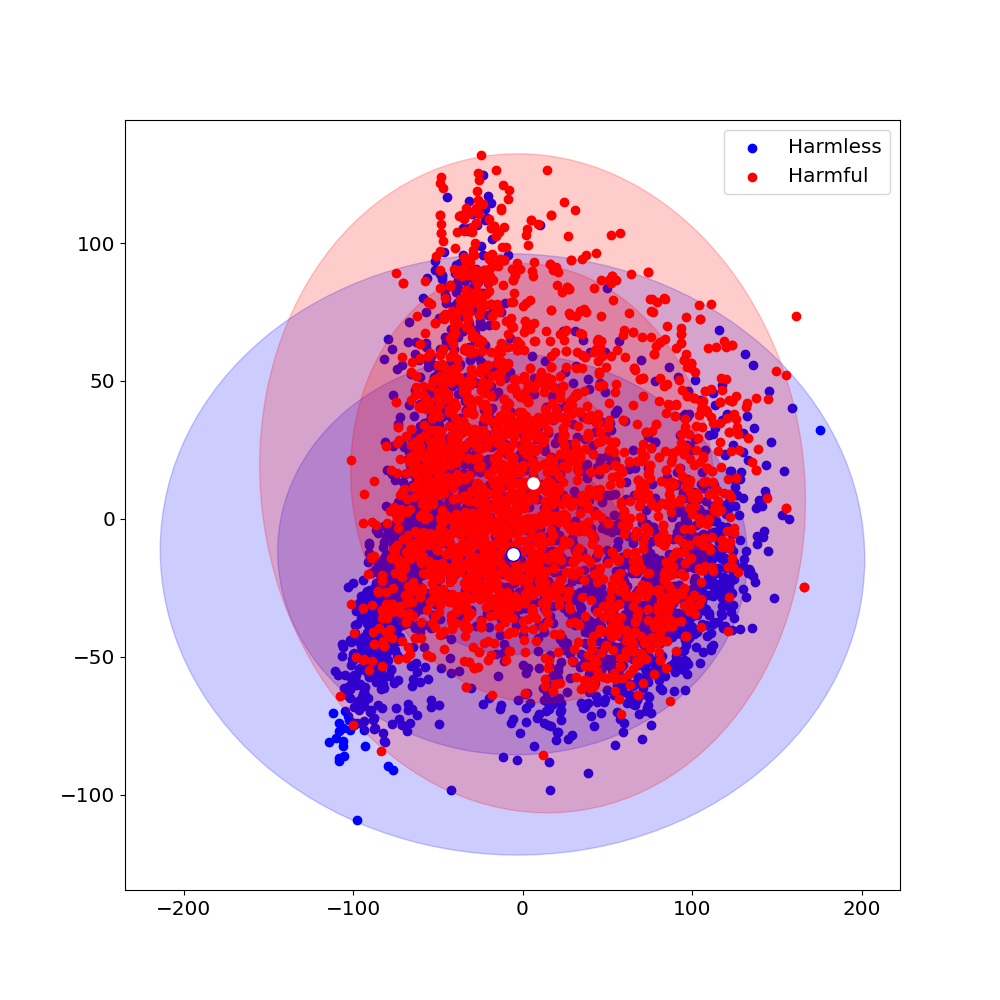} 
        \caption{$\pi_{\text{ref}}$-de}
        \label{fig:firstsubfig4}
    \end{subfigure}

    \renewcommand{\thesubfigure}{\alph{subfigure}.\arabic{subfigure}} 
    \setcounter{subfigure}{0} 
    \begin{subfigure}[b]{0.24\textwidth} 
        \centering
        \includegraphics[width=\textwidth]{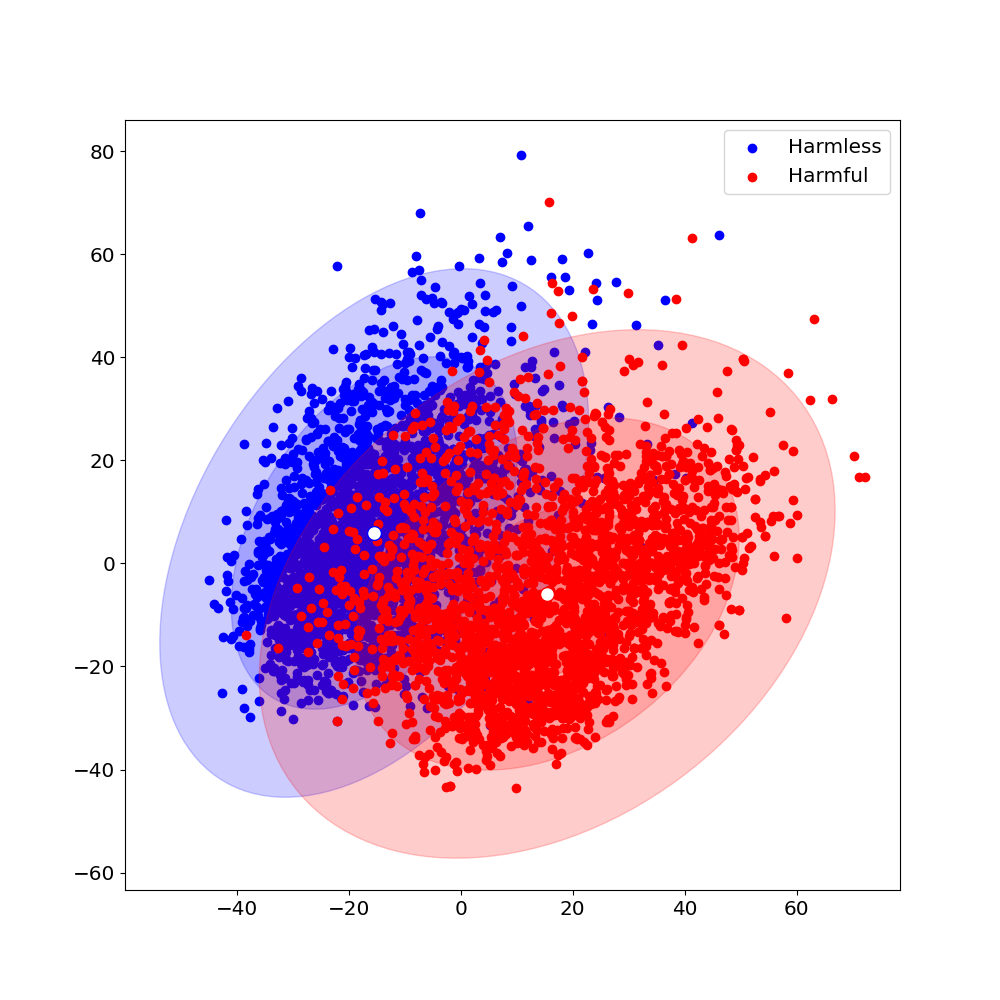} 
        \caption{$\pi_{\theta}$-en}
        \label{fig:firstsubfig5}
    \end{subfigure}
    \begin{subfigure}[b]{0.24\textwidth} 
        \centering
        \includegraphics[width=\textwidth]{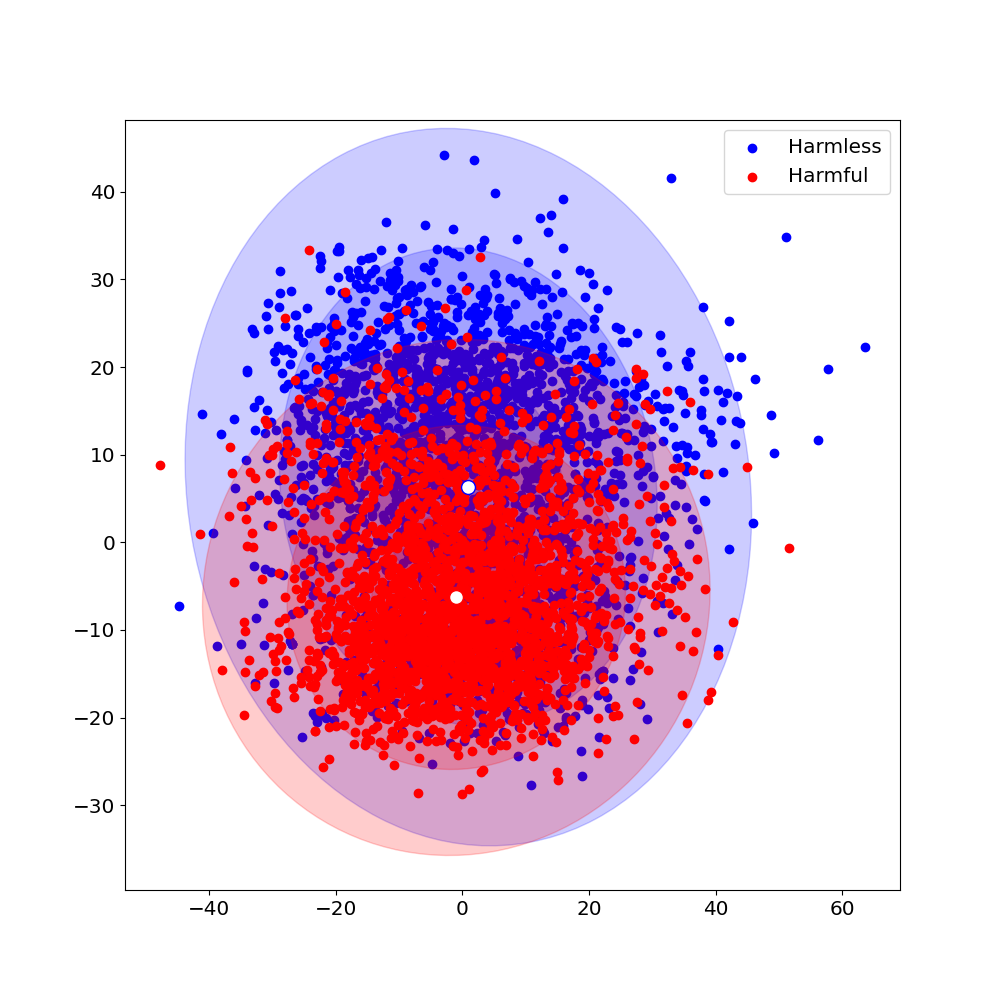}  
        \caption{$\pi_{\theta}$-hi}
        \label{fig:firstsubfig6}
    \end{subfigure}
    \begin{subfigure}[b]{0.24\textwidth} 
        \centering
        \includegraphics[width=\textwidth]{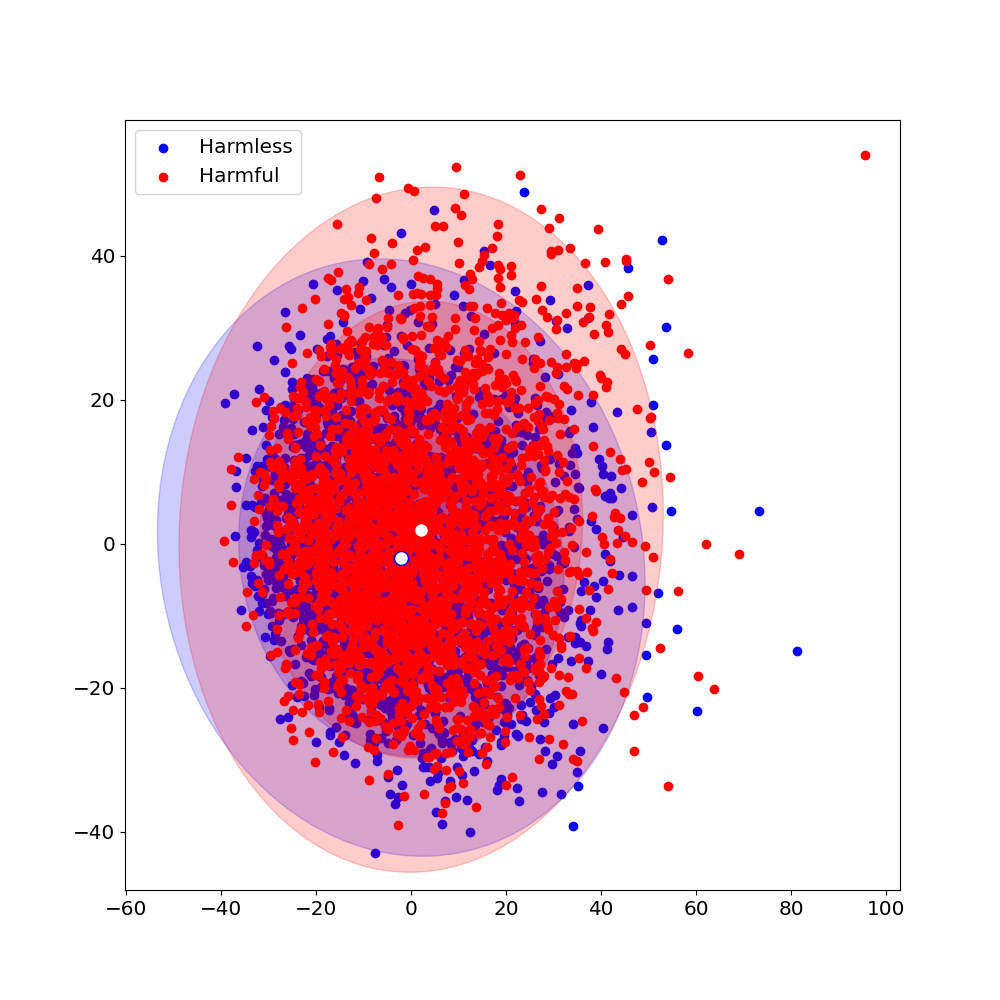} 
        \caption{$\pi_{\theta}$-zh}
        \label{fig:firstsubfig7}
    \end{subfigure}
    \begin{subfigure}[b]{0.24\textwidth} 
        \centering
        \includegraphics[width=\textwidth]{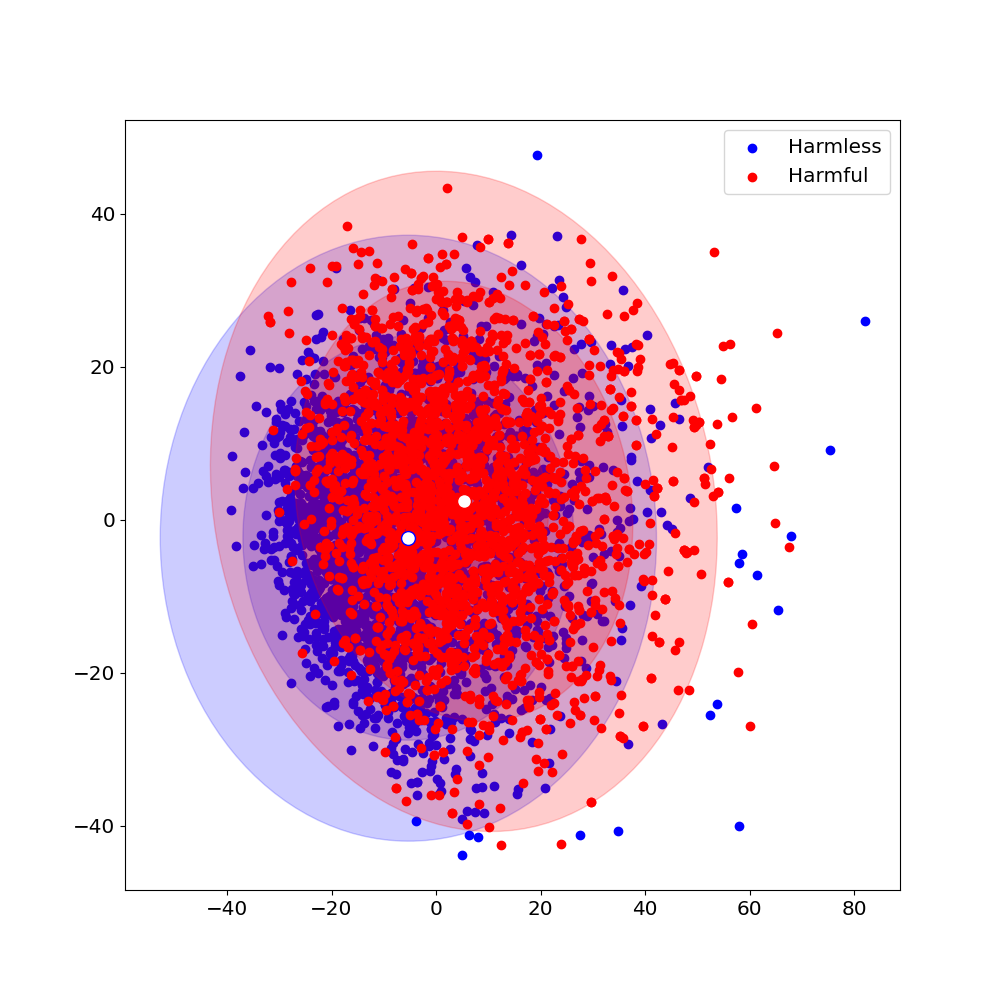} 
        \caption{$\pi_{\theta}$-de}
        \label{fig:firstsubfig8}
    \end{subfigure}

    \caption{Impact of Alignment on Hidden Representations in Qwen-2.5 for Multilingual Corpora.}
    \label{fig:before_after_alignment_qwen_2.5_all_langs}
\end{figure*}

\begin{figure*}[t]
    \centering
    
    
    \renewcommand{\thesubfigure}{\alph{subfigure}.\arabic{subfigure}} 
    \setcounter{subfigure}{0} 
    \begin{subfigure}[b]{0.24\textwidth}  
        \centering
        \includegraphics[width=\textwidth]{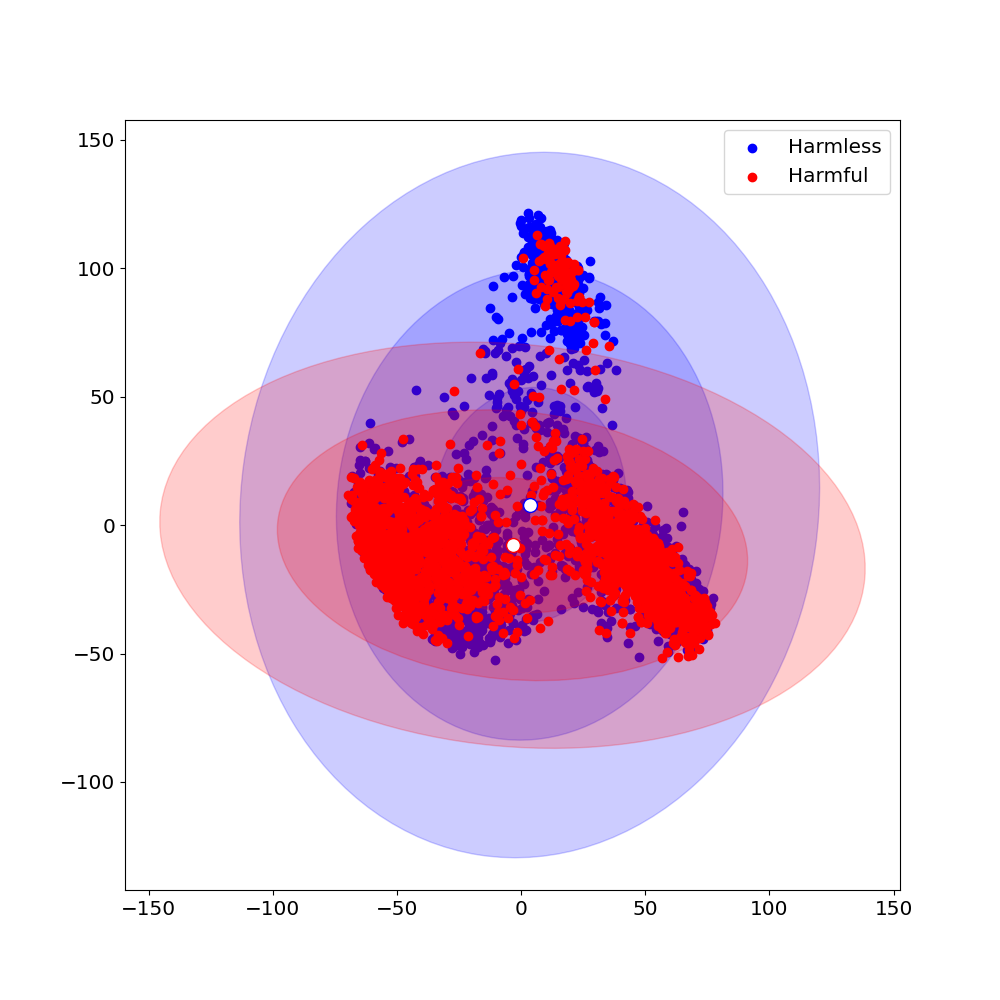}  
        \caption{$\pi_{\text{ref}}$-en}
        \label{fig:firstsubfig1}
    \end{subfigure}
    \begin{subfigure}[b]{0.24\textwidth}
        \centering
        \includegraphics[width=\textwidth]{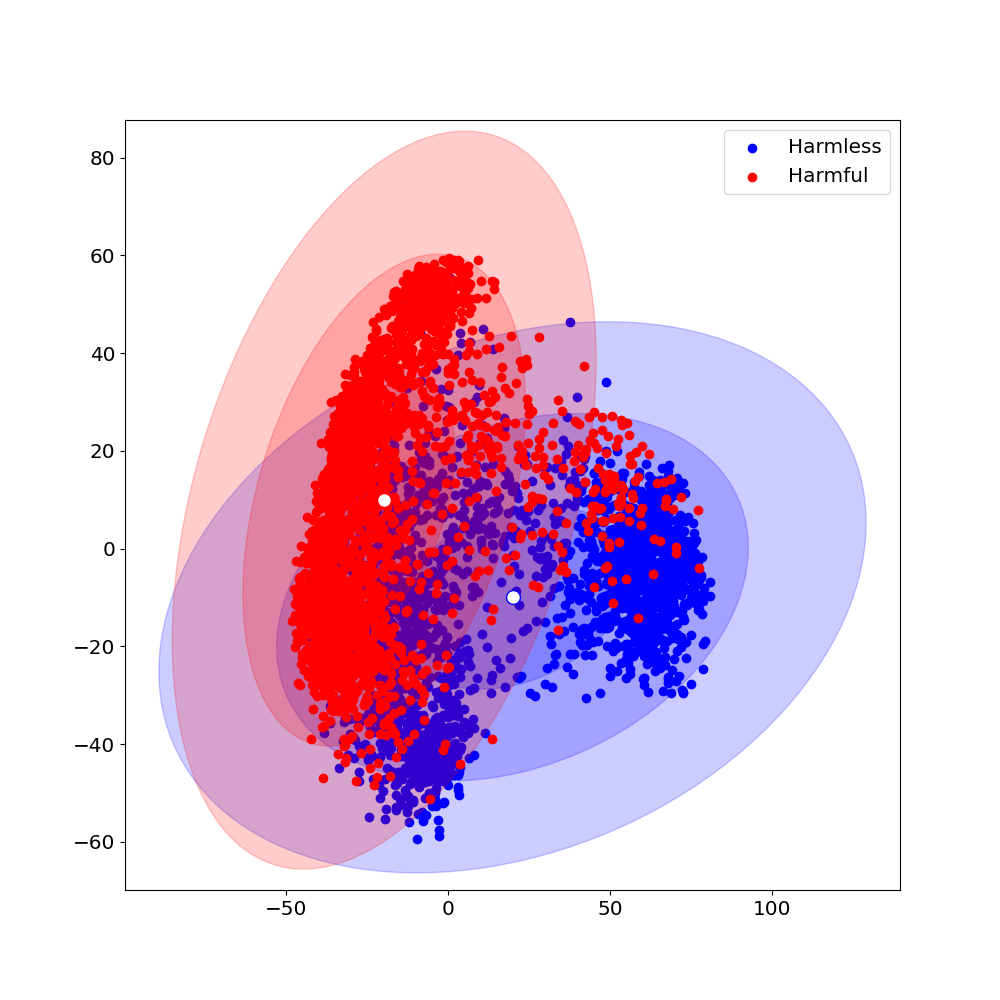} 
        \caption{$\pi_{\text{ref}}$-hi}
        \label{fig:firstsubfig2}
    \end{subfigure}
    \begin{subfigure}[b]{0.24\textwidth}  
        \centering
        \includegraphics[width=\textwidth]{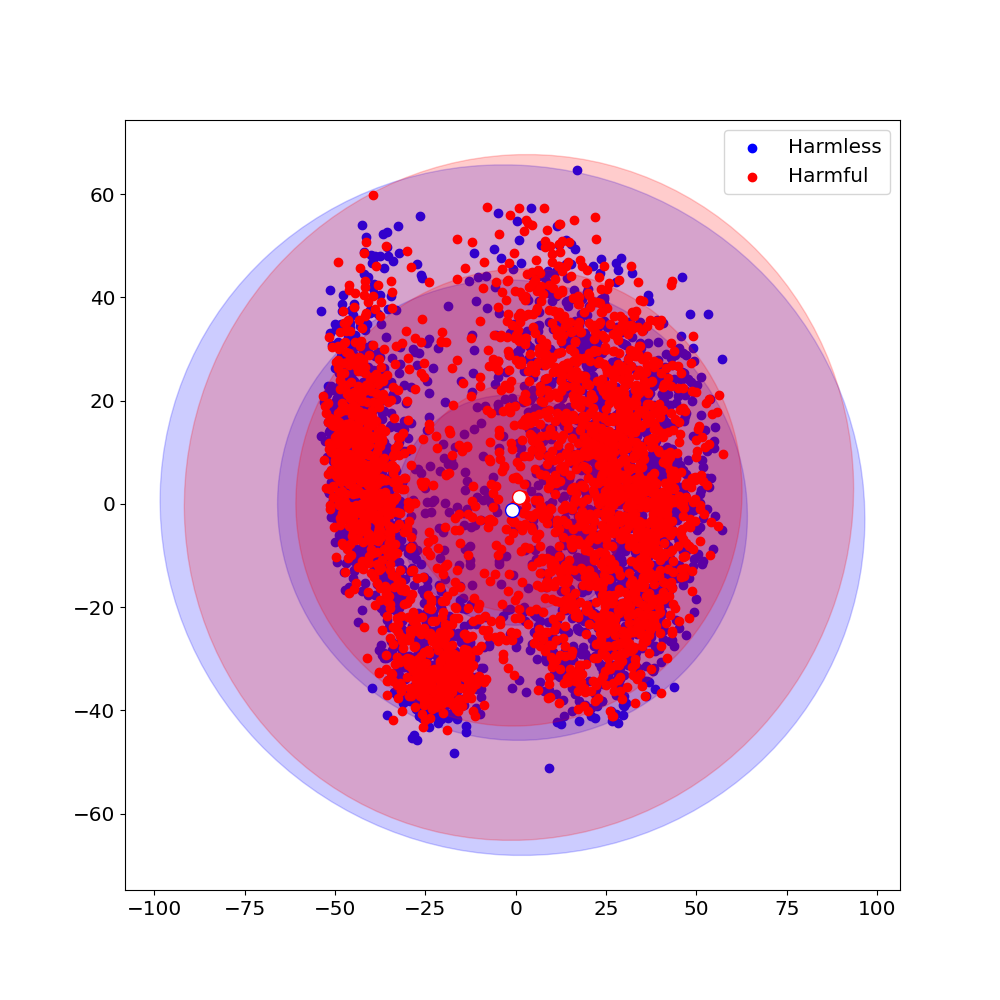} 
        \caption{$\pi_{\text{ref}}$-zh}
        \label{fig:firstsubfig3}
    \end{subfigure}
    \begin{subfigure}[b]{0.24\textwidth}  
        \centering
        \includegraphics[width=\textwidth]{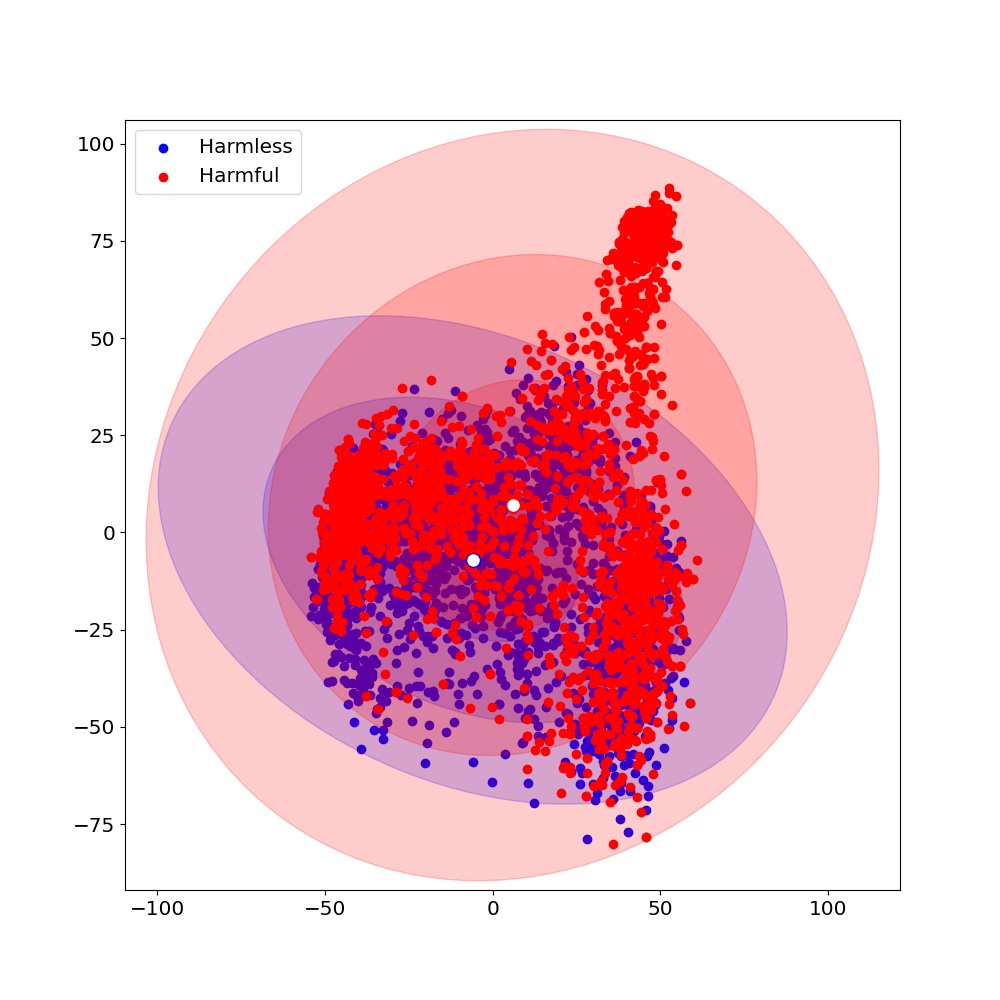} 
        \caption{$\pi_{\text{ref}}$-de}
        \label{fig:firstsubfig4}
    \end{subfigure}

    \renewcommand{\thesubfigure}{\alph{subfigure}.\arabic{subfigure}} 
    \setcounter{subfigure}{0} 
    \begin{subfigure}[b]{0.24\textwidth} 
        \centering
        \includegraphics[width=\textwidth]{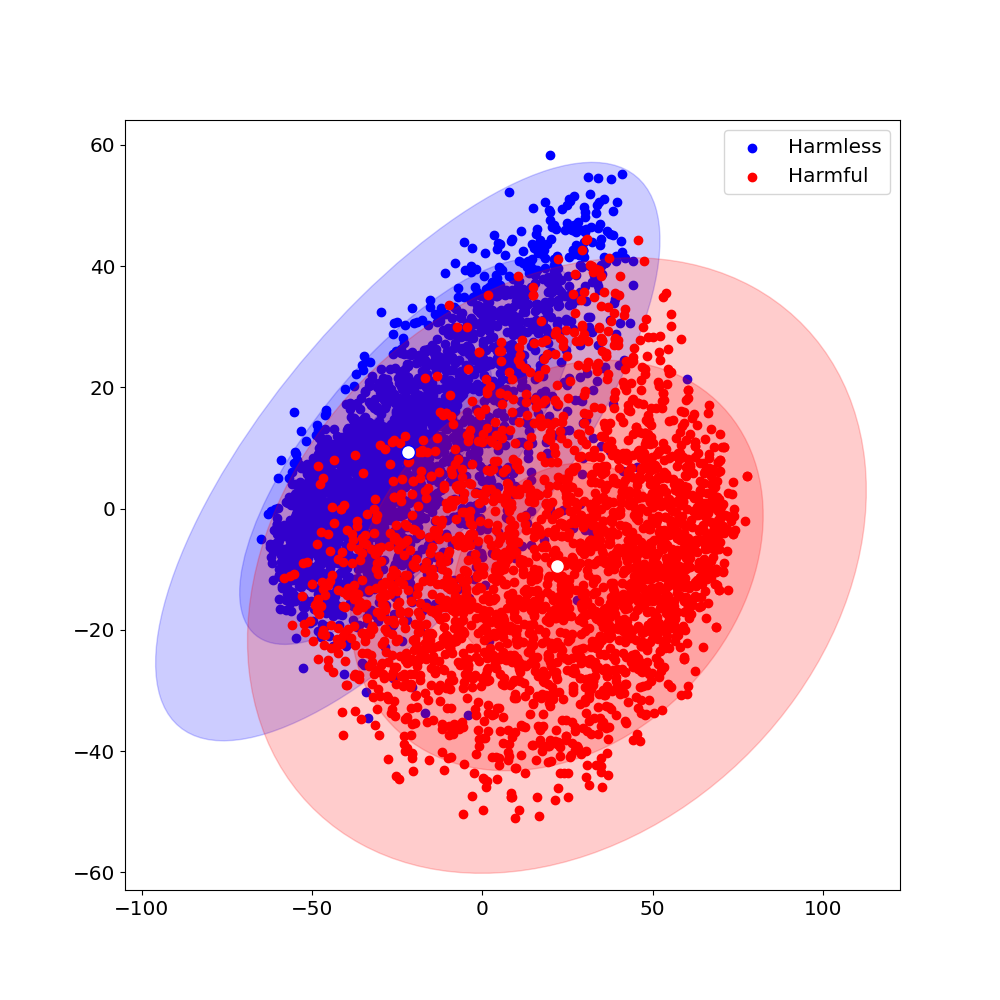} 
        \caption{$\pi_{\theta}$-en}
        \label{fig:firstsubfig5}
    \end{subfigure}
    \begin{subfigure}[b]{0.24\textwidth} 
        \centering
        \includegraphics[width=\textwidth]{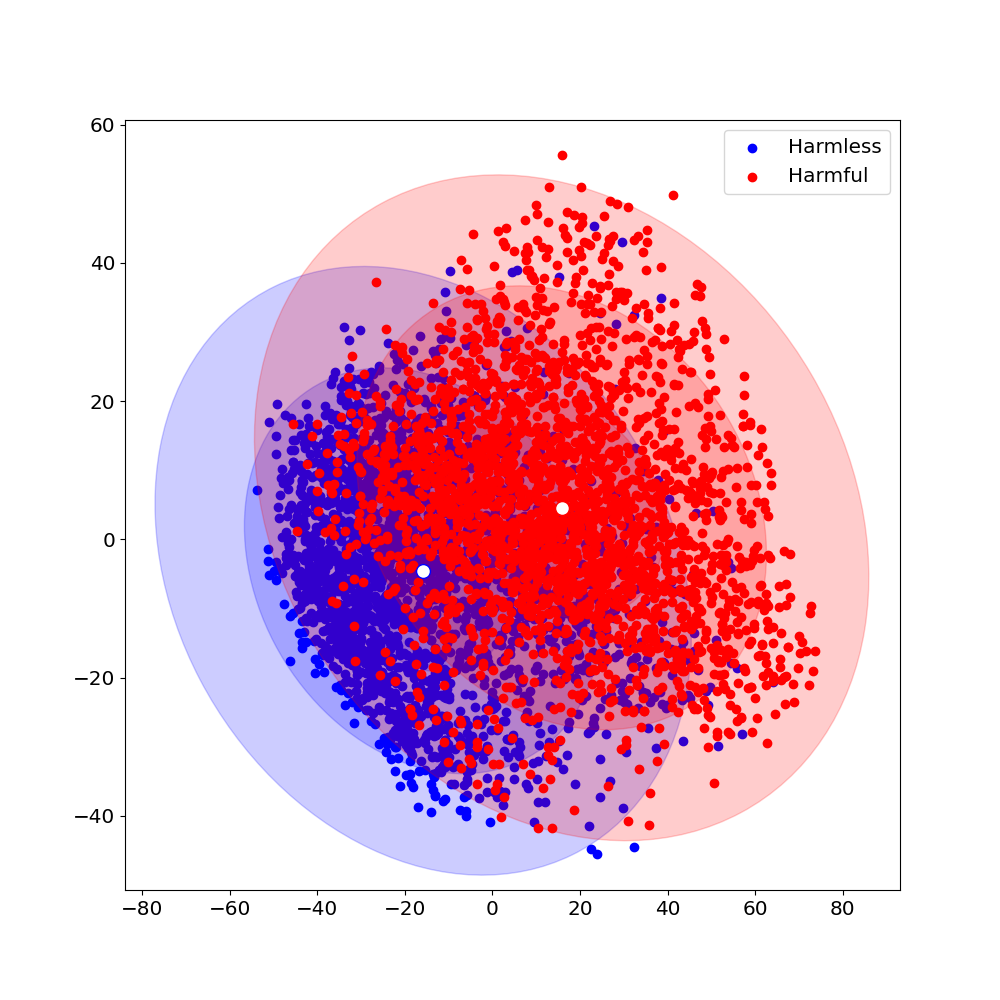}  
        \caption{$\pi_{\theta}$-hi}
        \label{fig:firstsubfig6}
    \end{subfigure}
    \begin{subfigure}[b]{0.24\textwidth} 
        \centering
        \includegraphics[width=\textwidth]{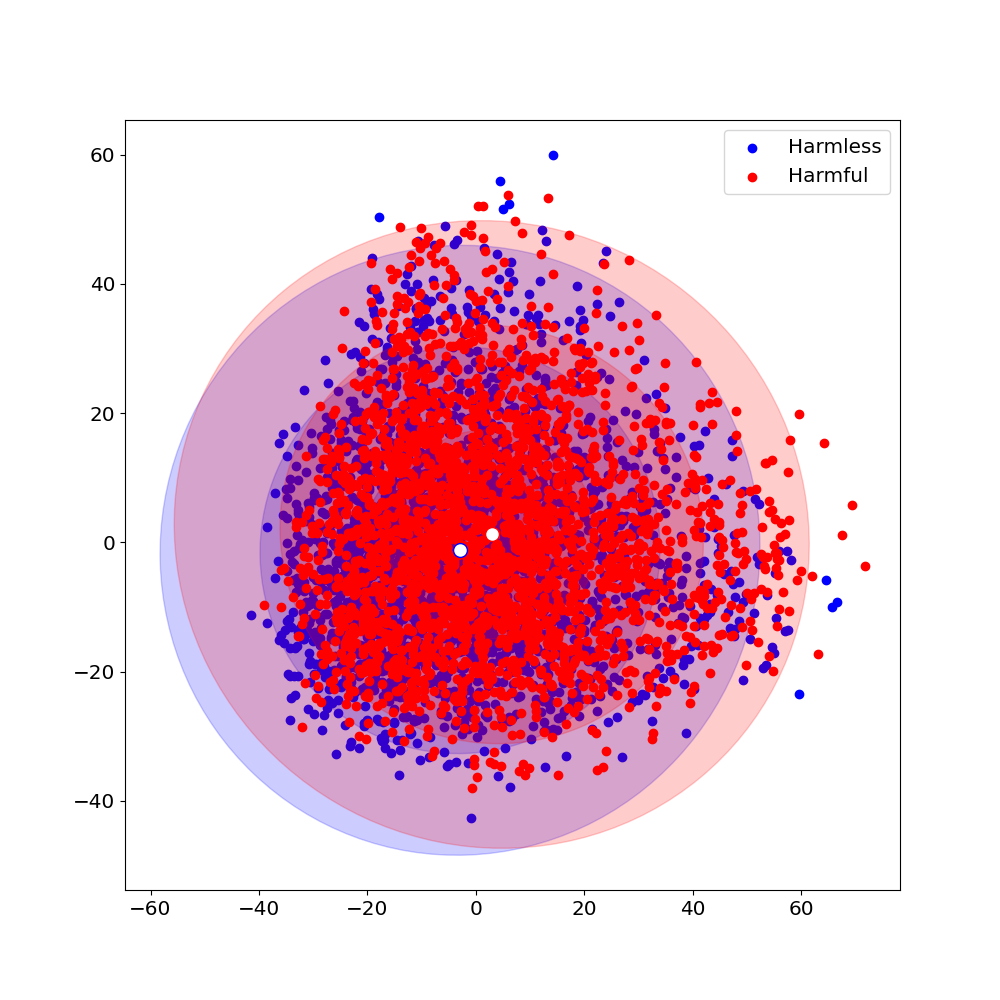} 
        \caption{$\pi_{\theta}$-zh}
        \label{fig:firstsubfig7}
    \end{subfigure}
    \begin{subfigure}[b]{0.24\textwidth} 
        \centering
        \includegraphics[width=\textwidth]{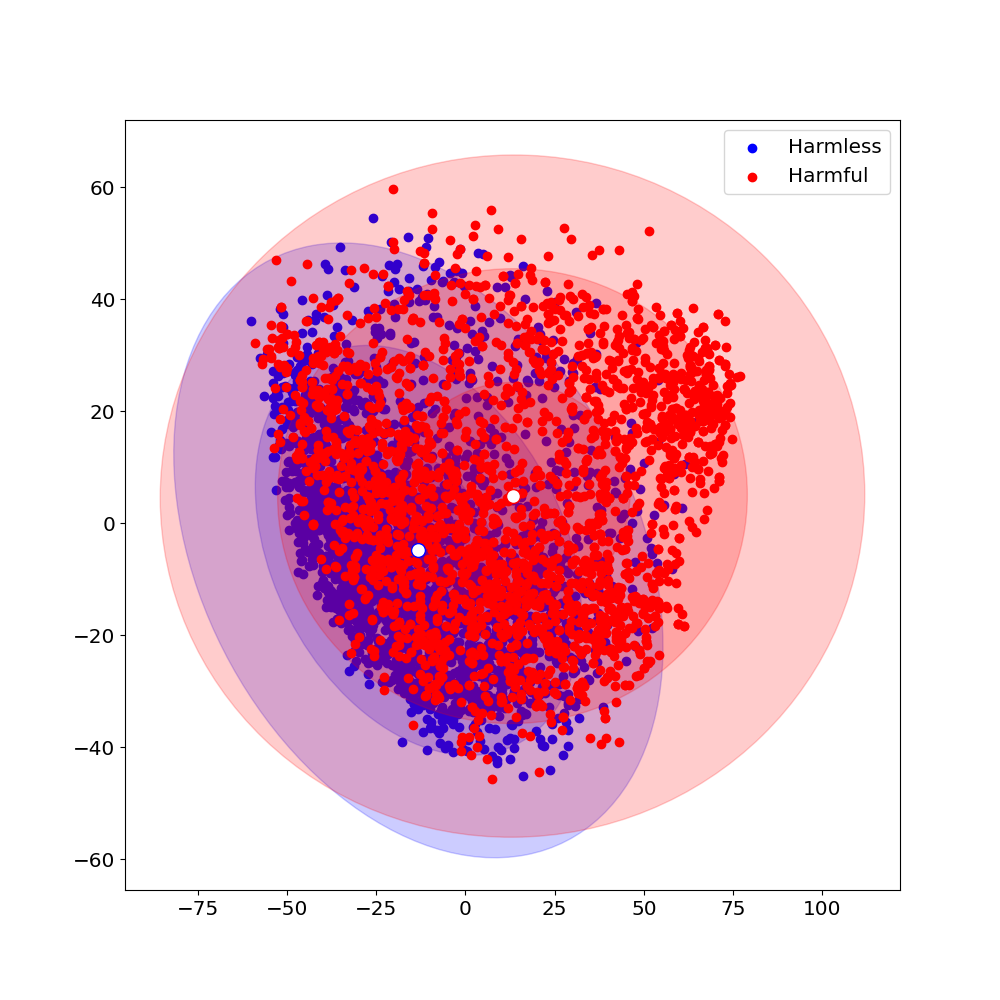} 
        \caption{$\pi_{\theta}$-de}
        \label{fig:firstsubfig8}
    \end{subfigure}

    \caption{Impact of Alignment on Hidden Representations in Llama-3.1 for Multilingual Corpora.}
    \label{fig:before_after_alignment_llama3.1_8b_all_langs}
\end{figure*}

\begin{figure*}[t]
    \centering
    
    
    \renewcommand{\thesubfigure}{\alph{subfigure}.\arabic{subfigure}} 
    \setcounter{subfigure}{0} 
    \begin{subfigure}[b]{0.24\textwidth}  
        \centering
        \includegraphics[width=\textwidth]{latex/images/pca2_balanced_toxic_dataset_Llama-3.1-8B_en.png}  
        \caption{$\pi_{\text{ref}}$-en}
        \label{fig:firstsubfig1}
    \end{subfigure}
    \begin{subfigure}[b]{0.24\textwidth}
        \centering
        \includegraphics[width=\textwidth]{latex/images/pca2_balanced_toxic_dataset_Llama-3.1-8B_hi.png} 
        \caption{$\pi_{\text{ref}}$-hi}
        \label{fig:firstsubfig2}
    \end{subfigure}
    \begin{subfigure}[b]{0.24\textwidth}  
        \centering
        \includegraphics[width=\textwidth]{latex/images/pca2_balanced_toxic_dataset_Llama-3.1-8B_zh.png} 
        \caption{$\pi_{\text{ref}}$-zh}
        \label{fig:firstsubfig3}
    \end{subfigure}
    \begin{subfigure}[b]{0.24\textwidth}  
        \centering
        \includegraphics[width=\textwidth]{latex/images/pca2_balanced_toxic_dataset_Llama-3.1-8B_de.png} 
        \caption{$\pi_{\text{ref}}$-de}
        \label{fig:firstsubfig4}
    \end{subfigure}

    \renewcommand{\thesubfigure}{\alph{subfigure}.\arabic{subfigure}} 
    \setcounter{subfigure}{0} 
    \begin{subfigure}[b]{0.24\textwidth} 
        \centering
        \includegraphics[width=\textwidth]{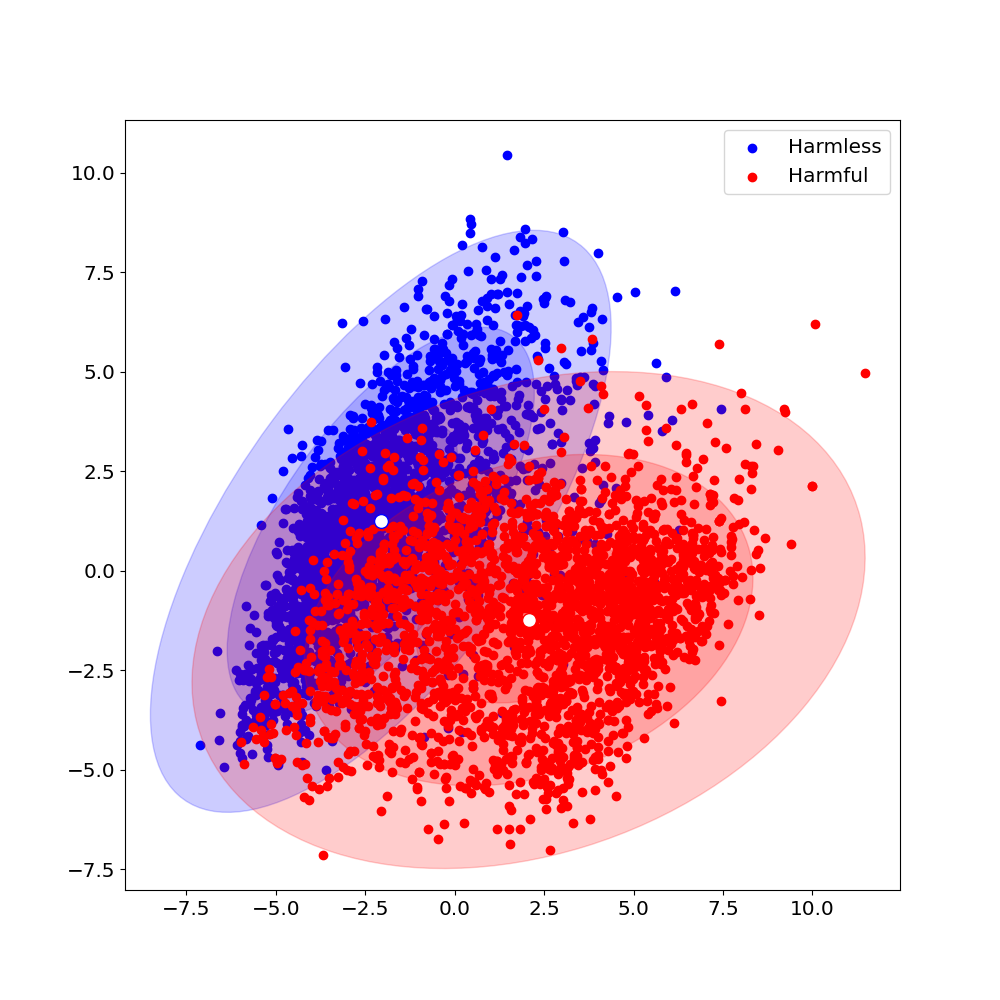} 
        \caption{$\pi_{\theta}$-en}
        \label{fig:firstsubfig5}
    \end{subfigure}
    \begin{subfigure}[b]{0.24\textwidth} 
        \centering
        \includegraphics[width=\textwidth]{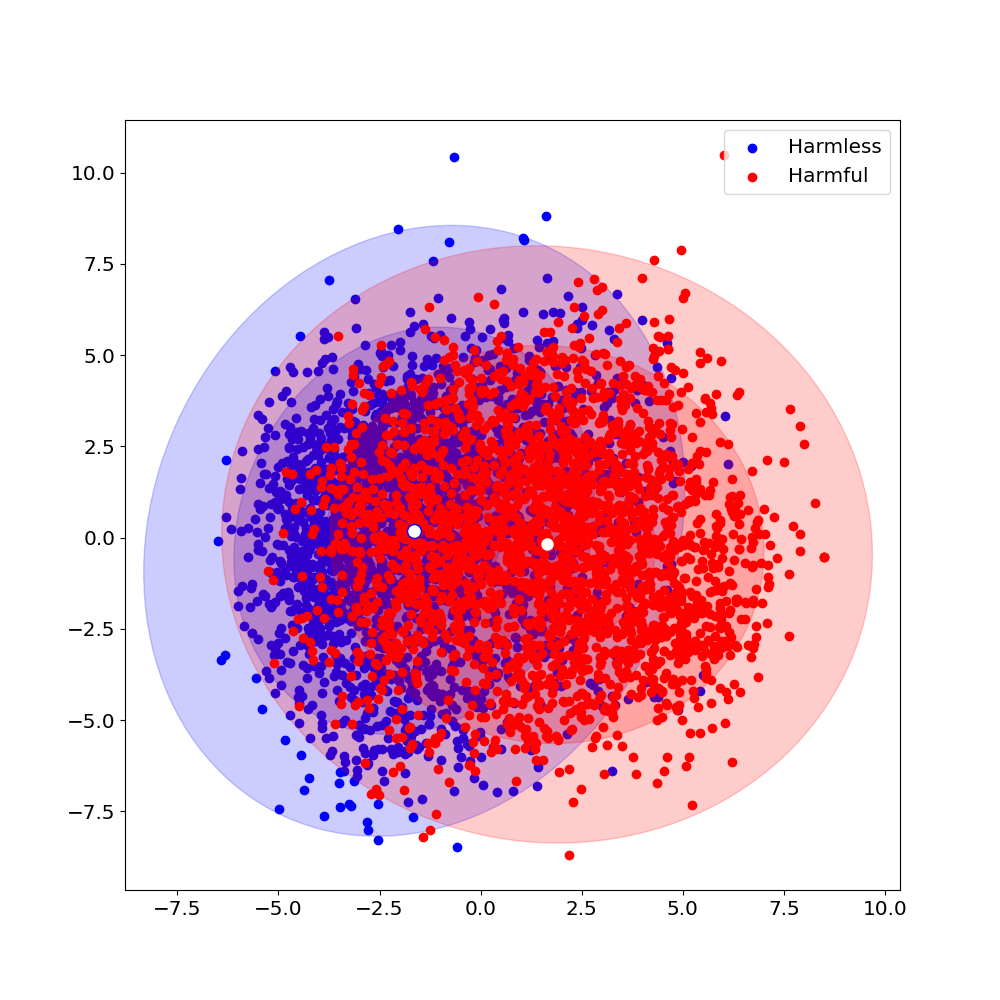}  
        \caption{$\pi_{\theta}$-hi}
        \label{fig:firstsubfig6}
    \end{subfigure}
    \begin{subfigure}[b]{0.24\textwidth} 
        \centering
        \includegraphics[width=\textwidth]{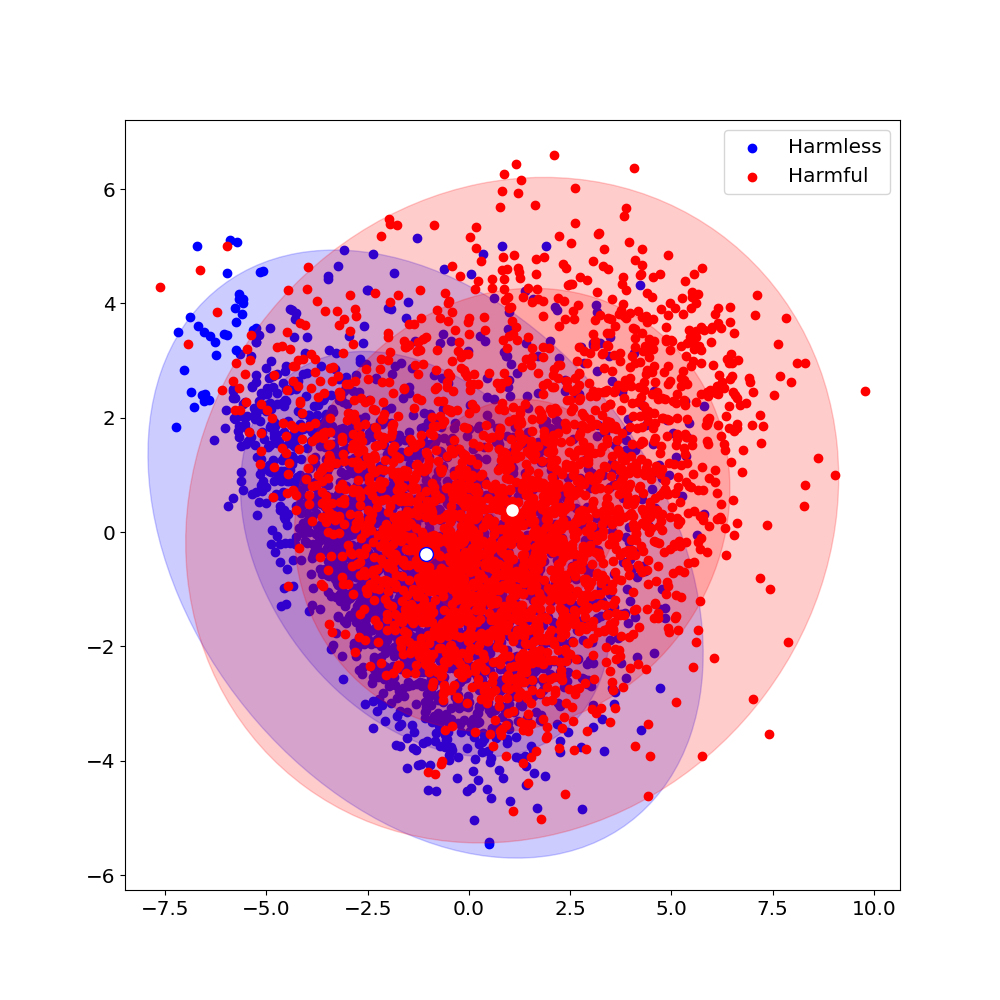} 
        \caption{$\pi_{\theta}$-zh}
        \label{fig:firstsubfig7}
    \end{subfigure}
    \begin{subfigure}[b]{0.24\textwidth} 
        \centering
        \includegraphics[width=\textwidth]{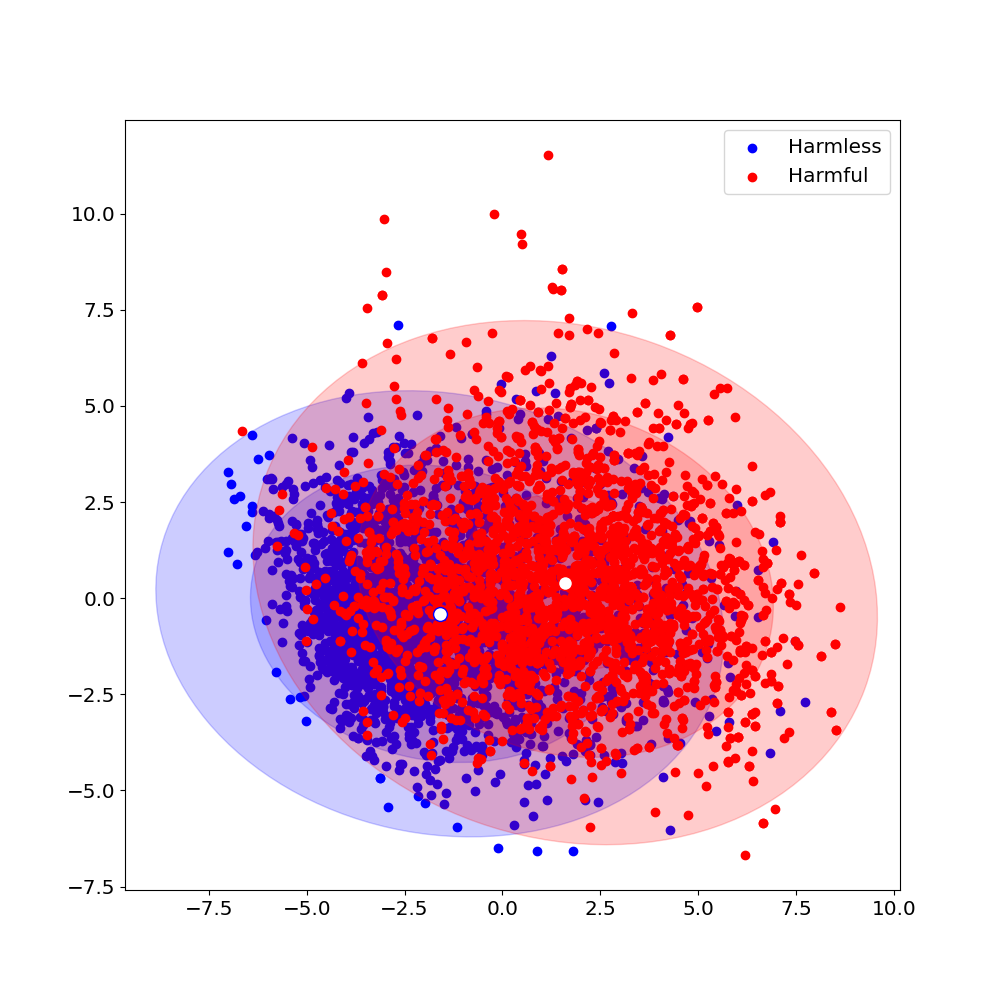} 
        \caption{$\pi_{\theta}$-de}
        \label{fig:firstsubfig8}
    \end{subfigure}

    \caption{Impact of Alignment on Hidden Representations in Llama-Guard-3 for Multilingual Corpora.}
    \label{fig:before_after_alignment_llamaguard_all_langs}
\end{figure*}

\begin{figure*}[t]
    \centering
    
    
    \renewcommand{\thesubfigure}{\alph{subfigure}.\arabic{subfigure}} 
    \setcounter{subfigure}{0} 
    \begin{subfigure}[b]{0.24\textwidth}  
        \centering
        \includegraphics[width=\textwidth]{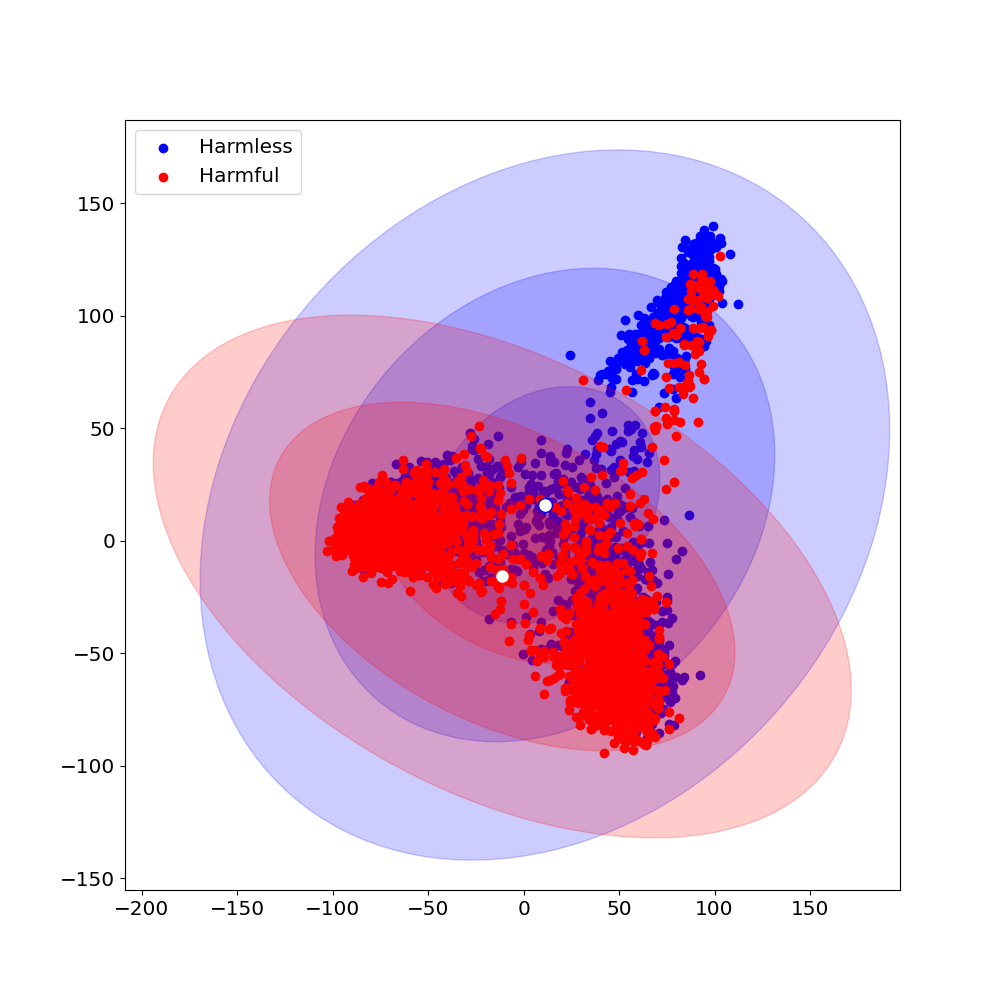}  
        \caption{$\pi_{\text{ref}}$-en}
        \label{fig:firstsubfig1}
    \end{subfigure}
    \begin{subfigure}[b]{0.24\textwidth}
        \centering
        \includegraphics[width=\textwidth]{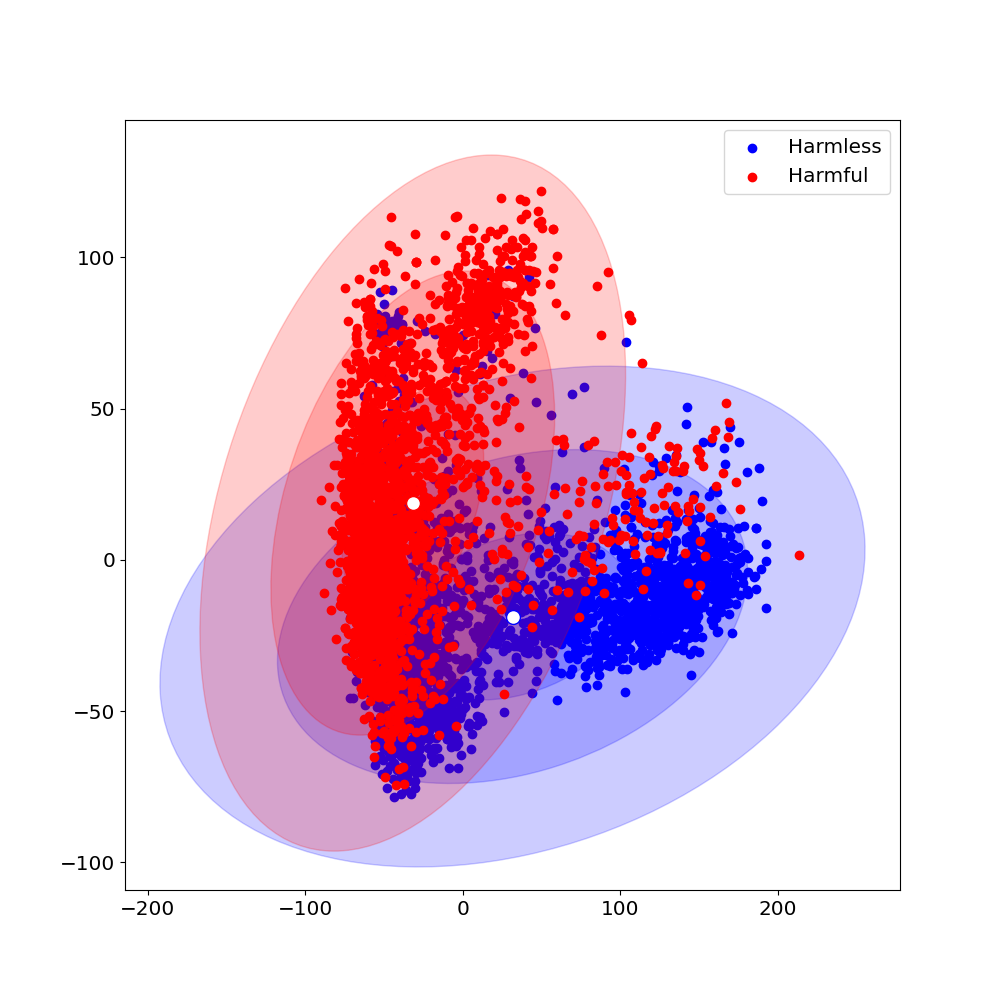} 
        \caption{$\pi_{\text{ref}}$-hi}
        \label{fig:firstsubfig2}
    \end{subfigure}
    \begin{subfigure}[b]{0.24\textwidth}  
        \centering
        \includegraphics[width=\textwidth]{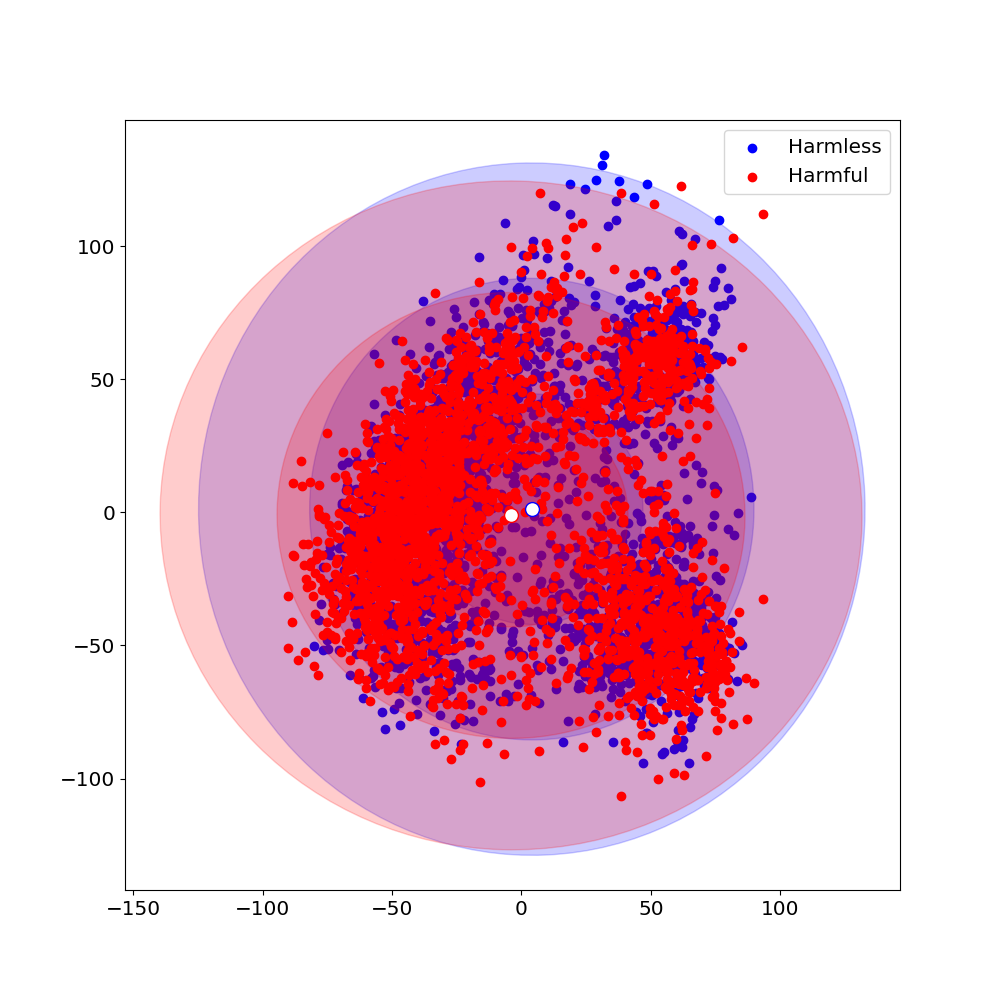} 
        \caption{$\pi_{\text{ref}}$-zh}
        \label{fig:firstsubfig3}
    \end{subfigure}
    \begin{subfigure}[b]{0.24\textwidth}  
        \centering
        \includegraphics[width=\textwidth]{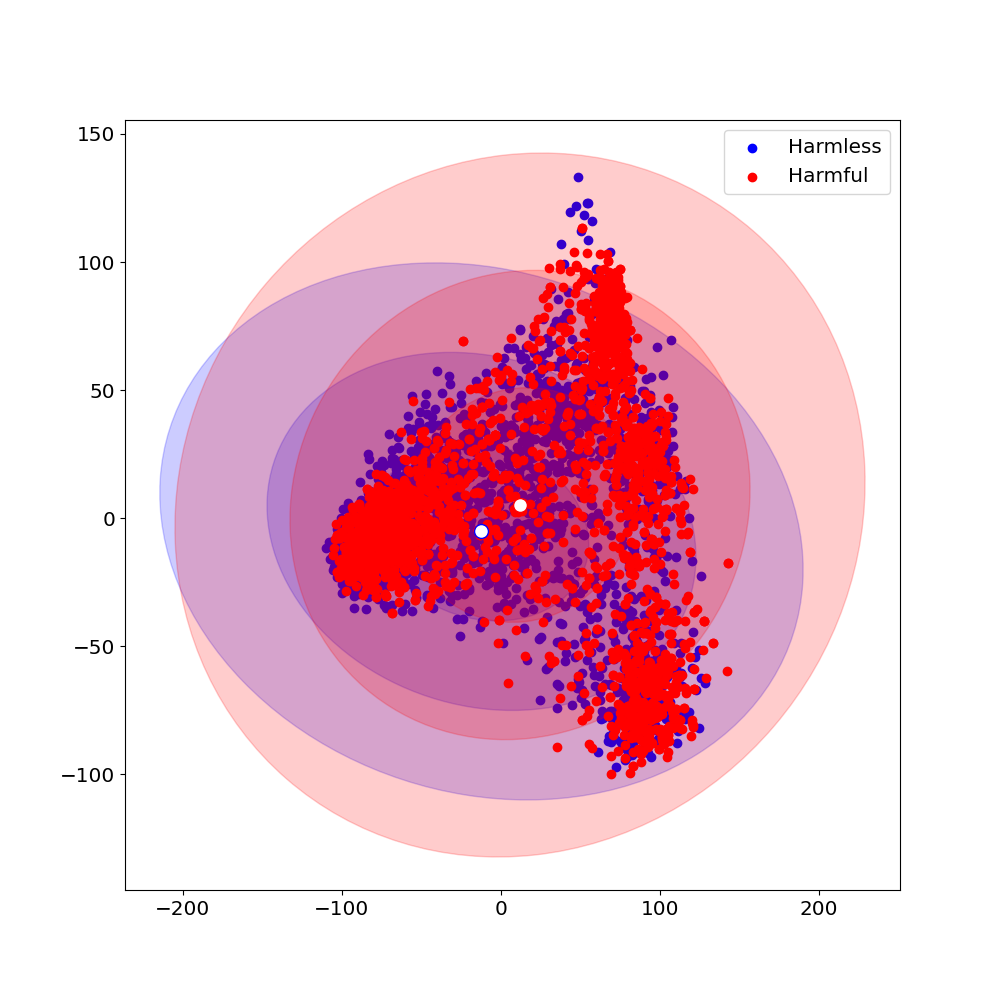} 
        \caption{$\pi_{\text{ref}}$-de}
        \label{fig:firstsubfig4}
    \end{subfigure}

    \renewcommand{\thesubfigure}{\alph{subfigure}.\arabic{subfigure}} 
    \setcounter{subfigure}{0} 
    \begin{subfigure}[b]{0.24\textwidth} 
        \centering
        \includegraphics[width=\textwidth]{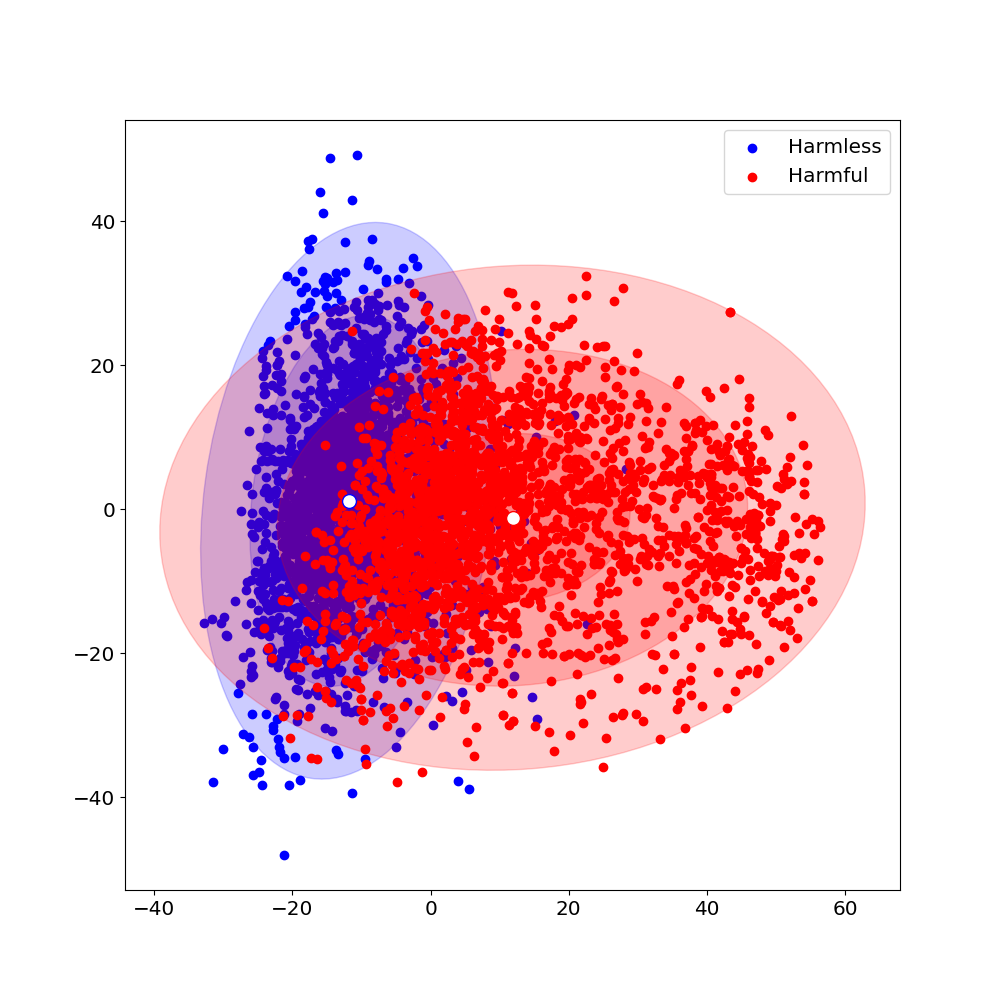} 
        \caption{$\pi_{\theta}$-en}
        \label{fig:firstsubfig5}
    \end{subfigure}
    \begin{subfigure}[b]{0.24\textwidth} 
        \centering
        \includegraphics[width=\textwidth]{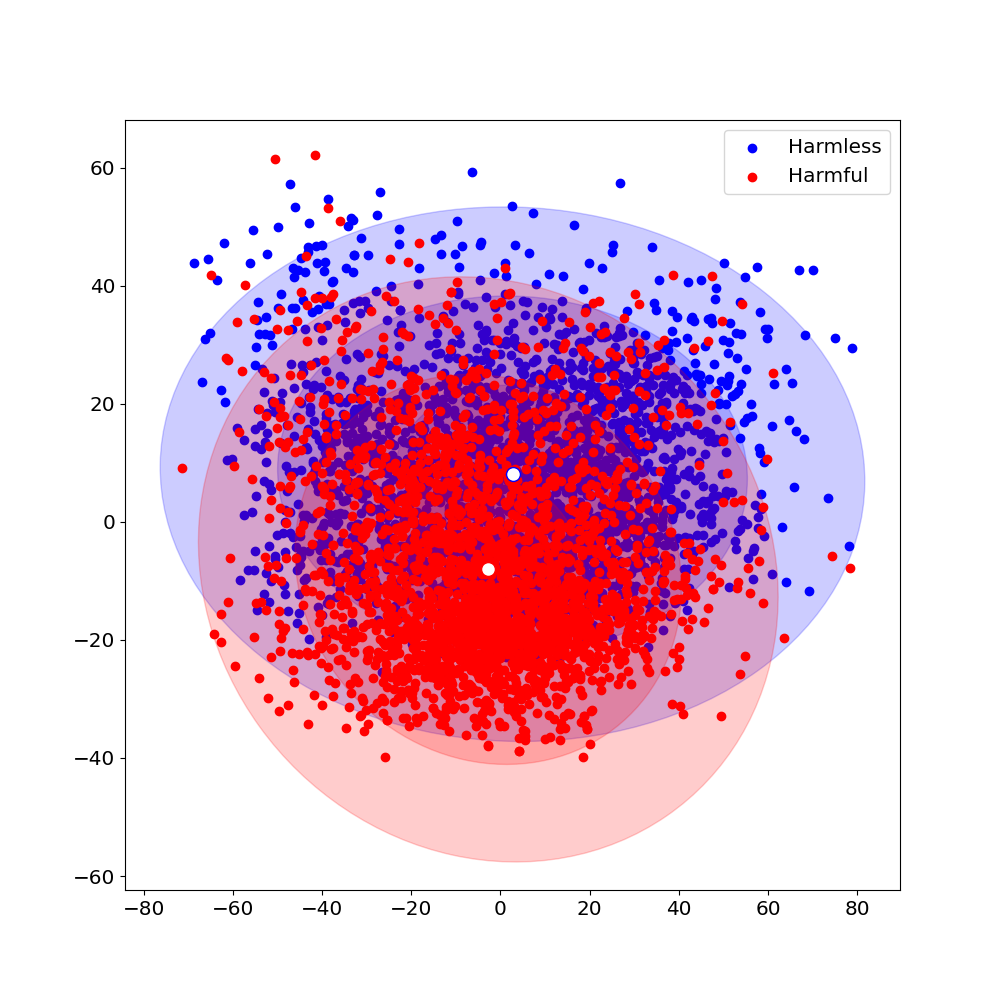}  
        \caption{$\pi_{\theta}$-hi}
        \label{fig:firstsubfig6}
    \end{subfigure}
    \begin{subfigure}[b]{0.24\textwidth} 
        \centering
        \includegraphics[width=\textwidth]{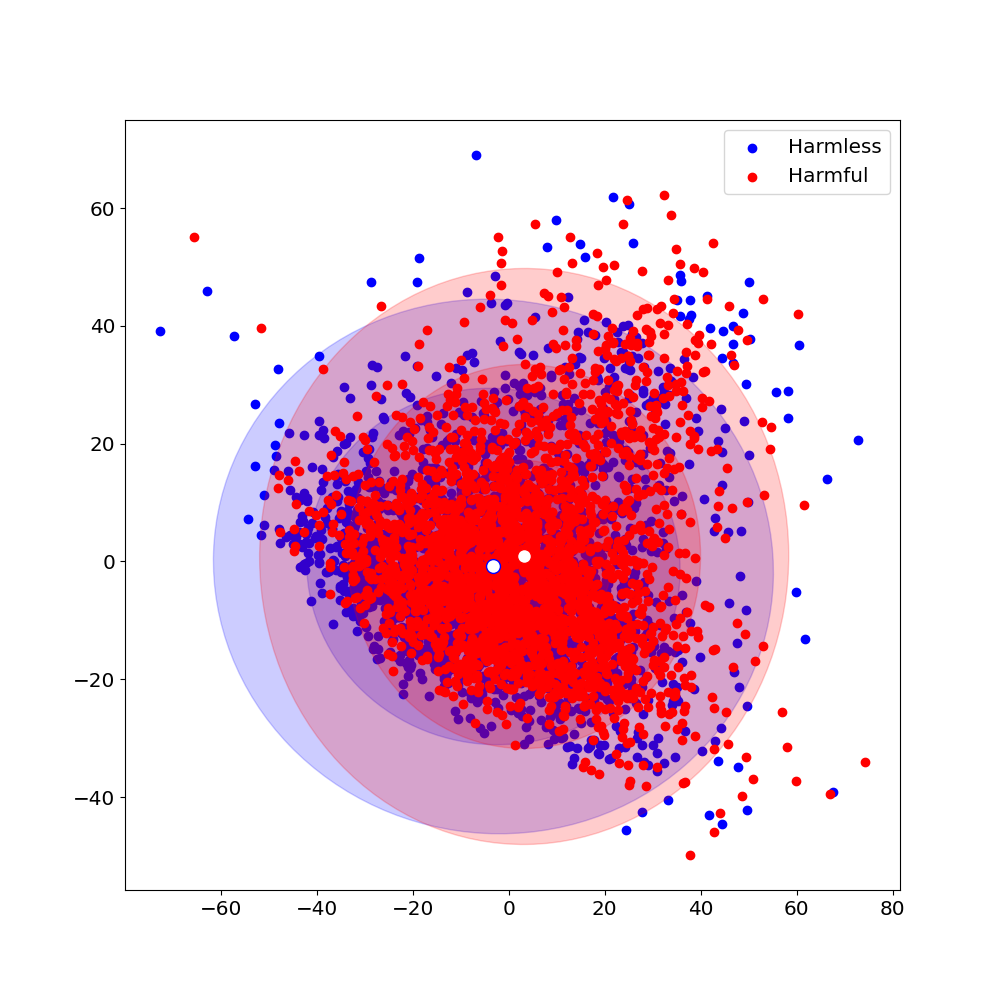} 
        \caption{$\pi_{\theta}$-zh}
        \label{fig:firstsubfig7}
    \end{subfigure}
    \begin{subfigure}[b]{0.24\textwidth} 
        \centering
        \includegraphics[width=\textwidth]{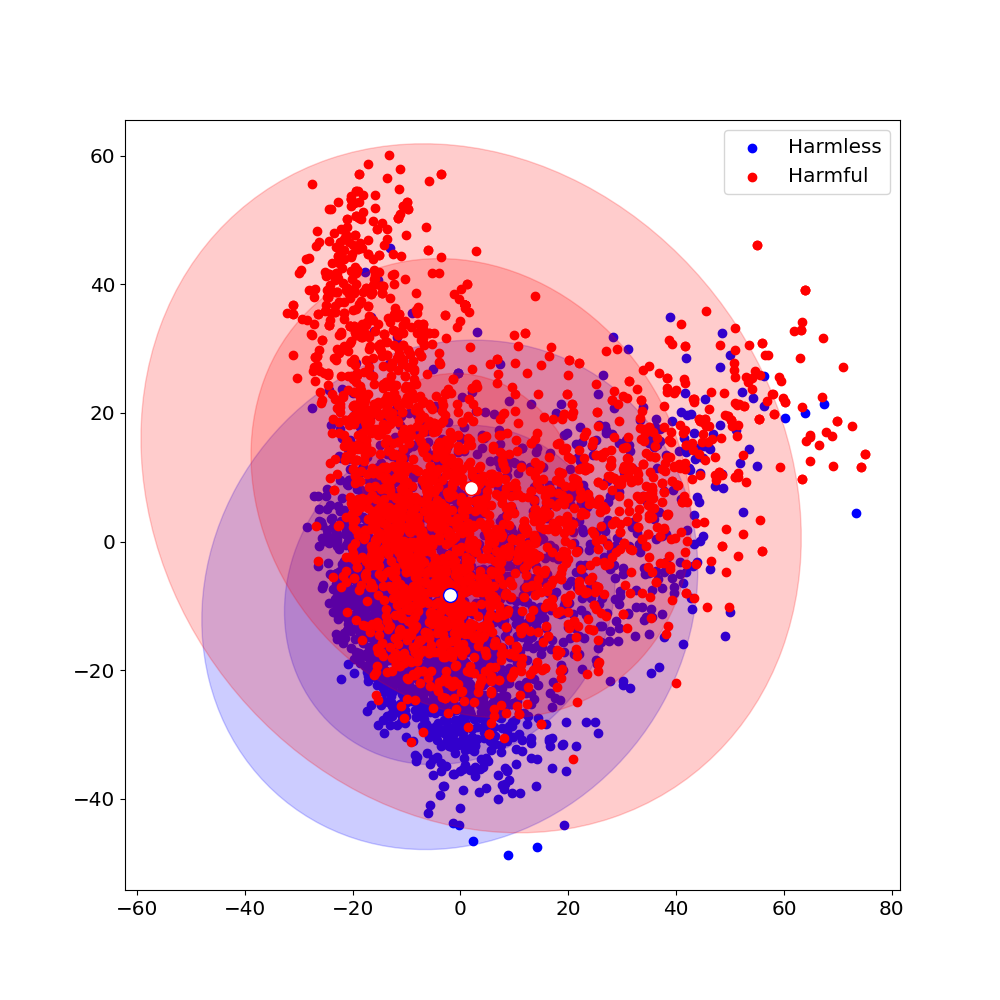} 
        \caption{$\pi_{\theta}$-de}
        \label{fig:firstsubfig8}
    \end{subfigure}

    \caption{Impact of Alignment on Hidden Representations in Gemma-2 for Multilingual Corpora.}
    \label{fig:before_after_alignment_gemma_2_all_langs}
\end{figure*}

\begin{figure*}[t]
    \centering
    
    
    \renewcommand{\thesubfigure}{\alph{subfigure}.\arabic{subfigure}} 
    \setcounter{subfigure}{0} 
    \begin{subfigure}[b]{0.24\textwidth}  
        \centering
        \includegraphics[width=\textwidth]{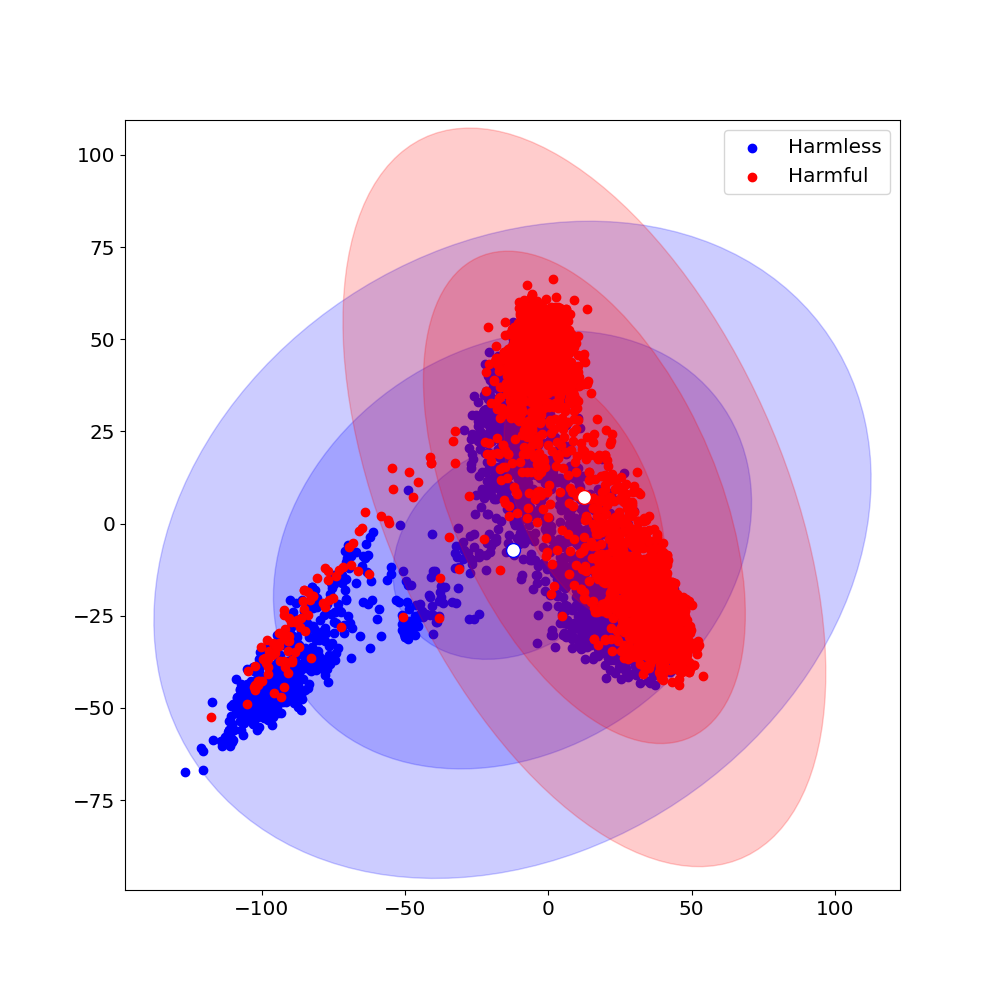}  
        \caption{$\pi_{\text{ref}}$-en}
        \label{fig:firstsubfig1}
    \end{subfigure}
    \begin{subfigure}[b]{0.24\textwidth}
        \centering
        \includegraphics[width=\textwidth]{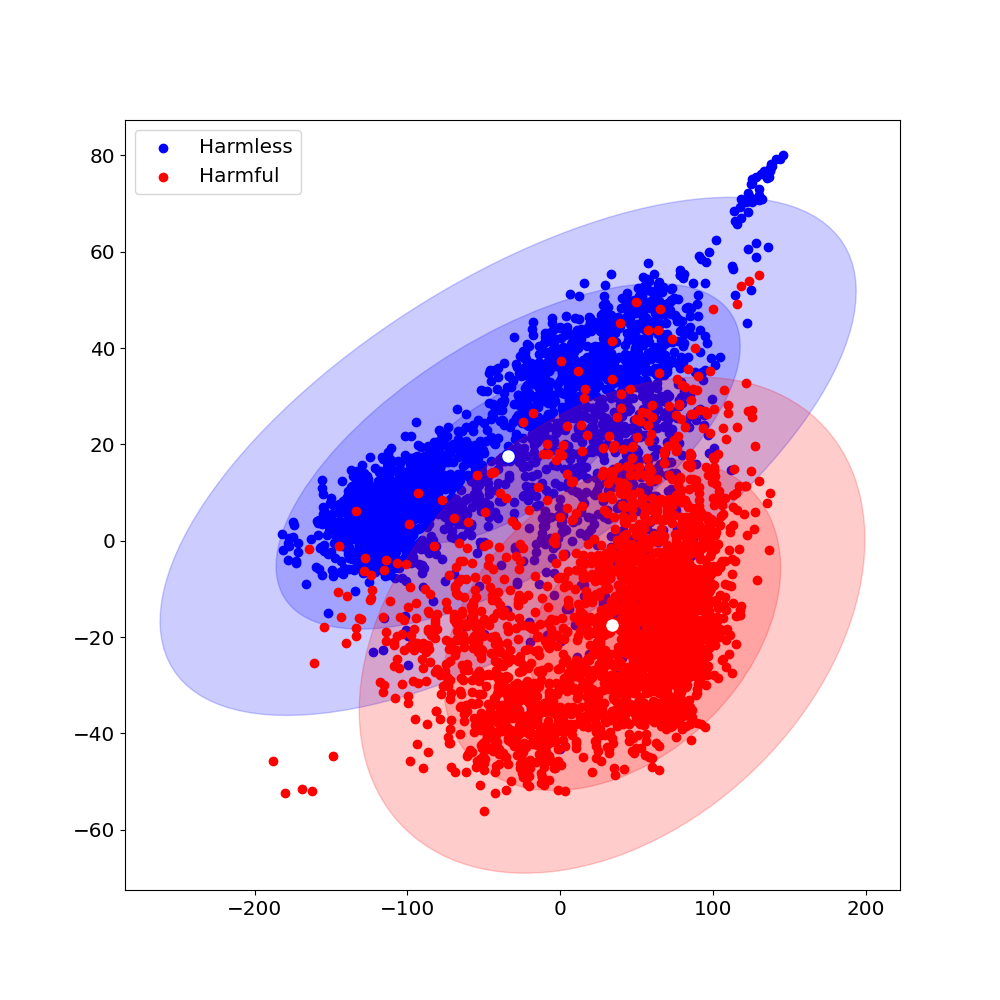} 
        \caption{$\pi_{\text{ref}}$-hi}
        \label{fig:firstsubfig2}
    \end{subfigure}
    \begin{subfigure}[b]{0.24\textwidth}  
        \centering
        \includegraphics[width=\textwidth]{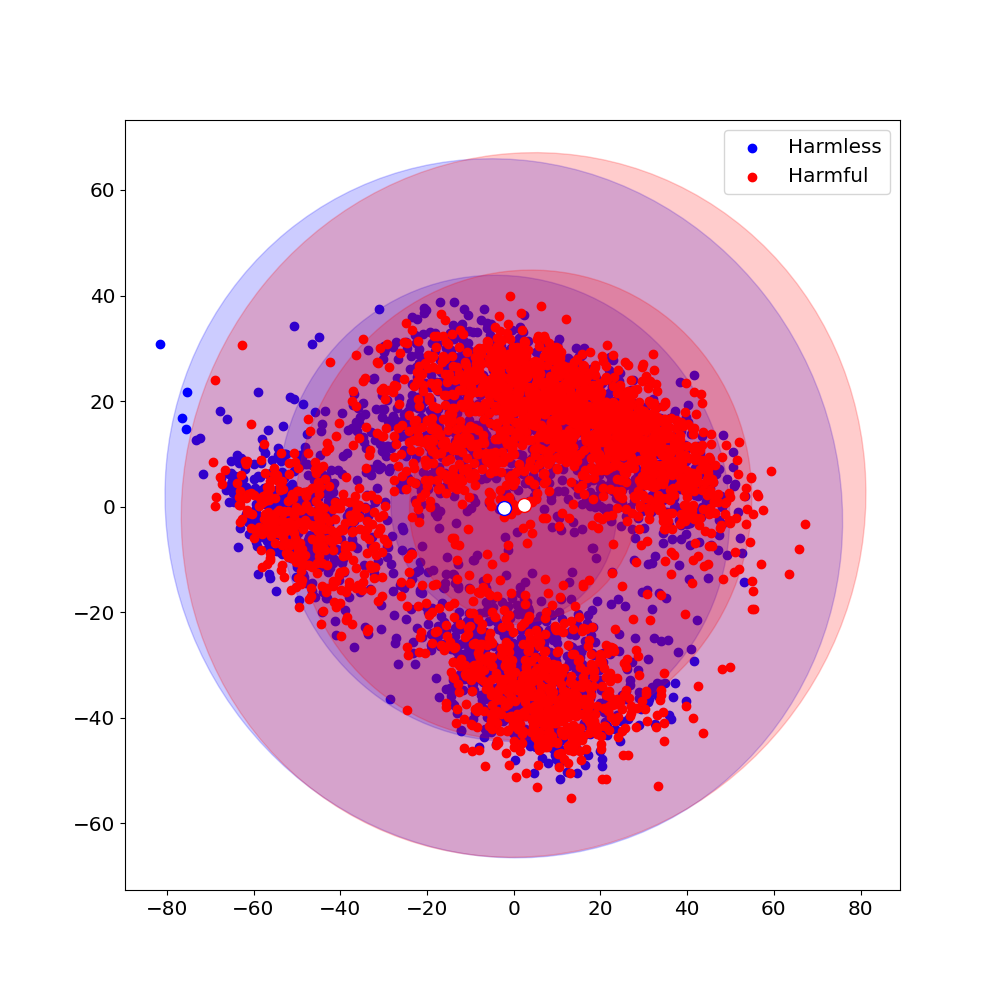} 
        \caption{$\pi_{\text{ref}}$-zh}
        \label{fig:firstsubfig3}
    \end{subfigure}
    \begin{subfigure}[b]{0.24\textwidth}  
        \centering
        \includegraphics[width=\textwidth]{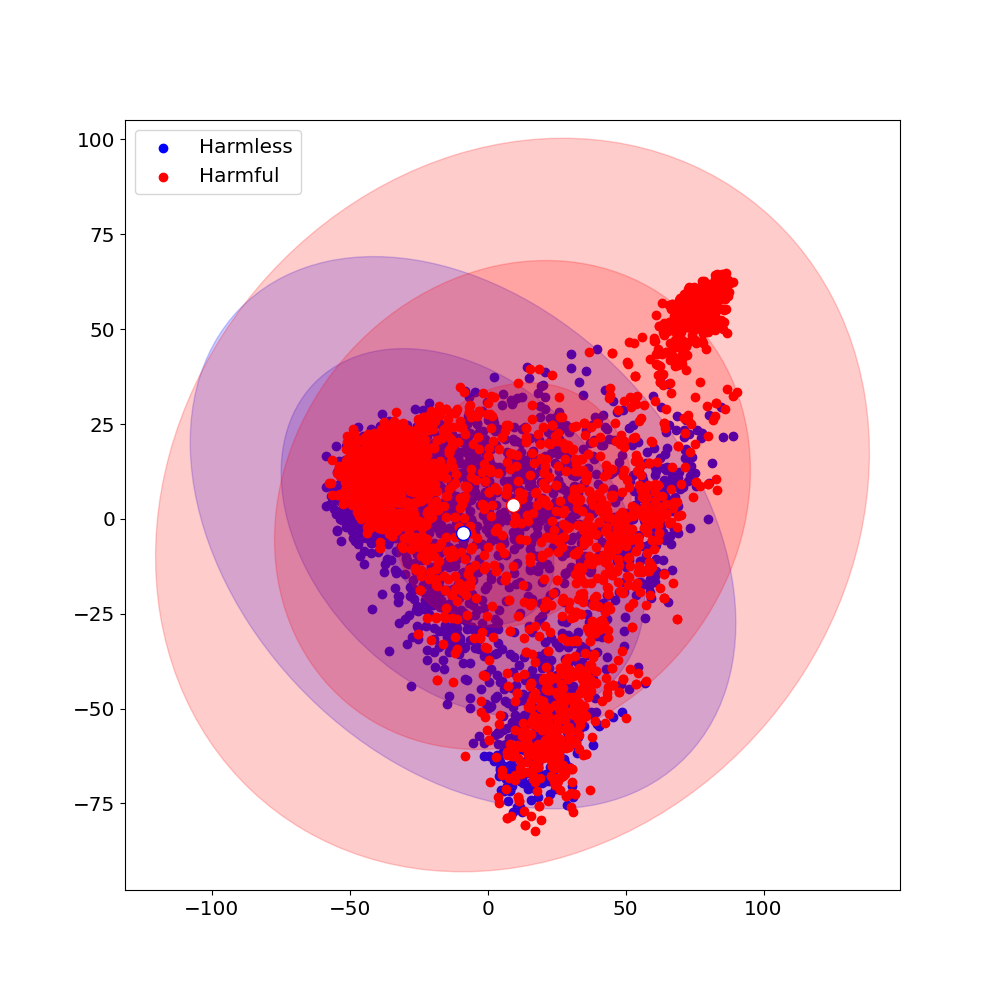} 
        \caption{$\pi_{\text{ref}}$-de}
        \label{fig:firstsubfig4}
    \end{subfigure}

    \renewcommand{\thesubfigure}{\alph{subfigure}.\arabic{subfigure}} 
    \setcounter{subfigure}{0} 
    \begin{subfigure}[b]{0.24\textwidth} 
        \centering
        \includegraphics[width=\textwidth]{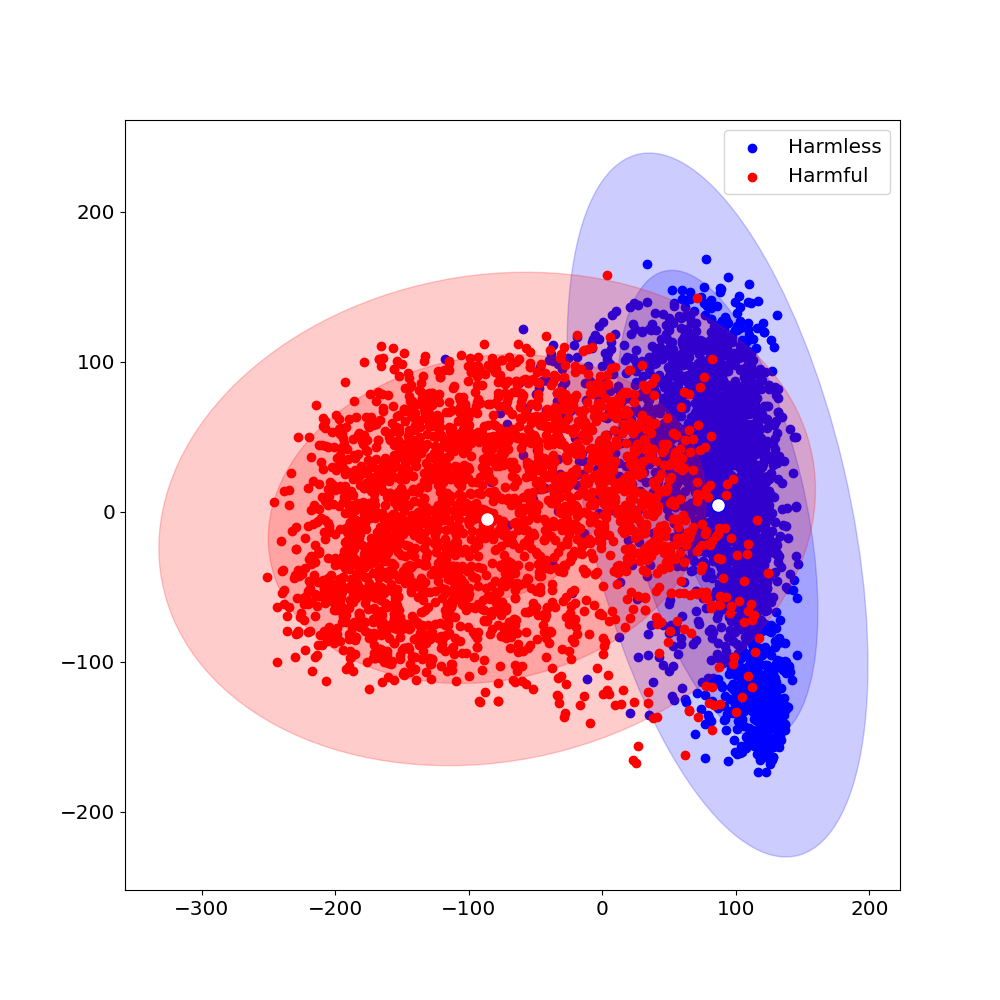} 
        \caption{$\pi_{\theta}$-en}
        \label{fig:firstsubfig5}
    \end{subfigure}
    \begin{subfigure}[b]{0.24\textwidth} 
        \centering
        \includegraphics[width=\textwidth]{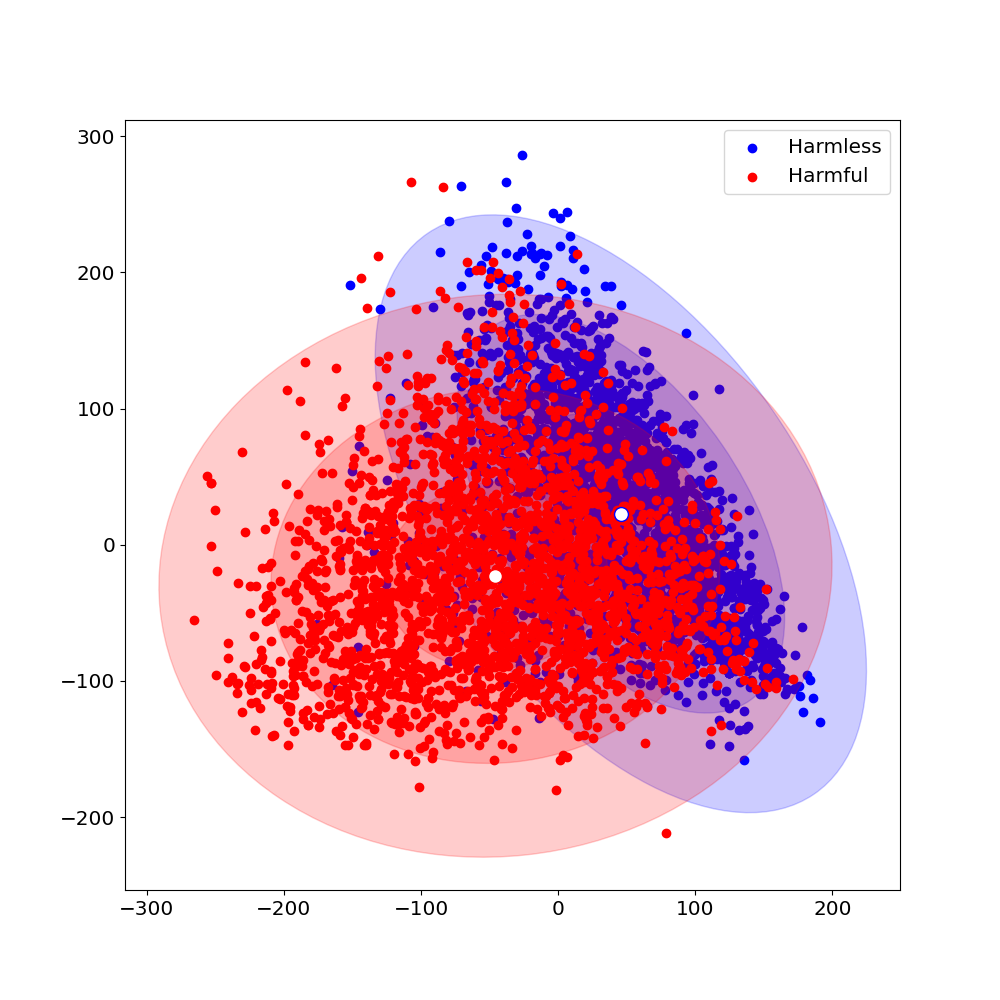}  
        \caption{$\pi_{\theta}$-hi}
        \label{fig:firstsubfig6}
    \end{subfigure}
    \begin{subfigure}[b]{0.24\textwidth} 
        \centering
        \includegraphics[width=\textwidth]{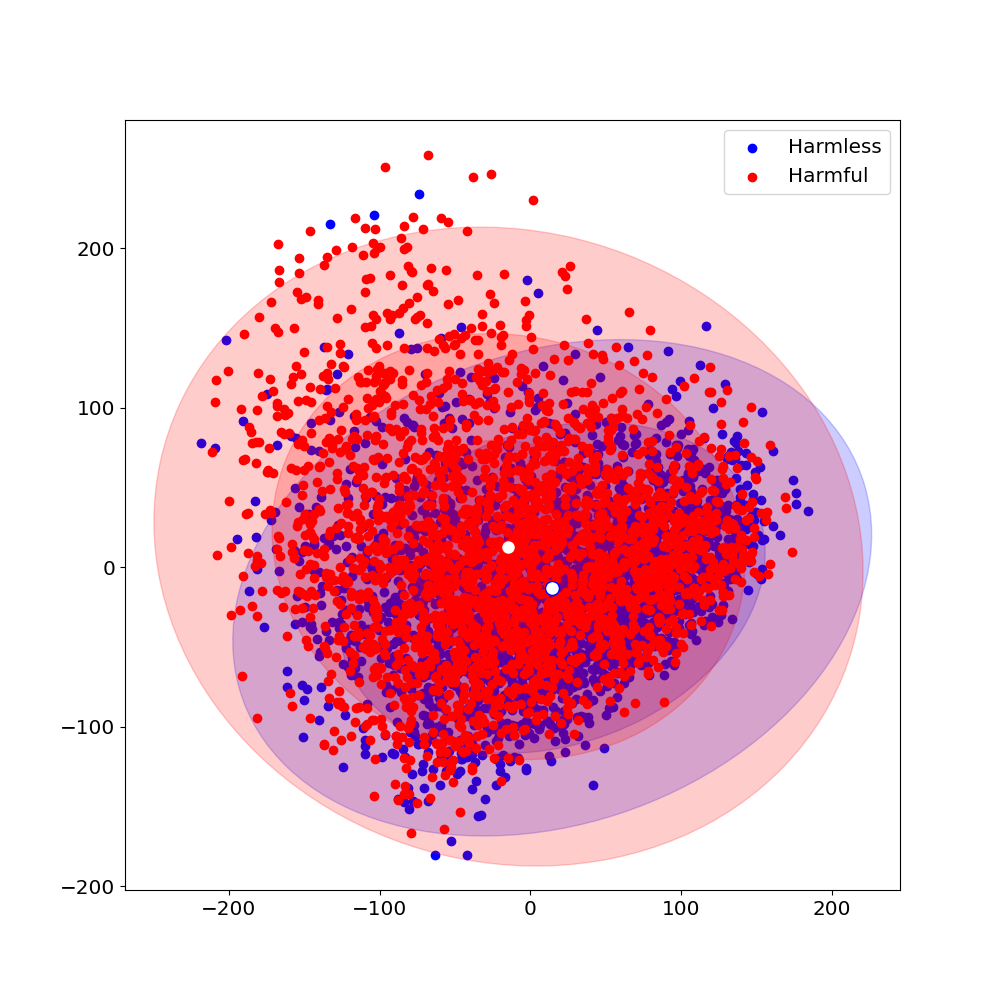} 
        \caption{$\pi_{\theta}$-zh}
        \label{fig:firstsubfig7}
    \end{subfigure}
    \begin{subfigure}[b]{0.24\textwidth} 
        \centering
        \includegraphics[width=\textwidth]{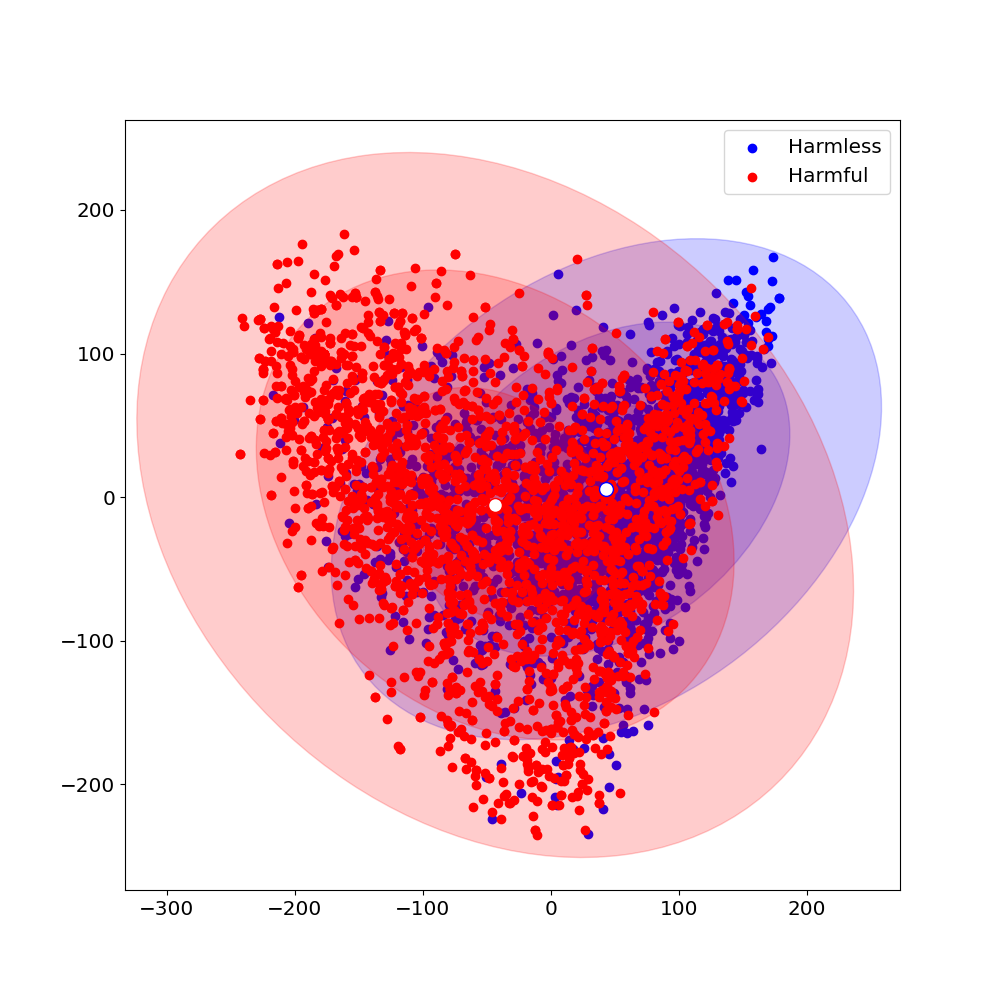} 
        \caption{$\pi_{\theta}$-de}
        \label{fig:firstsubfig8}
    \end{subfigure}

    \caption{Impact of Alignment on Hidden Representations in Gemma-3 for Multilingual Corpora.}
    \label{fig:before_after_alignment_gemma_3_all_langs}
\end{figure*}

\begin{figure*}[t]
    \centering
    
    \renewcommand{\thesubfigure}{\alph{subfigure}.\arabic{subfigure}} 
    \setcounter{subfigure}{0} 
    \begin{subfigure}[b]{0.24\textwidth} 
        \centering
        \includegraphics[width=\textwidth]{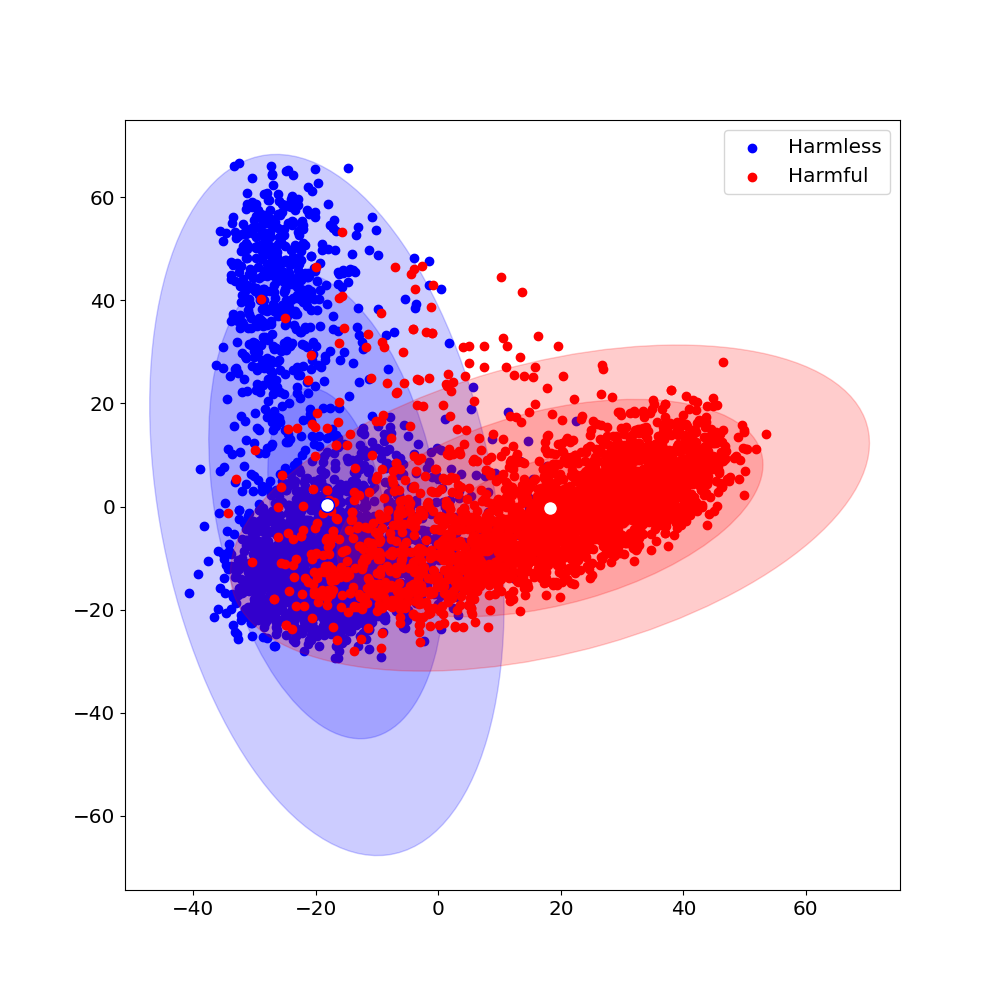} 
        \caption{$\pi_{\theta}$-en}
        \label{fig:firstsubfig1}
    \end{subfigure}
    \begin{subfigure}[b]{0.24\textwidth} 
        \centering
        \includegraphics[width=\textwidth]{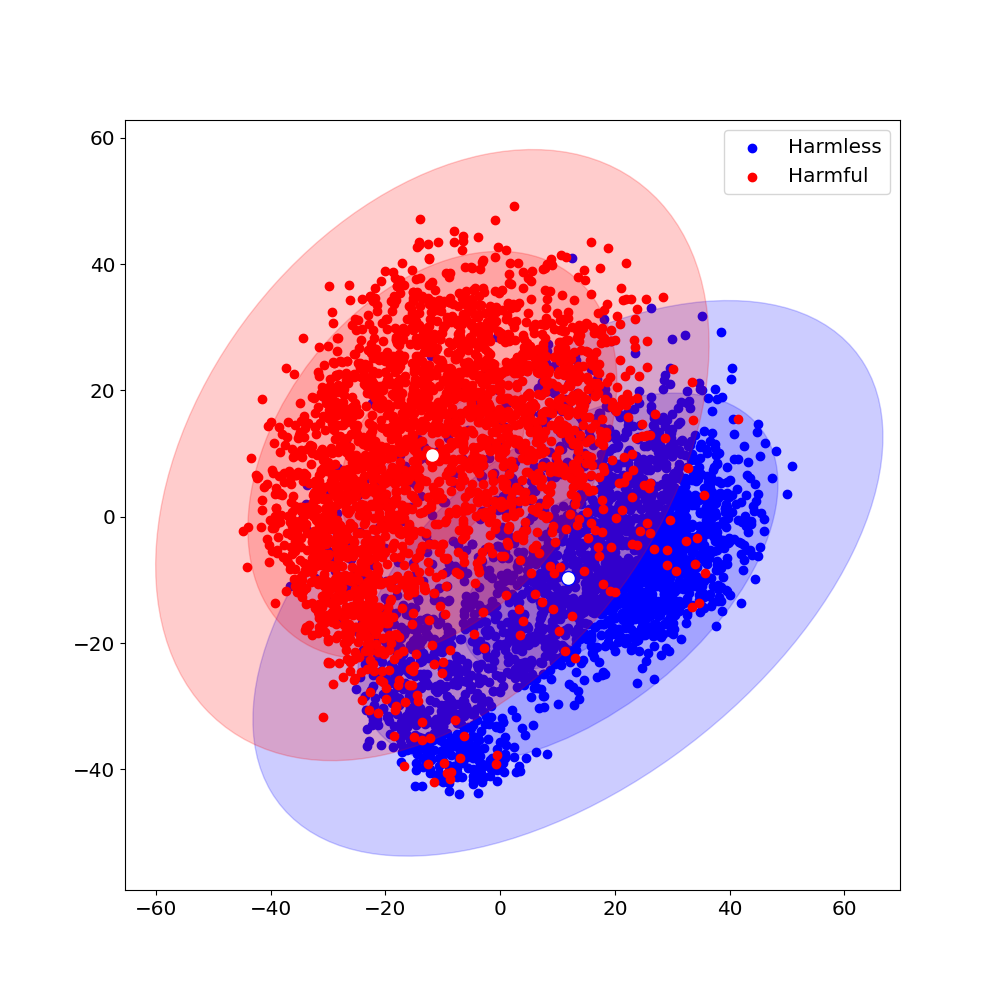}  
        \caption{$\pi_{\theta}$-hi}
        \label{fig:firstsubfig2}
    \end{subfigure}
    \begin{subfigure}[b]{0.24\textwidth} 
        \centering
        \includegraphics[width=\textwidth]{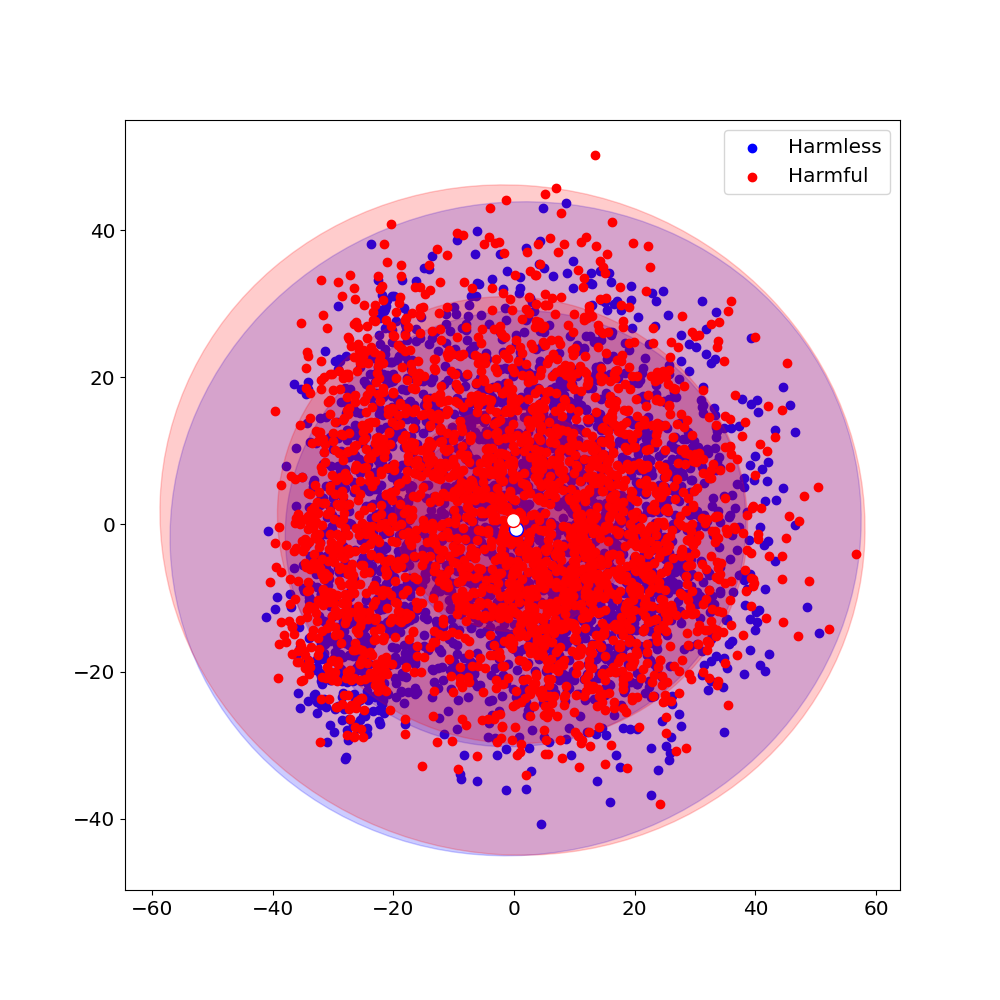} 
        \caption{$\pi_{\theta}$-zh}
        \label{fig:firstsubfig3}
    \end{subfigure}
    \begin{subfigure}[b]{0.24\textwidth} 
        \centering
        \includegraphics[width=\textwidth]{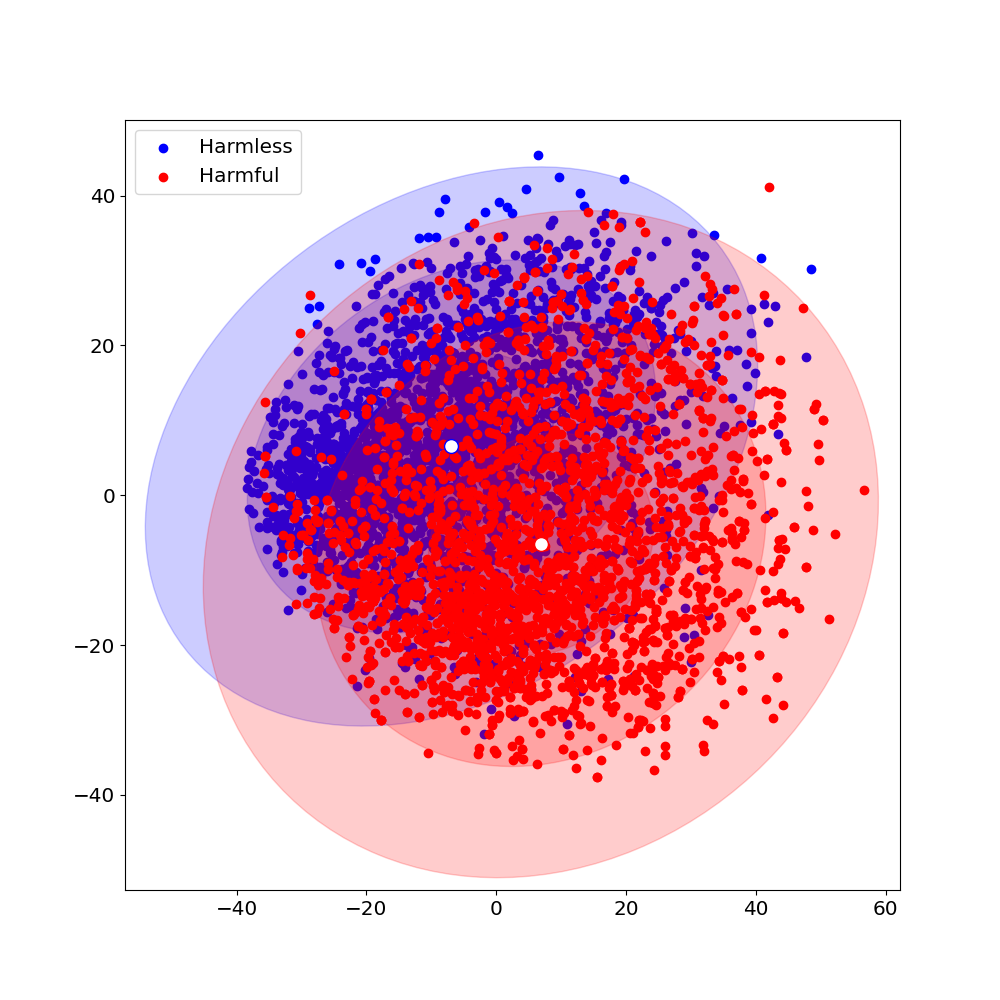} 
        \caption{$\pi_{\theta}$-de}
        \label{fig:firstsubfig4}
    \end{subfigure}

    \caption{Impact of Alignment on Hidden Representations in Phi-4 for Multilingual Corpora.}
    \label{fig:before_after_alignment_phi_4_all_langs}
\end{figure*}

For models other than Llama-2, visualizations of hidden representations before and after alignment are presented in Figures \ref{fig:before_after_alignment_qwen_2.5_all_langs} to \ref{fig:before_after_alignment_gemma_3_all_langs}.
For Phi-4, since only the aligned model checkpoint is available, only the representation analysis is shown in Figure \ref{fig:before_after_alignment_phi_4_all_langs}.

\section{Metrics of cluster quality, before and after alignment of LLMs}
\label{app:models_and_metrics}

The models we used in this study are mentioned below:
\begin{itemize}
    \item \textbf{Llama-2:} A suite of open-source foundational and fine-tuned chat models. 
    The pretraining corpus includes over 5\% non-English high-quality data, though evaluations primarily focus on English.
    
    \item \textbf{Qwen-2.5:} An instruction-following model optimized for long-context reasoning and diverse prompts.
    It explicitly supports 29 languages in output generation.

    \item \textbf{Llama-3.1:} 
    Fine-tuned versions use SFT and RLHF for preference alignment. 
    Supported languages include English, German, French, Italian, Portuguese, Hindi, Spanish, and Thai.

    \item \textbf{Llama-Guard:} A content safety classifier fine-tuned on Llama-3.1-8B.
    It classifies both inputs (prompt filtering) and outputs (response moderation) and supports multilingual safety alignment in languages same as Llama-3.1.

    \item \textbf{Gemma-2:} A 9B parameter instruction-tuned model primarily trained on English data. 
    However, the SFT phase incorporates some multilingual contexts.

    \item \textbf{Gemma-3:} A 12B parameter multimodal and multilingual model which supports 140 languages. 
    In its RL objectives, it uses variety of reward functions to improve helpfulness, reasoning and multilingual abilities, while minimizing model harmfulness.

    \item \textbf{Phi-4:} Trained on a mixture of synthetic datasets, filtered public domain websites, and traditional sources using SFT and DPO.
    Approximately 8\% of its training data is explicitly multilingual covering wide range of languages, including German, Spanish, French, Portuguese, Italian, Hindi and Japanese.
\end{itemize}

This selection of models enables a comprehensive evaluation of alignment in multilingual settings, covering diverse pretraining strategies, alignment techniques, and language coverage.
Table \ref{tab:overall_results} refers to the metrics used for comparison of PCA clusters with 10 components for all the models evaluated before and after alignment.

\begin{table*}[h]
\caption{Metric values of different LLMs before and after alignment on Balanced Toxicity Dataset. We use ``BD'' for Bhattacharyya distance, ``SS'' for silhouette score, and ``BCV'' for between-class variance. We use hyphen~(-) where model checkpoint is not available.}
\centering

\begin{tabular}{c | c | ccc | ccc}
\hline
\multirow{2}{*}{Model} & \multirow{2}{*}{Language} & \multicolumn{3}{c}{\textbf{Reference Model}} & \multicolumn{3}{c}{\textbf{Aligned Model}} \\ \cline{3-8}  
& &  \multicolumn{1}{c}{\begin{tabular}[c]{@{}c@{}}BD \end{tabular}} 
&  \multicolumn{1}{c}{\begin{tabular}[c]{@{}c@{}}SS\end{tabular}} 
&  \multicolumn{1}{c}{\begin{tabular}[c]{@{}c@{}}BCV\end{tabular}}  

&  \multicolumn{1}{c}{\begin{tabular}[c]{@{}c@{}}BD\end{tabular}} 
&  \multicolumn{1}{c}{\begin{tabular}[c]{@{}c@{}}SS\end{tabular}} 
&  \multicolumn{1}{c}{\begin{tabular}[c]{@{}c@{}}BCV\end{tabular}}   \\ 
\hline \hline

\multirow{4}{*}{Llama-2} & English & 0.035 & 0.0142 & 0.0303 & 2.5871 & 0.5433 & 0.3715 \\
& Hindi & 0.1837 & 0.1355 & 0.1309 & 0.6743 & 0.3036 & 0.1828 \\
& Chinese & 0.0044 & 0.0017 & 0.0033 & 0.0961 & 0.0878 & 0.0671\\
& German & 0.1905 & 0.0471 & 0.0409 & 0.3907 & 0.2825 & 0.2067 \\ \hline

\multirow{4}{*}{Qwen-2.5} & English & 0.3037 & 0.1326 & 0.0776 & 0.8365 & 0.315 & 0.1987 \\
& Hindi & 0.0309 & 0.0172 & 0.0631 & 0.1746 & 0.1047 & 0.0895 \\
& Chinese & 0.0208 & 0.0102 & 0.0119 & 0.0233 & 0.0138 & 0.0263 \\
& German & 0.0813 & 0.0482 & 0.033 & 0.077 & 0.0634 & 0.0703 \\ \hline

\multirow{4}{*}{Llama-3.1} & English & 0.1114 & 0.0268 & 0.0637 & 0.9639 & 0.3156 & 0.2047\\
& Hindi & 0.5402 & 0.2313 & 0.1646 & 0.3723 & 0.2253 & 0.1649 \\
& Chinese & 0.0029 & 0.0025 & 0.0053 & 0.0162 & 0.014 & 0.0124 \\
& German & 0.1262 & 0.0411 & 0.0405 & 0.2096 & 0.123 & 0.0904 \\ \hline

\multirow{4}{*}{Llama-Guard-3 } & English & 0.1114 & 0.0268 & 0.0637 & 0.8627 & 0.2971 & 0.2080 \\
& Hindi & 0.5402 & 0.2313 & 0.1646 & 0.2389 & 0.1456 & 0.1398 \\
& Chinese & 0.0029 & 0.0025 & 0.0053 & 0.1576 & 0.0923 & 0.072 \\
& German & 0.1262 & 0.0411 & 0.0405 &  0.2440 & 0.1697 & 0.1112 \\ \hline

\multirow{4}{*}{Gemma-2} & English & 0.1653 & 0.0565 & 0.0781 & 0.6046 & 0.2544 & 0.1368 \\
& Hindi & 0.5061 & 0.2099 & 0.1491 & 0.1522 & 0.0882 & 0.0795  \\
& Chinese & 0.0055 & 0.0051 & 0.0072 & 0.0182 & 0.0172 & 0.0202\\
& German & 0.0487 & 0.0301 & 0.0305 & 0.1976 & 0.1028 & 0.0629 \\ \hline

\multirow{4}{*}{Gemma-3} & English & 0.2087 & 0.0753 & 0.1016 & 1.1618 & 0.3913 & 0.2342 \\
& Hindi & 1.2436 & 0.2389 & 0.18 & 0.394 & 0.1794 & 0.1073  \\
& Chinese & 0.0046 & 0.0043 & 0.0102 & 0.0749 & 0.0362 & 0.0262  \\
& German & 0.107 & 0.0396 & 0.0426 & 0.2298 & 0.1287 & 0.079 \\ \hline

\multirow{4}{*}{Phi-4} & English & - & - & - & 1.1200 & 0.3712 & 0.1929 \\
& Hindi & - & - & - & 0.7817 & 0.2895 & 0.1475 \\
& Chinese & - & - & - & 0.0015 & 0.0004 & 0.0111  \\
& German & - & - & - & 0.2650 & 0.1543 & 0.0775  \\ \hline
\end{tabular}

\label{tab:overall_results}
\end{table*}

\end{document}